\documentclass{article}

\PassOptionsToPackage{numbers, compress}{natbib}

\usepackage[usenames,dvipsnames]{xcolor}

\usepackage{enumitem}
\usepackage[preprint,nonatbib]{neurips_2023}

\usepackage{wrapfig}
\usepackage[utf8]{inputenc} 
\usepackage[T1]{fontenc}    
\usepackage{hyperref}       
\usepackage{url}            
\usepackage{booktabs}       
\usepackage{amsmath}
\usepackage{amssymb}
\usepackage{mathtools}
\usepackage{amsthm}
\usepackage{amsfonts}
\usepackage{nicefrac}       
\usepackage{microtype}      
\usepackage{xcolor}         
\usepackage{comment}

\usepackage[pdftex]{graphicx}
\usepackage[export]{adjustbox}
\usepackage[font=small,labelfont=bf]{caption}
\usepackage{soul}
\usepackage{subfig}
\usepackage{xparse}
\newcommand{\bnm}{\begin{newmath}}
\newcommand{\enm}{\end{newmath}}

\newcommand{\bea}{\begin{eqnarray*}}%
\newcommand{\eea}{\end{eqnarray*}}%

\newcommand{\bne}{\begin{newequation}}
\newcommand{\ene}{\end{newequation}}

\newcommand{\bal}{\begin{newalign}}
\newcommand{\eal}{\end{newalign}}

\newenvironment{newalign}{\begin{align}%
\setlength{\abovedisplayskip}{4pt}%
\setlength{\belowdisplayskip}{4pt}%
\setlength{\abovedisplayshortskip}{6pt}%
\setlength{\belowdisplayshortskip}{6pt} }{\end{align}}

\newenvironment{newmath}{\begin{displaymath}%
\setlength{\abovedisplayskip}{4pt}%
\setlength{\belowdisplayskip}{4pt}%
\setlength{\abovedisplayshortskip}{6pt}%
\setlength{\belowdisplayshortskip}{6pt} }{\end{displaymath}}

\newenvironment{newequation}{\begin{equation}%
\setlength{\abovedisplayskip}{4pt}%
\setlength{\belowdisplayskip}{4pt}%
\setlength{\abovedisplayshortskip}{6pt}%
\setlength{\belowdisplayshortskip}{6pt} }{\end{equation}}

\newcounter{ctr}

\newcounter{mytable}
\def\mytable{\begin{centering}\refstepcounter{mytable}}
\def\endmytable{\end{centering}}

\newcounter{myfig}
\def\myfig{\begin{centering}\refstepcounter{myfig}}
\def\endmyfig{\end{centering}}

\newlength{\saveparindent}
\newlength{\saveparskip}
\newcommand{\E}{{\rm I\kern-.3em E}}

\renewcommand{\eqref}[1]{\mbox{Equation~(\ref{#1})}}

\def \part {part}

\DeclareMathOperator*{\argmax}{argmax}

\renewcommand{\paragraph}[1]{\vspace*{6pt}\noindent\textbf{#1}\;}

\def \blackslug{\hbox{\hskip 1pt \vrule width 4pt height 8pt
    depth 1.5pt \hskip 1pt}}
\def \qed{\quad\blackslug\lower 8.5pt\null\par}

\newcounter{mynote}[section]

\newcommand\ignore[1]{}

\newcounter{rcnote}[section]

\newcounter{mrnote}[section]

\newcounter{fknote}[section]

\newcounter{anote}[section]

\DeclareMathSymbol{\mlq}{\mathord}{operators}{``}
\DeclareMathSymbol{\mrq}{\mathord}{operators}{`'}

\newcommand{\rhf}[2]{R_{f, \gamma}}

\DeclareDocumentCommand{\edist}{o o}{
  \ensuremath{
    \IfNoValueTF{#1}{{d}}{{\sf d}(#1,#2)}
  }
}

\newcommand{\olrk}[1]{\ifx\nursymbol#1\else\!\!\mskip4.5mu plus 0.5mu\left(\mskip0.5mu plus0.5mu #1\mskip1.5mu plus0.5mu \right)\fi}

\NewDocumentCommand{\indseq}{ O{1} O{r} }{{#1}\ldots {#2}}

\usepackage{multirow}
\usepackage[textsize=tiny]{todonotes}
\usepackage{acronym}
\acrodef{IC}{integrated circuit}
\acrodef{EDA}{electronic design automation}
\acrodef{HDL}{hardware description language}
\acrodef{AIG}{and-inverter-graph}
\acrodef{RL}{reinforcement learning}
\acrodef{ML}{machine learning}
\acrodef{IP}{intellectual property}
\acrodef{RTL}{register transfer level}
\acrodef{DAG}{directed acyclic graph}
\acrodef{GCN}{graph convolutional network}
\acrodef{MCTS}{Monte Carlo tree search}
\acrodef{SOTA}{state-of-the-art}
\acrodef{PPA}{power, performance, and area}
\acrodef{QoR}{quality of result}
\acrodef{ADP}{area-delay product}
\acrodef{OOD}{out-of-distribution}
\newcommand{\solution}{INVICTUS}

\title{INVICTUS: Optimizing Boolean Logic Circuit Synthesis via Synergistic Learning and Search}
\usepackage{pifont}
\newcommand{\cmark}{\ding{51}}%
\newcommand{\xmark}{\ding{55}}%

\newcommand{\specialcell}[2][c]{%
  \begin{tabular}[#1]{@{}c@{}}#2\end{tabular}}

\usepackage{todonotes}
\usepackage{xargs}                      
\newcommandx{\marco}[2][1=]{\todo[linecolor=blue,backgroundcolor=blue!25,bordercolor=blue,#1]{#2}}
\usepackage{cleveref}
\author{%
  Animesh B. Chowdhury\thanks{Corresponding authors: abc586@nyu.edu,sg175@nyu.edu}  \\
  New York University\\
  \And
  Marco Romanelli \\
  New York University\\
  \AND
  Benjamin Tan \\
  University of Calgary \\
  \And
  Ramesh Karri \\
  New York University\\
  \And
  Siddharth Garg \\
  New York University\\
}

\begin{document}

\maketitle

\begin{abstract}
  Logic synthesis is the  first and most vital step in chip design. This steps converts a chip specification written in a hardware description language (such as Verilog) into an optimized implementation using Boolean logic gates. State-of-the-art logic synthesis algorithms have a large number of logic minimization heuristics, typically applied sequentially based on human experience and intuition. The choice of the order greatly impacts the quality (e.g., area and delay) of the synthesized circuit. In this paper, we propose \solution{}, a model-based offline reinforcement learning (RL) solution that automatically generates a sequence of logic minimization heuristics ("synthesis recipe") based on a training dataset of previously seen designs. A key challenge is that new designs can range from being very similar to past designs (e.g., adders and multipliers) to being completely novel (e.g., new processor instructions). 
  \solution{} is the first solution that uses a mix of RL and search methods joint with an online out-of-distribution detector to generate synthesis recipes over a wide range of benchmarks. Our results demonstrate significant improvement in area-delay product (ADP) of synthesized circuits with up to 30\% improvement over state-of-the-art techniques. Moreover, \solution{} achieves up to $6.3\times$ runtime reduction (iso-ADP) compared to the state-of-the-art.
  
\end{abstract}

\section{Introduction}
Modern chips are designed using sophisticated \ac{EDA} algorithms that automate
the conversion of the description of a function, for example, in a \ac{HDL} like Verilog or VHDL, to a physical layout that can be manufactured at a semiconductor foundry. 
EDA involves a sequence of steps, the first of which is \emph{logic synthesis}: this step converts a high-level HDL chip description into a low-level ``netlist'' of Boolean logic gates that implements the desired function. A netlist is a graph whose nodes are logic gates (\emph{e.g.}, ANDs, NOTs, ORs) and whose edges represent wires or connections between gates. Subsequent \ac{EDA} steps, referred to as physical design, place gates in the netlist in a chip layout and route wires between them. 

\begin{figure}[t]
\centering
\includegraphics[width=\columnwidth]{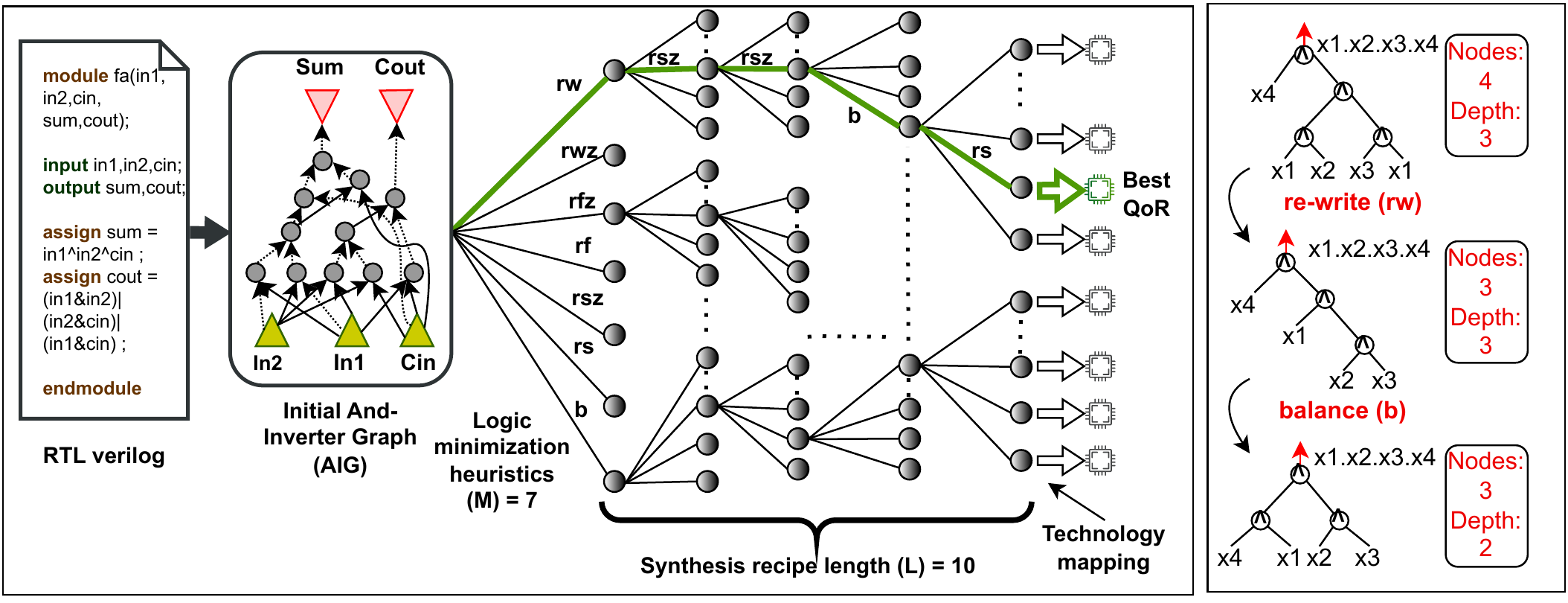}
\caption{(Left) A hardware design in Verilog is first transformed into an \ac{AIG}, i.e., a netlist containing only AND and NOT gates. Then a sequence of functionality-preserving transformations 
(here, picked from set \{\textbf{rw}, \textbf{rwz}, \ldots, \textbf{b} \}) is applied to generate an optimized \ac{AIG}. Each such sequence is called a synthesis recipe. The optimized \ac{AIG} is ``technology-mapped" to a larger set of Boolean logic gates, producing the final netlist. The synthesis recipe with the best \ac{QoR} (e.g., area or delay) is shown in green. (Right) Applying \textbf{rw} and \textbf{b} to an AIG results results in an optimized AIG with fewer nodes and lower depth.}
\label{fig:networkarchitecture1}
\end{figure}

EDA tools seek to optimize quality metrics like area, delay, and power consumption of the final chip. As the first step in the \ac{EDA} flow, the quality of the netlist produced by logic synthesis is crucial for the quality of all downstream steps and the final chip design. Beginning with an unoptimized netlist implementing a design, state-of-art logic synthesis algorithms perform a sequence of functionality-preserving transformations such as
redundant node elimination, reordering Boolean formulas, and streamlining node representations, to arrive at a final optimized netlist~\cite{yang2012lazy,dag_mischenko,riener2019scalable,cunxi_iccad,neto2022flowtune} (see Figure~\ref{fig:networkarchitecture1}). 
A specific sequence of transformations is called a ``\textbf{synthesis recipe}.'' Typically, designers use experience and intuition to pick a ``good" synthesis recipe from the solution space of all recipes and iterate if the quality of result is poor. This manual process is costly and time-consuming, especially for modern, complex chips.

Recent work has explored the use of machine learning and reinforcement learning (RL) ~\cite{googlerl,kurin2020can,lai2022maskplace,schmitt2021neural,yolcu2019learning,vasudevan2021learning,yang2022versatile, firstWorkDL_synth,lsoracle,cunxi_dac,drills,mlcad_abc,bullseye} 
to rapidly explore the solution space at different stages of the EDA flow, 
including for identifying high-quality synthesis recipes~\cite{cunxi_dac,drills,mlcad_abc,cunxi_iccad,neto2022flowtune,bullseye}. 
One line of work~\cite{cunxi_iccad,neto2022flowtune} proposes heuristic search methods, Monte-Carlo tree search (MCTS) in particular, to smartly explore the solution space for a given design. Although ~\cite{cunxi_iccad,neto2022flowtune} trains an agent during iterations of MCTS, these methods do \emph{not} learn from historical data---public or private repositories of past designs that are abundant in semiconductor companies and increasingly on the internet~\cite{openabc}. Recent work~\cite{bullseye} has shown that a predictive QoR model trained on past data in conjunction with simulated annealing-based search can outperform prior search-only methods. 


Here we propose \solution{}, a new approach that synergistically leverages both learning and search 
to rapidly identify high-quality synthesis recipes for a new design. \solution{} has three main components: 
(1) a pre-trained offline RL agent trained on a dataset of past designs; (2) RL agent-guided MCTS search over the synthesis recipe space for new designs; and (3) when new designs are novel with respect to the training set, out-of-distribution (OOD) to select between the learned policy and pure search. Via ablations, we show that all three components, learning, search, and OOD, are critical for high-quality results. OOD detection, in particular, reflects real-world semiconductor design---new designs use a mix of previously seen modules (adders, multipliers, communication buses etc.) while also including novel functionality. We find these trends reflected in standard logic synthesis benchmark sets. 

Evaluated on standard MCNC~\cite{mcnc} and EPFL~\cite{epfl} benchmarks, \solution{} achieves up to 10\% and 30\% reductions in area-delay product (ADP) compared to the state-of-the-art~\cite{neto2022flowtune,bullseye}. Conversely, \solution{} achieves the same ADP as prior work upto $6.3\times$ faster compared to \cite{neto2022flowtune} at iso-ADP. \solution{} successfully classified all \ac{OOD} benchmarks in MCNC and EPFL(except one) further pushing the \ac{ADP} reduction upto $7\%$.

\section{Proposed Approach}
\subsection{Problem Statement}
We begin by formally defining the optimization problem we seek to solve. The definition is in the 
context of ABC~\cite{abc}, the leading open-source logic synthesis tool that also forms the 
basis of commercial tools. 
As shown in Figure~\ref{fig:networkarchitecture1}, ABC first converts a Verilog description 
into an unoptimized \ac{AIG}, i.e., a graph $G_{0} \in \mathcal{G}$, where  
$\mathcal{G}$ is the set of all finite directed, acyclic, and bi-colored graphs. 
Note that since these attributes end-up having no bearing on the  problem, 
we will not discuss them further.
Next, ABC performs a functionality-preserving transformation on $G_{0}$. We view these 
as a finite set of $M$ actions, $\mathcal{A} = 
\{\textrm{rf}, \textrm{rm}, \ldots, \textrm{b}\}$
(see \S\ref{subsec:ls_heuristics} for more details). For ABC, $M=7$. Applying an action on an AIG 
yields a new AIG as determined by the synthesis function $\mathbf{S}: \mathcal{G} \times \mathcal{A} \rightarrow \mathcal{G}$.
Finally, a synthesis recipe $R \in \mathcal{A}^L$ is a sequence of $L$ actions that are applied to $G_{0}$ in order. Given a synthesis recipe $P = \{a_{0}, a_{1}, \ldots, a_{L-1}\}$ ($a_{i} \in \mathcal{A}$), then we obtain $G_{i+1} = \mathbf{S}(G_{i}, a_{i})$ for all $i\in [0,L-1]$ where $G_{L}$ is the final optimized AIG. 

Finally, let $\mathbf{QoR}: \mathcal{G} \rightarrow \mathbb{R}$ measure the quality of 
graph $G$, for instance, its inverse area-delay product (so larger is better).
Then, we seek to solve the following optimization problem:
\begin{align}
\argmax_{P \in \mathcal{A}^L}  \mathbf{QoR}(G_{L}), \, \, s.t. \, \, G_{i+1} = \mathbf{S}(G_{i}, a_{i}) \, \, \forall i\in [0,L-1].
\end{align}
We now discuss \solution{}, our proposed approach to solve this optimization problem. We note 
that in addition to $G_{0}$, the AIG to be synthesized, we will assume access to a training set of AIGs that can be used to aid optimization.

\subsection{Baseline MCTS-based Optimization}
The tree-structured solution space motivated prior work~\cite{cunxi_iccad,neto2022flowtune} to adopt an
MCTS-based approach that we briefly review here. A state $s$ in this setting is an input AIG $G_{0}$ and sequence of $l \leq L$ actions, i.e., $\{a_{0}, a_{1}, \ldots, a_{l}\}$. In a given state, any action $a \in \mathcal{A}$ can be picked as described above. Finally, the reward $\mathbf{QoR}(G_{L})$ is delayed to the final synthesis step.

While we refer the reader to past work~\cite{cunxi_iccad,neto2022flowtune} for more details.
In iteration $k$ of the search, \ac{MCTS} keeps track of two functions: $Q_{MCTS}^{k}(s,a)$ which is measure the ``goodness" of a state action pair, and $U_{MCTS}^{k}(s,a)$ which represents upper confidence tree (UCT) factor that encourages exploration of unseen or less visited states and actions. Exploiting known good states is balanced against exploration by selecting a policy $\pi_{MCTS}^k(s)$ that depends on both factors: 
\begin{equation}
\pi_{MCTS}^k(s) = \argmax_{a \in \mathcal{A}} \left(Q_{MCTS}^k(s, a) + U_{MCTS}^k(s, a)\right).
\label{eq:search-policy}
\end{equation}
Mor details on how these terms are updated are presented in \S\ref{subsec:baselineMCTS}, but we note that over iterations of MCTS, the policy tends towards the optimal with the exploration factor reducing and the exploitation factor increasing.




\subsection{RL-Agent Training and Architecture} \label{sec:Training-Prior}

\textbf{RL-agent training:} Building on the same principles as~\cite{silver2016mastering}, \solution{} improves \ac{MCTS} by training a \acf{RL} agent on previously seen circuits so as to guide \ac{MCTS} search on a new circuit to ``good" parts of the search space. 
Specifically, we use a dataset of $N_{tr}$ training circuits  to learn a policy $\pi_\theta(s,a)$ that outputs the probability of
taking action $a$ in state $s$ and
approximates the pure \ac{MCTS} policy on the training set. Here, $\theta$ represents the trainable parameters of the policy agent.

Given a new circuit, the upper confidence tree (UCT in~\autoref{eq:search-policy}) of \ac{MCTS} is biased towards favorable paths by computing a 
new $U_{MCTS}^{*k}(s, a)$ as:

\begin{equation}
    U_{MCTS}^{*k}(s, a)=\pi_{\theta}(s, a)\cdot{}U_{MCTS}^k(s, a).
    \label{eq:puct-loss-fn} 
\end{equation}
Here, the learned policy term $\pi_\theta(s,a)$
biases MCTS against exploring states that are learned to yield bad QoR, i.e., when $\pi_{\theta}(s, a)$ is small.


 Policy $\pi_{\theta}(s, a)$ is learned using a cross-entropy loss between the learned policy and the MCTS policy over samples picked from a replay buffer.
 We outline the pseudocode for RL-training in the appendix (\autoref{alg:qi}).

\begin{wrapfigure}[26]{r}{0.35\textwidth}
\centering
    \includegraphics[width=0.3\textwidth]{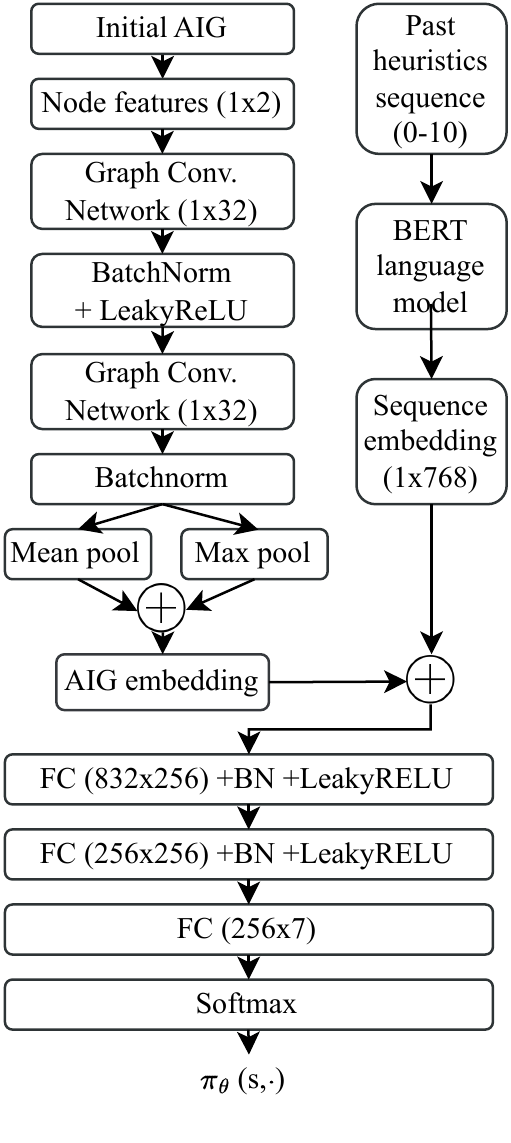}
    \caption{Policy network architecture. BN: Batch Normalization, FC: Fully connected layer}
\label{fig:networkArchitecture}
\vspace{-1em}
\end{wrapfigure}

\textbf{Policy network architecture:} Our policy network (\autoref{fig:networkArchitecture}) takes two inputs, (1) an initial AIG $G_{0}$, and (2) a sequence of $l\le L$ actions taken thus far, and outputs a probability distribution over the next action. Because the two inputs are in different formats, the policy network has two parallel branches that learn embeddings of the AIG and partial recipe. These embeddings are then concatenated and followed by additional layers to produce the final output.

For the AIG input, we employ a 3-layer \ac{GCN}~\cite{kipf2016semi} architecture to represent the \ac{AIG} as a netlist embedding ($h_{AIG}$). We use LeakyRELU as the activation function and apply batch normalization before each layer. (See appendix \S\ref{subsec:aigNetwork} for details.)  
For recipe embeddings, We used a pre-trained BERT (Bidirectional Encoder Representations from Transformers)~\cite{devlin2018bert} model to encode synthesis recipes. BERT embeddings capture the context of a sequence of minimization heuristics and concatenates it with the $h_{AIG}$. The concatenated outputs are passed through three additional fully-connected layers.

In contrast to previous research~\cite{bullseye}, we use BERT two primary reasons: 1) BERT's capacity to retain the contextual relationships within a sequence of actions in a synthesis recipe, thanks to its transformer-based architecture. Its self-attention mechanism is capable of discerning and encoding the inter-dependencies among various steps in the action sequence. 2) It can process variable-length inputs, producing fixed-length outputs, aligning with our requirements for generating consistent vector representations.

\subsection{Synergistic Learning and Search}



As we noted before, hardware designs frequently contain familiar and entirely new components. 
While we expect the learned RL-agent to help significantly on inputs similar to those in the training data, learning can hurt performance on novel inputs by biasing search towards low QoR regions of the search space. 

Thus, we propose an \acf{OOD} solution for using \ac{MCTS} with pre-trained agents: i.e., we use \ac{MCTS} search with pre-trained agent if the new design is in-distribution with respect to training data, and otherwise, use pure \ac{MCTS}. 

Specifically, we use the cosine distance metric ($\Delta_{cos}(h_{k_1},h_{k_2})=1 - \frac{h_{k_1} \cdot h_{k_2}}{|h_{k_1}||h_{k_2}|}$) between the learned AIG representations of AIGs $G_1$ and $G_2$ to measure distance between AIGs.  
To modulate the balance between the prior learned policy and pure search, we update the UCT terms with a  hyper-parameter $\alpha \in [0,1]$ as follows:
    $U_{MCTS}^{*k}(s, a)=\pi_{\theta}(s, a)^{\alpha}\cdot{} U_{MCTS}^{k}(s, a)$. 
When $\alpha=0$ the policy network is turned off and we implicitly default to pure search. Alternately when $\alpha=1$ we revert to the prior learning and search-based solution. Although we later discuss how $\alpha$ can potentially be set to any value between 0 and 1, our proposed solution sets $\alpha$ to a binary value based on cosine distance:

\begin{align}
    \alpha= \begin{cases}
1  & \quad  \delta^{min}_{test} < \delta_{th}, \\
0 & \quad otherwise.
\end{cases}\label{eq:alpha-factor}
\end{align}
where threshold $\delta^{min}_{test}$ is the smallest cosine distance between the test input and AIGs in the training dataset, i.e., $\min\limits_{h_i \in D_{train}} \Delta_{cos}(h_{test},h_i)$, and $\delta_{th}$ is determined based on validation data. 
Specifically, we run each design in the validation set using the RL-agent guided and pure search. 
We then set $\delta_{th}$ to maximize geometric mean performance on the validation set.

\autoref{fig:invictusFlow} outlines the proposed \solution{} flow. 
Here, $h_k$ indicates \ac{AIG} embedding of design $k$. Cosine distance measures the similarity of two embeddings. 
For an unseen design $G_{test}$, we obtain its \ac{AIG} embedding $h_{test}$ by passing it through our pre-trained agent. We then measure $\Delta_{cos}$ $h_{test}$ to the embeddings of designs used in training. We consider the cosine distance with the closest embedding~\autoref{eq:alpha-factor} to decide how to proceed: with standard MCTS or Agent-guided search. 

\section{Empirical Evaluation}
\label{sec:empiricalEval}

We present our experimental setup 
and compare \solution{} against \ac{SOTA} methods on ADP and runtime reductions.

\subsection{Experimental Setup}


\begin{table}[t]
\centering
\caption{Datasets used in our work}
\footnotesize
\begin{tabular}{ccl}
\toprule
Dataset & Splits & Circuits \\
\midrule
\multirow{3}{*}{ MCNC } & Train & alu2, apex3, apex5, b2, C1355, C5315, C2670, prom2, frg1, i7, i8, m3, max512, table5 \\
& Valid & apex7, c1908, c3540, frg2, max128, apex6, c432, c499, seq, table3, i10 \\
& Test & pair, max1024, alu4, apex1, apex2, apex4, c6288, c7552, i9, m4, prom1, b9, c880 \\
\midrule
\multirow{4}{*}{\specialcell{ EPFL \\ arith}} & I & \textbf{Train:} adder, div, log2, sin, sqrt, multiplier,max \textbf{Test:} square,bar \\
& II & \textbf{Train:} max, square, bar, div, sin, multiplier \textbf{Test:} adder, sqrt, log2 \\
& III & \textbf{Train:} adder, div, log2, sqrt, max, square, bar \textbf{Test:} multiplier, sin \\
& IV & \textbf{Train:} adder, log2, sqrt, square, bar, multiplier, sin \textbf{Test:}div, max \\
\midrule
\multirow{5}{*}{\specialcell{ EPFL \\ random }} & I & \textbf{Train:} cavlc, ctrl, dec, i2c, int2float, mem\_ctrl, priority, router \textbf{Test:} arbiter, voter \\
& II & \textbf{Train:} arbiter, ctrl, i2c, int2float, mem\_ctrl, priority, voter \textbf{Test:} cavlc, router \\
& III & \textbf{Train:} arbiter, cavlc, i2c, int2float, mem\_ctrl, router, voter \textbf{Test:} ctrl, priority \\
& IV & \textbf{Train:} arbiter, cavlc, ctrl, i2c, int2float, priority, router, voter \textbf{Test:} mem\_ctrl \\
& V & \textbf{Train:} arbiter, cavlc, ctrl, dec, mem\_ctrl, priority, router, voter \textbf{Test:} i2c, int2float \\
\bottomrule
\end{tabular}
\label{tab:datasetSplit}
\end{table}
\textbf{Datasets:} We consider three popular datasets used by the logic synthesis community: MCNC~\cite{mcnc}, EPFL arithmetic and EPFL random control benchmarks~\cite{epfl}. MCNC benchmarks have 38 circuits ranging from 100 to 8000 node \ac{AIG}s. EPFL benchmarks are of two different types: arithmetic and random control. The EPFL arithmetic benchmarks perform operations like additions, multiplications etc. and have between
1000-44000 nodes. These benchmarks share common sub-modules, for examples, multipliers are typically implemented by stacking adders.
Finally, the EPFL random control benchmarks consist of finite-state machines, routing logic and 
other random functions with between 100 to 46000 nodes.

\textbf{Train-test split:} We train dataset-specific \ac{RL} agents to evaluate performance of \solution{}
using train-validation-test splits motivated by ~\cite{fu2020d4rl}. The splits used for each datatset are discussed below:
\begin{enumerate}  [noitemsep,nolistsep,leftmargin=0.5cm]
    \item \textbf{MCNC:} We divide MCNC dataset circuits into three sets (~\autoref{tab:datasetSplit}): training, validation and test sets. The training set contains 14 circuits; validation and test data consists of 12 circuits each. 
    We train a single MCNC agent since it is the largest benchmark suite and provides
    sufficiently many circuits for training, validation and test. 
    \item \textbf{EPFL arithmetic:} We create four variants of the EPFL arithmetic benchmark suite. In the first three variants, we train arithmetic agents I, II and III using a 7-2 split of training and test data. 
    Arithmetic agent IV is trained using a 6-3 split of training and test data. This strategy ensures that each of the nine EPFL arithmetic circuits appears in atleast one (in fact exactly one) test set, thus allowing us to report results on each circuit in the benchmark suite. The splits are performed randomly. 
    For validation data, we combine training circuits for each agent with unseen circuits from the MCNC benchmark suite (\texttt{alu2} and \texttt{apex7}), along with four EPFL control circuits. 
    We use validation data to set $\delta_{th}$( \ac{OOD} hyperparameter). 
    \item \textbf{EPFL random control:} Similar to arithmetic benchmarks, we divide random control benchmarks into four 7-2 split and one 8-1 split and train five \ac{RL} agents. We create validation data following the above strategy. 
\end{enumerate}

\textbf{Optimization objective and metrics:} We seek to identify the best $L=10$ synthesis recipes. Consistent with prior works~\cite{drills,mlcad_abc,neto2022flowtune}, we use area-delay product (ADP) as evaluation metric. Area and delay values are obtained using a 7nm technology library post technology mapping of the synthesized AIG. As a baseline, we compare against the ADP of the \texttt{resyn2} synthesis recipe as is also done in prior work~\cite{neto2022flowtune,bullseye}. In addition to ADP reduction, we also report runtime reduction of 
\solution{} at iso-ADP, i.e., how much faster \solution{} is in reaching the best ADP achieved by competing methods.

\textbf{Training details and hyper-parameters:} We present our network architecture in ~\autoref{fig:networkArchitecture}. We use He initialization~\cite{he2015delving} for the weights of our RL agents. Following~\cite{googleRLempirical}, we multiply the weights of the final layer with $0.01$ to prevent bias towards any action. We train our agents for $50$ epochs. We used the Adam optimizer set the initial learning rate to $0.01$.

In each epoch, we perform MCTS on each training circuit. The MCTS search budget ($K$) is set to $512$ at each level of synthesis. At each level, the replay buffer stores the best experience tuple having information about the state and action probability distribution scores collected from Monte Carlo rollouts.
After performing MCTS simulations on training circuits, we sample $L\times N_{tr}$ ($N_{tr}$ is the number of training circuits) experiences from the replay buffer (of size $2\times L \times N_{tr}$) to train the agent. The \ac{RL} agent minimizes the cross entropy loss between $\pi_\theta$ and the MCTS agent $\pi_{MCTS}$. To stabilize training, we normalize our \ac{QoR} rewards (see Appendix~\ref{subsec:qorRewardNorm}) and  clip it to $[-1,+1]$~\cite{mnih2015human}. 

We performed the training on a server machine with one NVIDIA RTX A4000 with 16GB VRAM. 
The major bottleneck during training is the synthesis time for running ABC; actual gradient updates are relatively inexpensive.
Agent training took around 27 hours for MCNC and ~7 days for EPFL arithmetic and random control.

\begin{wrapfigure}[20]{r}{0.5\textwidth}
\centering
\setlength\tabcolsep{3.5pt}
\resizebox{0.5\textwidth}{!}{%
\begin{tabular}{lrrrrrr}
\toprule
\multirow{3}{*}{Designs}& \multicolumn{5}{c}{ADP reduction (in \%)} & \multirow{3}{*}{\begin{tabular}{@{}c@{}}Iso-\\ADP\\Speed-\\Up\end{tabular}} \\
\cmidrule(lr){2-6}
& \multirow{2}{*}{\begin{tabular}{@{}c@{}}Online-\\RL\\\cite{mlcad_abc}\end{tabular}}& \multirow{2}{*}{\begin{tabular}{@{}c@{}}SA+\\Pred.\\\cite{bullseye}\end{tabular}} & \multirow{2}{*}{\begin{tabular}{@{}c@{}}MCTS\\\cite{neto2022flowtune}\end{tabular}} &
\multicolumn{2}{c}{\solution{}} & \\
\cmidrule(lr){5-6}
& & & & $\alpha=1$ & +OOD & \\
\midrule
alu4 & 20.61 & 17.58 & 17.05 & \textbf{21.95} &  \textbf{21.95}(\cmark) & 4.5x \\
apex1 & 6.58 & 17.01 & 15.95 & \textbf{17.54} & \textbf{17.54}(\cmark)& 2.6x\\
apex2 & 8.12 & 15.58 & 13.06 & \textbf{17.51} & \textbf{17.51}(\cmark) & 4.7x \\
apex4 & 13.53 & 13.01 & 13.01 & \textbf{13.95} & \textbf{13.95}(\cmark) & 3.2x\\
i9 & 39.35& 46.45 & 46.89 & \textbf{53.97} & \textbf{53.97}(\cmark) & 1.6x\\
m4 & \textbf{20.95} & 18.16 & 14.98 & 20.05 & 20.05(\cmark) & 1.7x \\
prom1 & 4.97& 8.53 & 6.50 & \textbf{11.23} & \textbf{11.23}(\cmark) & 2.5x \\
b9 & 17.92 & 23.65 & 23.21 & \textbf{24.10}  &\textbf{24.10}(\cmark) & 6.3x\\
c880 & 16.23 & 19.95 & 17.75 & \textbf{24.58} & \textbf{24.58}(\cmark) & 6.3x \\
c7552 & 20.21 & 17.62 & \textbf{20.45} & 12.78  & \textbf{20.45}(\xmark) & 1.0x\\
pair & 4.73 & 10.02 & \textbf{13.10} & 12.65 &  \textbf{13.10}(\xmark)& 1.0x\\
max1024 & 11.39 & \textbf{20.27} & 19.65 & 18.32 & 19.65(\xmark) & 1.0x\\
\midrule
Geomean & 12.80 & 17.34 & 16.66 & 18.84 & \textbf{19.76} & 2.5x \\
Win ratio & 1/12 & 1/12 & 2/12 & 8/12 & 10/12 & 9/12\\
\bottomrule
\end{tabular}
}
    \captionof{table}{Area-delay reduction over \texttt{resyn2} on MCNC benchmarks. \cmark denotes \solution{} used RL agent during search whereas {\xmark} denotes standard MCTS}
\label{tab:qorMCNC}
\end{wrapfigure}

\textbf{Evaluation:} We compare \solution{} with three main methods: (1) standard \ac{MCTS}~\cite{neto2022flowtune}; 
(2) MCTS augmented with an \ac{RL} agent trained online (i.e., on the circuit being optimized) but not on past training data ~\cite{mlcad_abc}; and (3) simulated annealing (SA) with \ac{QoR} predictor learned from training data~\cite{bullseye}. Methods (1) and (3) and SA are the current \ac{SOTA} methods. For completeness, we 
also compare with (2) although it has already been shown to underperform (1). During evaluations on test circuits, we give each technique a  budget of $100$ synthesis runs.

\subsection{Results}

\begin{figure}[t]
\centering
   \subfloat[alu4]{{{\includegraphics[width=0.25\columnwidth, valign=c]{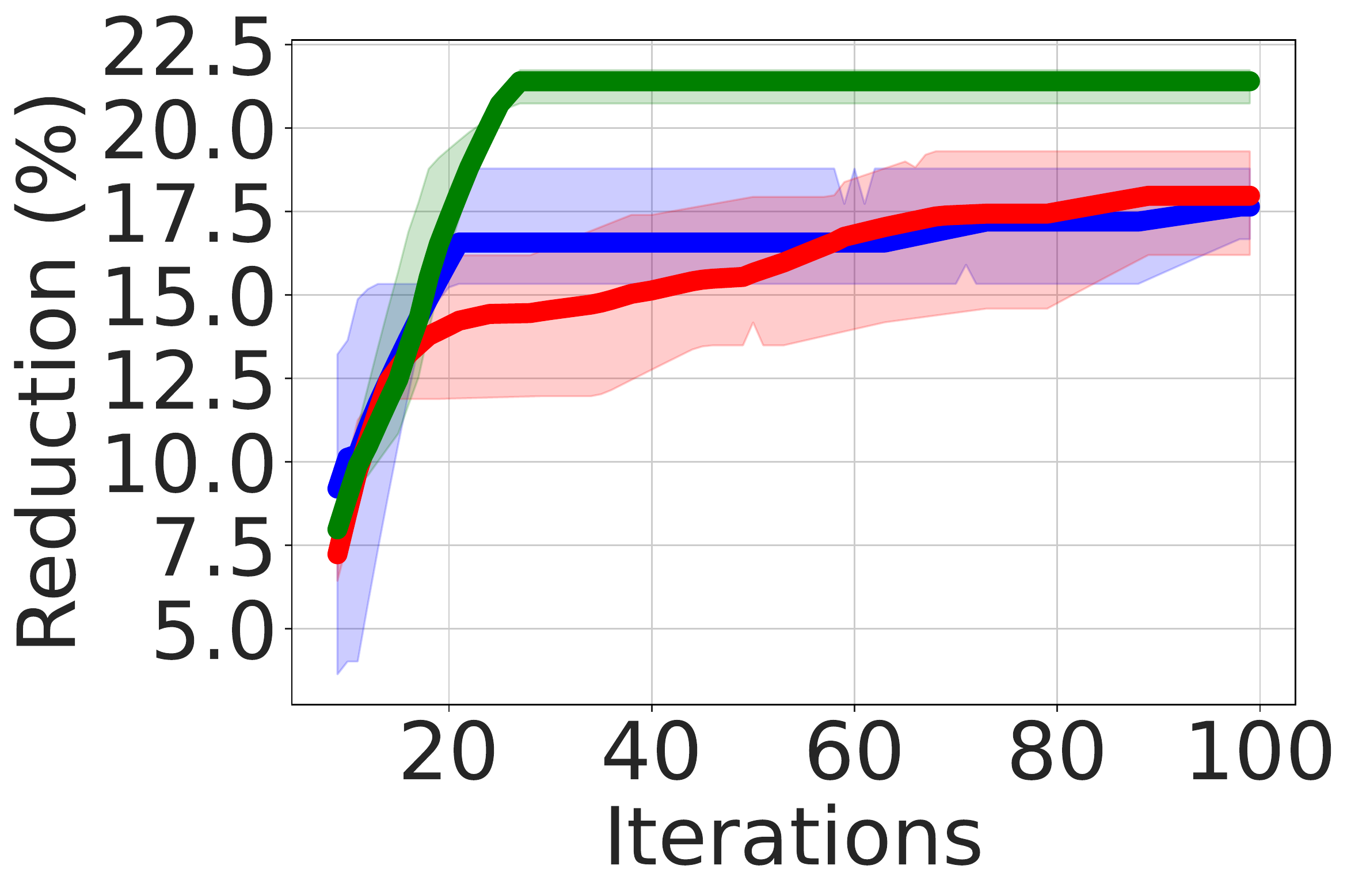} }}%
	}
	\subfloat[apex1]{{{\includegraphics[width=0.25\columnwidth, valign=c]{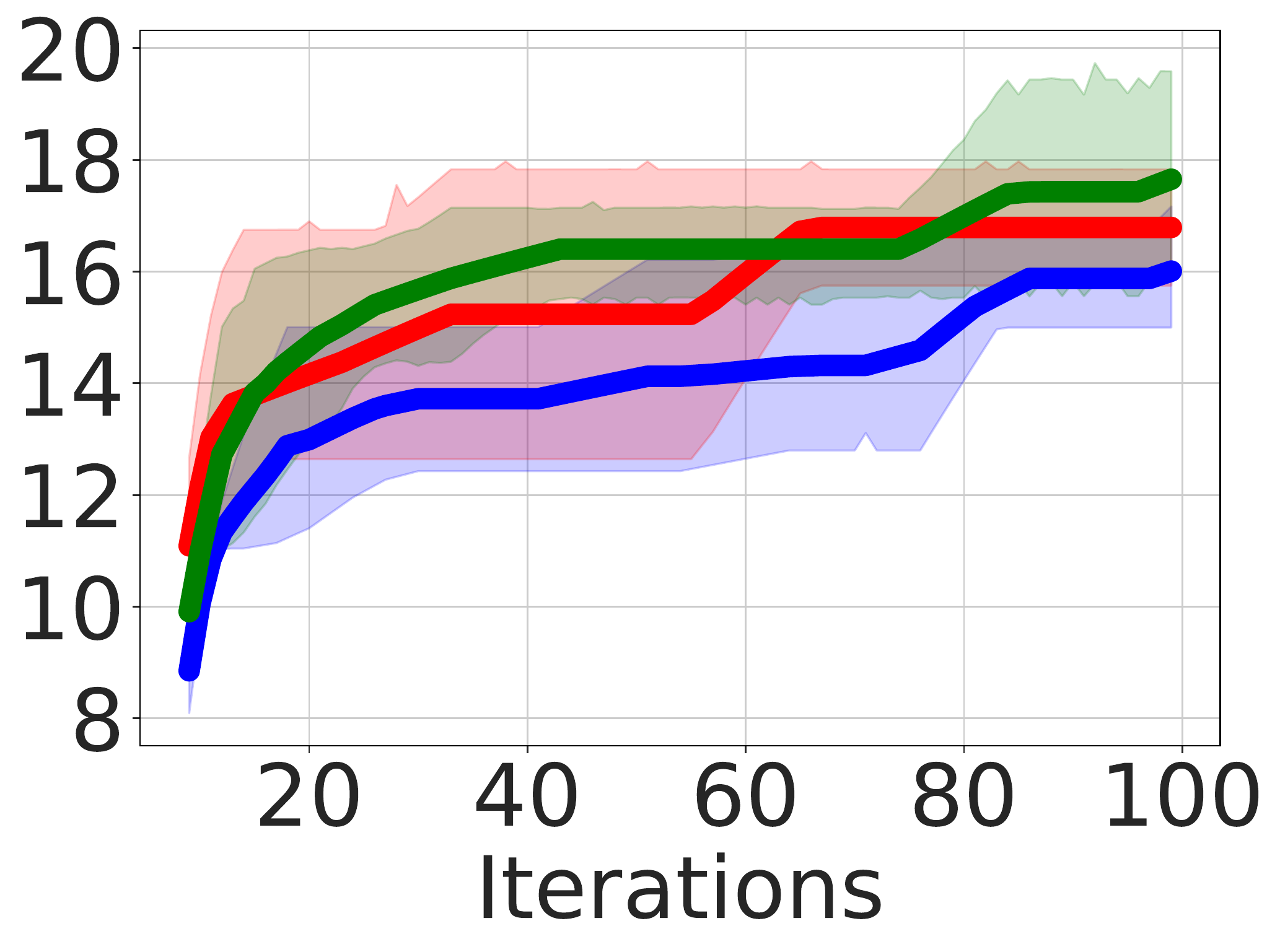} }}
	}
        \subfloat[apex2]{{{\includegraphics[width=0.25\columnwidth, valign=c]{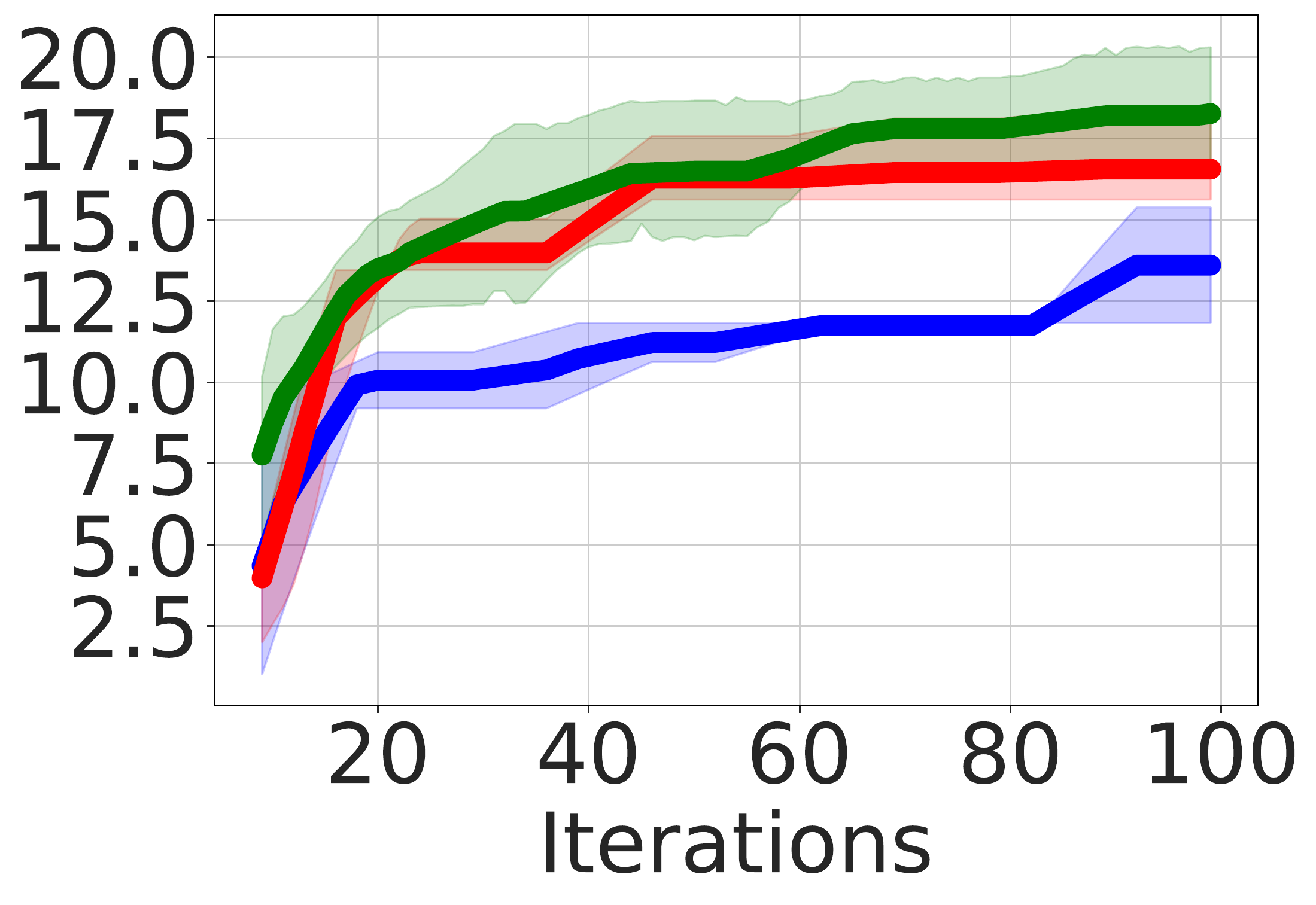} }}
	}
	\subfloat[apex4]{{{\includegraphics[width=0.25\columnwidth, valign=c]{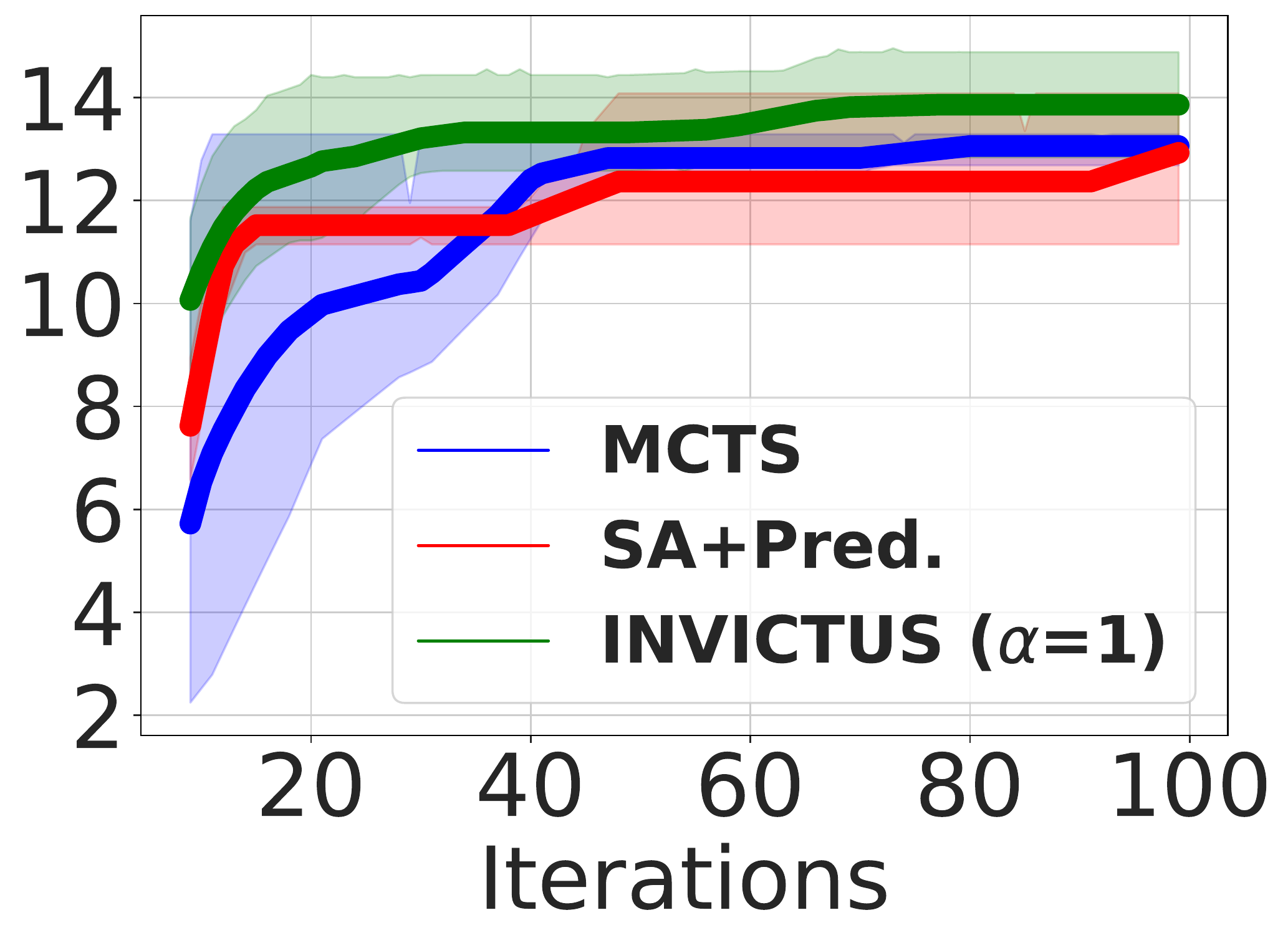} }}
	}
\vspace{-0.2in}
	\subfloat[b9]{{{\includegraphics[width=0.25\columnwidth, valign=c]{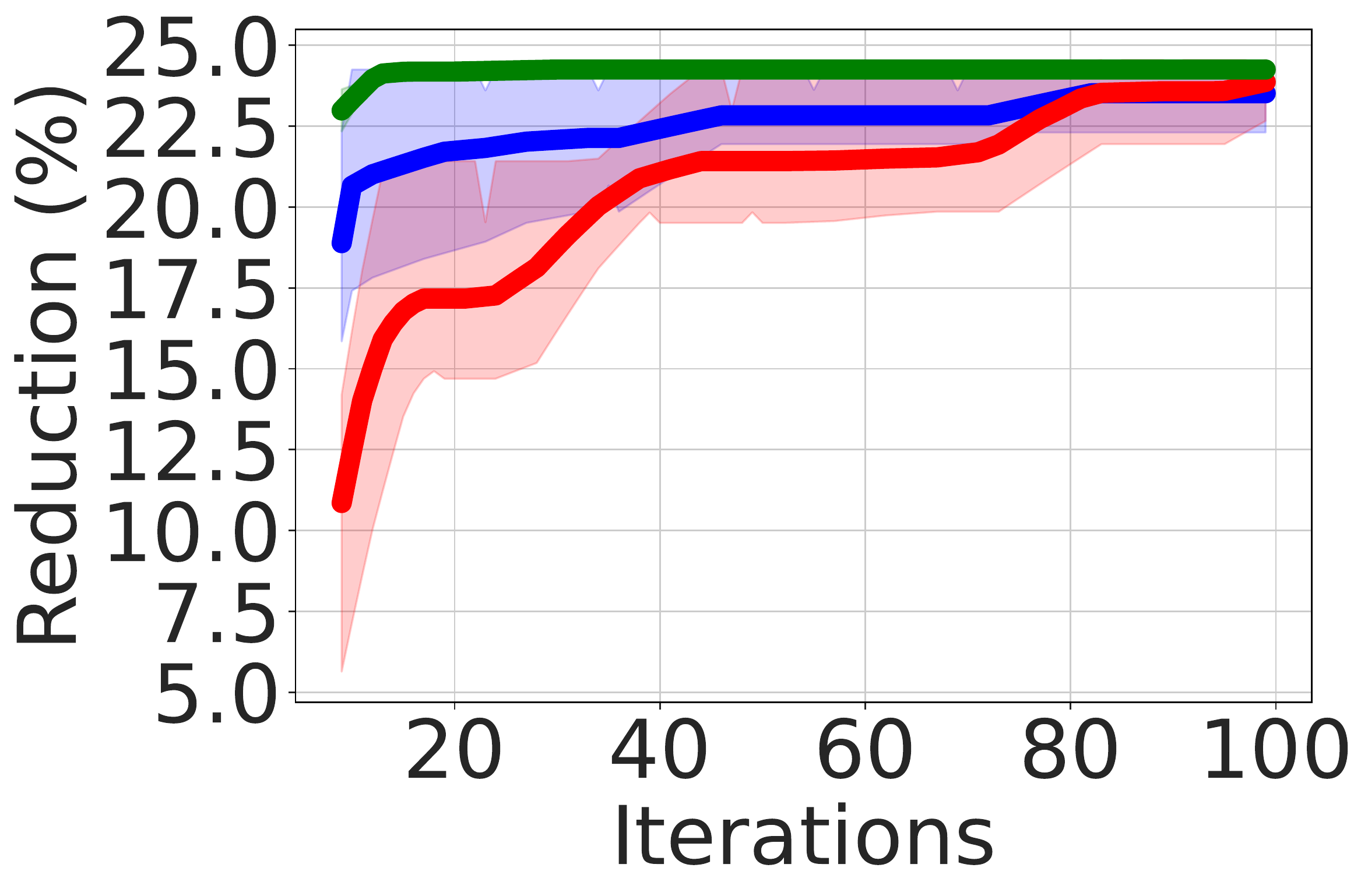} }}}
        \subfloat[c880]{{{\includegraphics[width=0.25\columnwidth, valign=c]{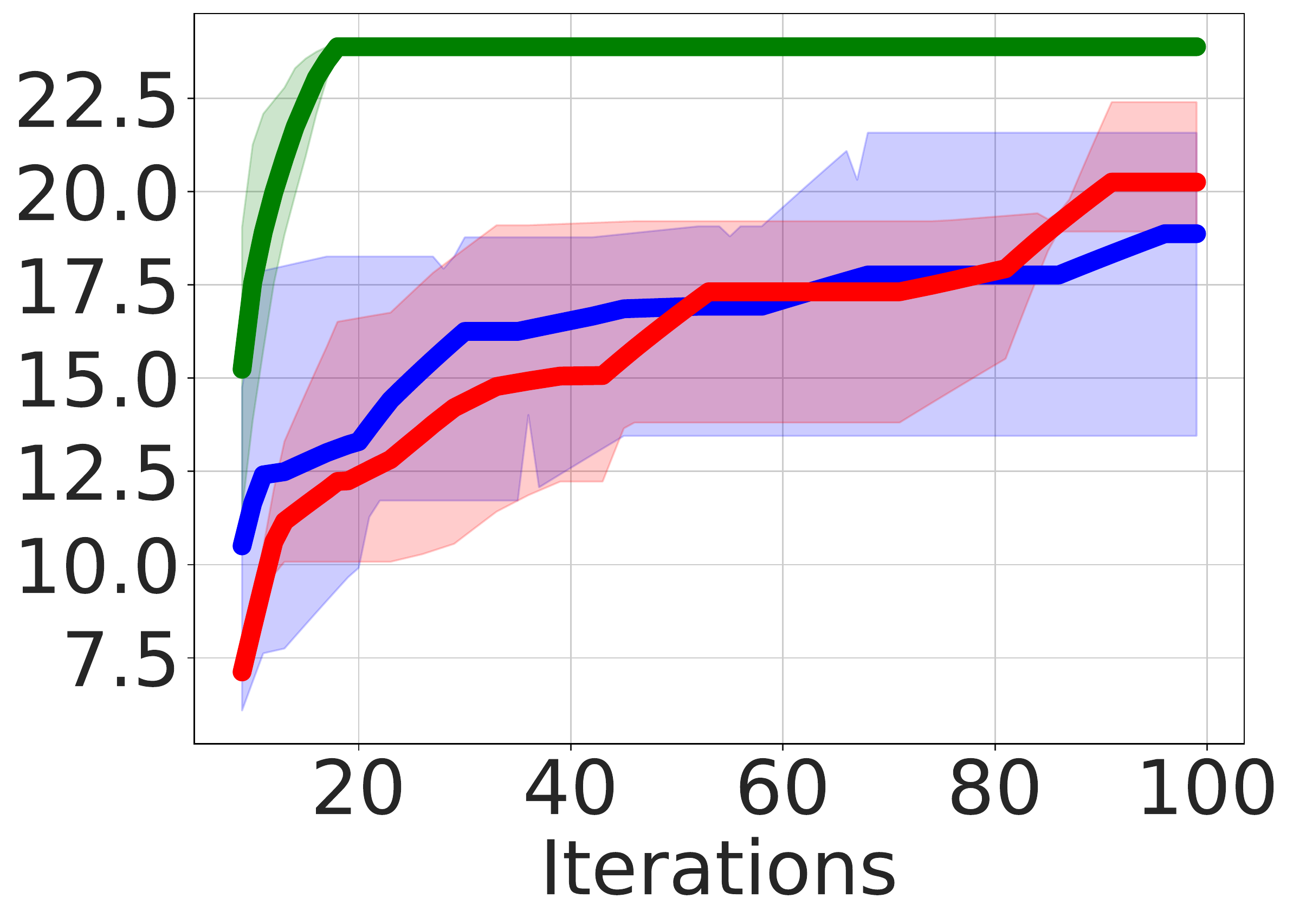} }}
	}
	\subfloat[prom1]{{{\includegraphics[width=0.25\columnwidth, valign=c]{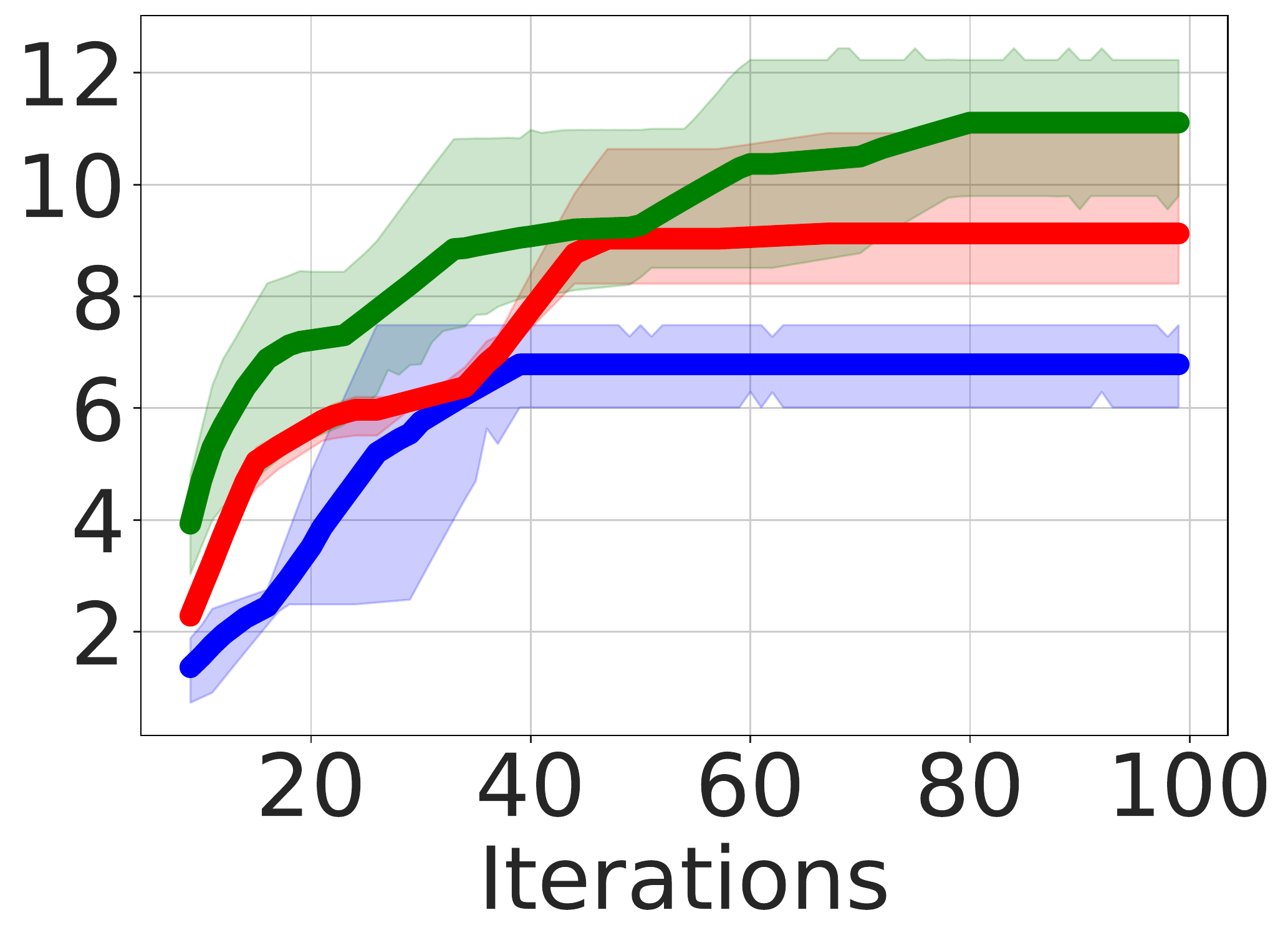} }}
	}
	\subfloat[i9]{{{\includegraphics[width=0.25\columnwidth, valign=c]{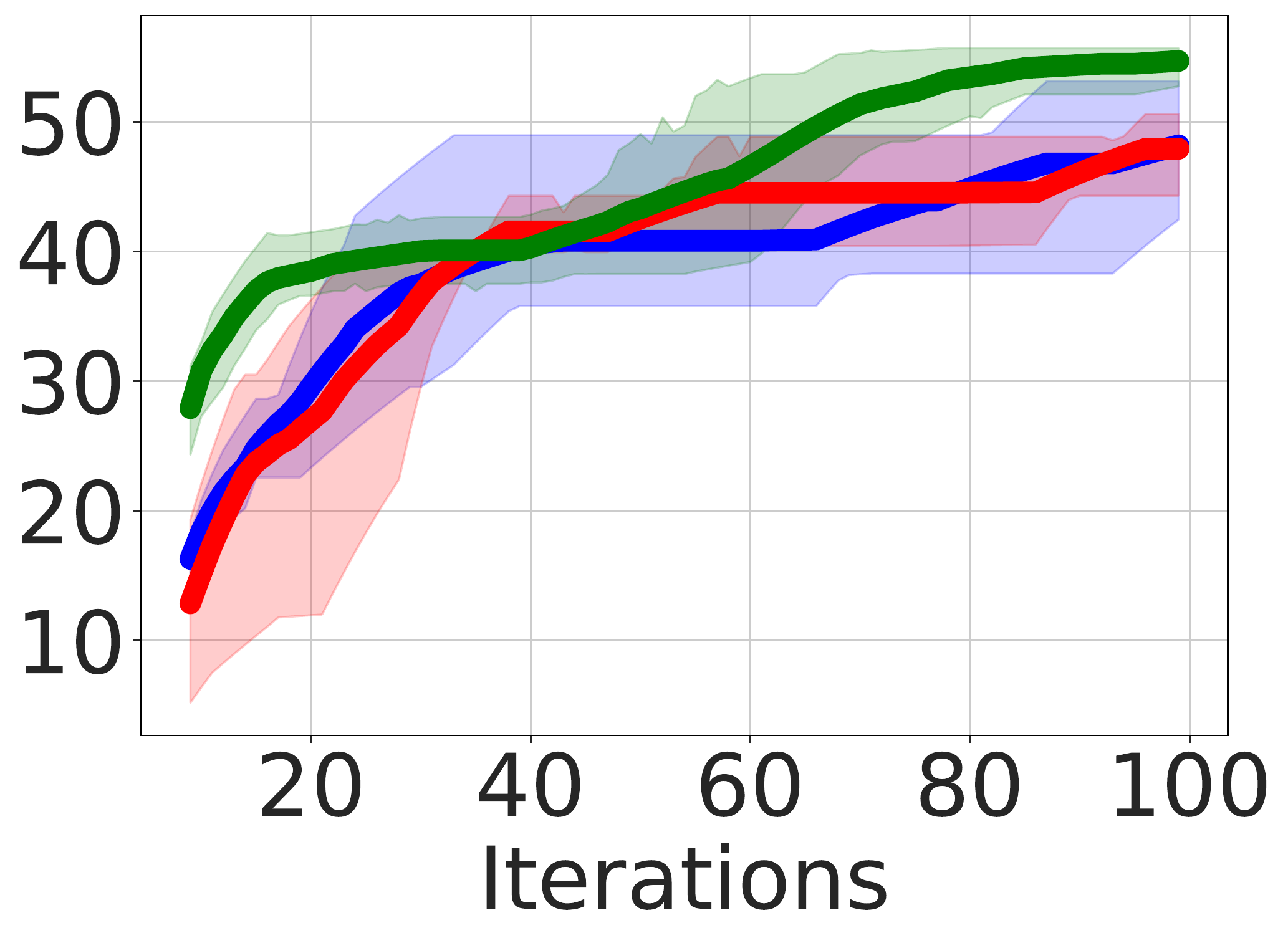} }}}
\vspace{-0.2in}
        \subfloat[m4]{{{\includegraphics[width=0.25\columnwidth, valign=c]{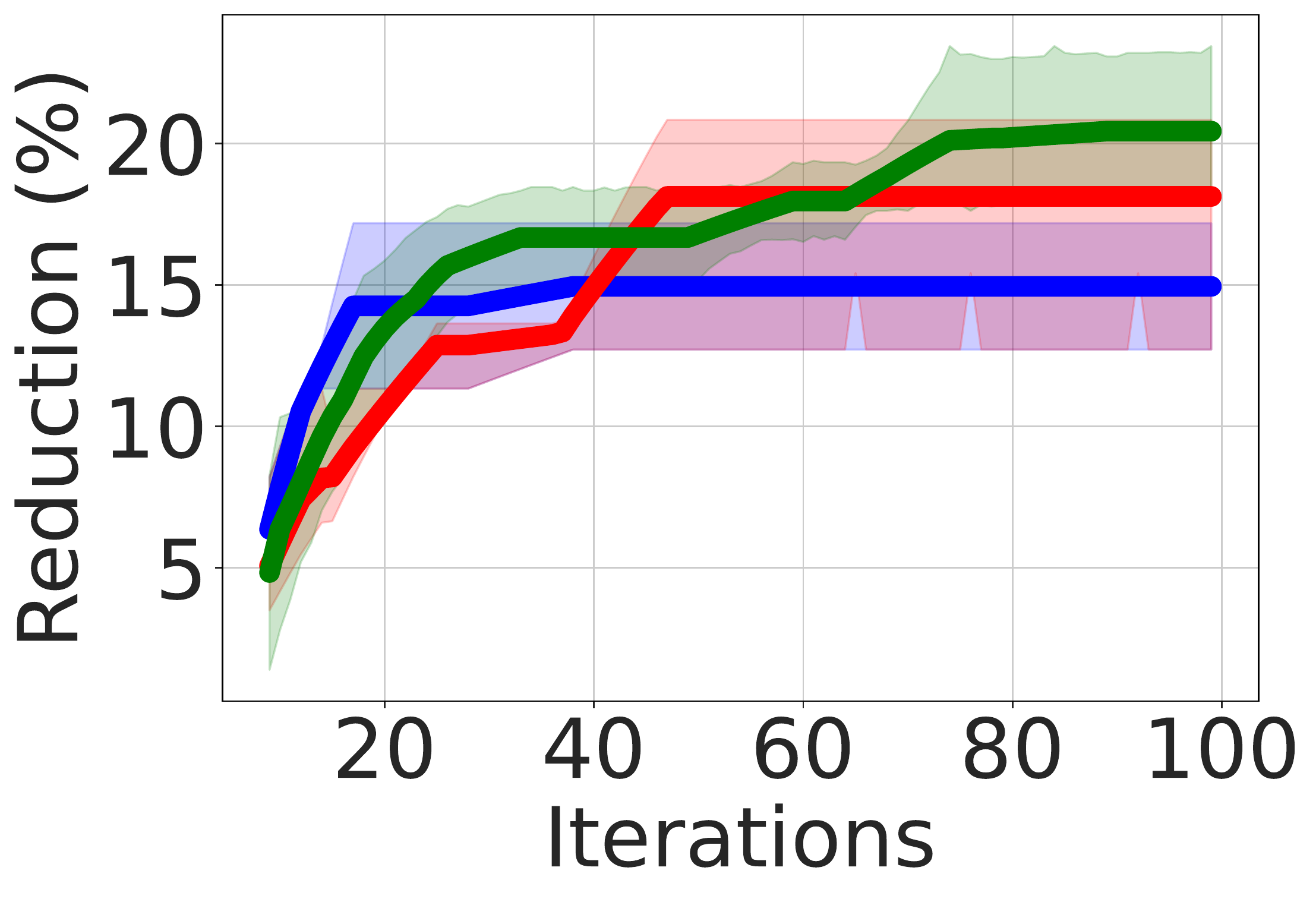} }}
	}
         \subfloat[\color{red}{pair$^*$}]{{{\includegraphics[width=0.25\columnwidth, valign=c]{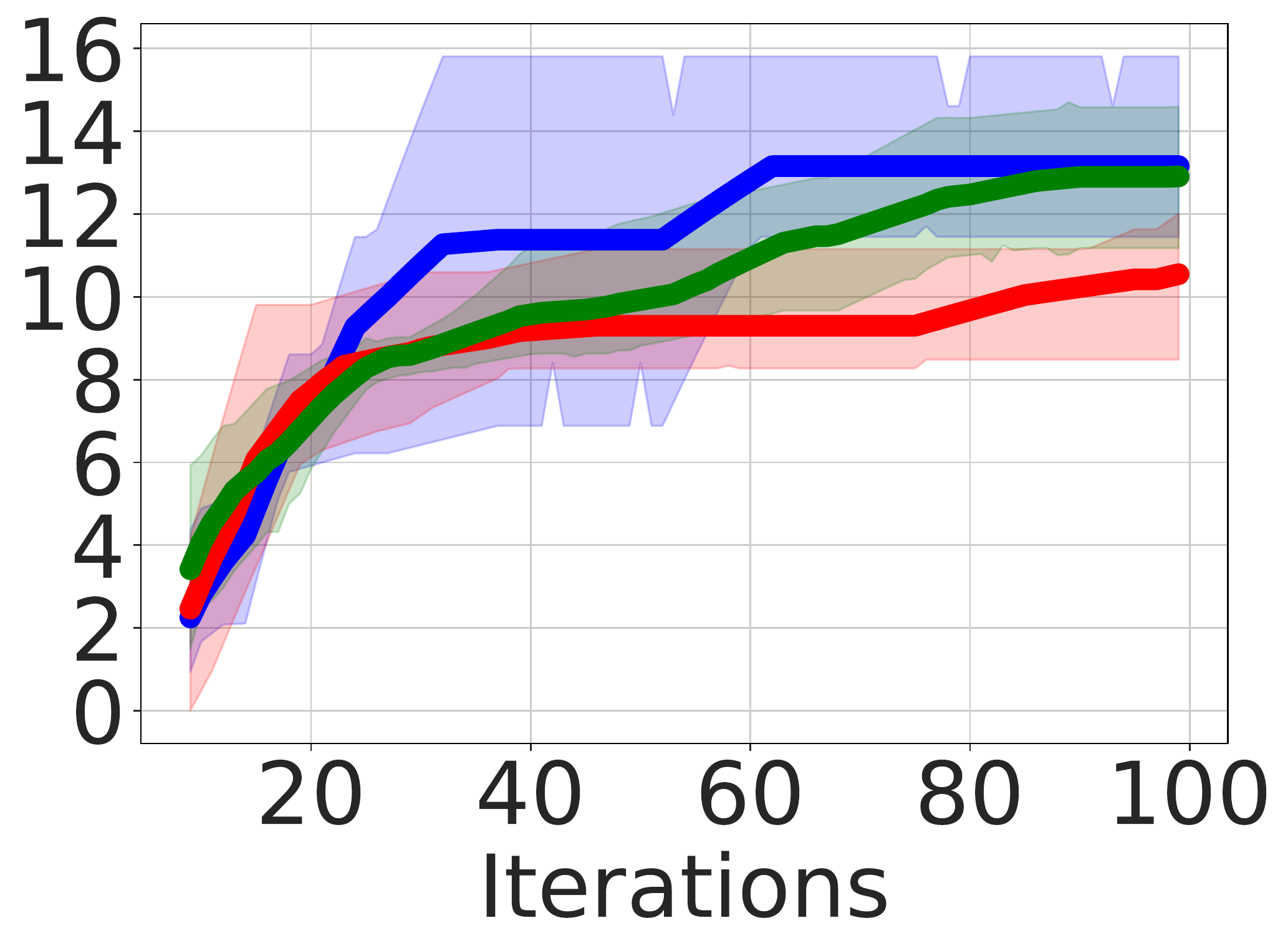} }}
	}
        \subfloat[\color{red}{max1024$^*$}]{{{\includegraphics[width=0.25\columnwidth, valign=c]{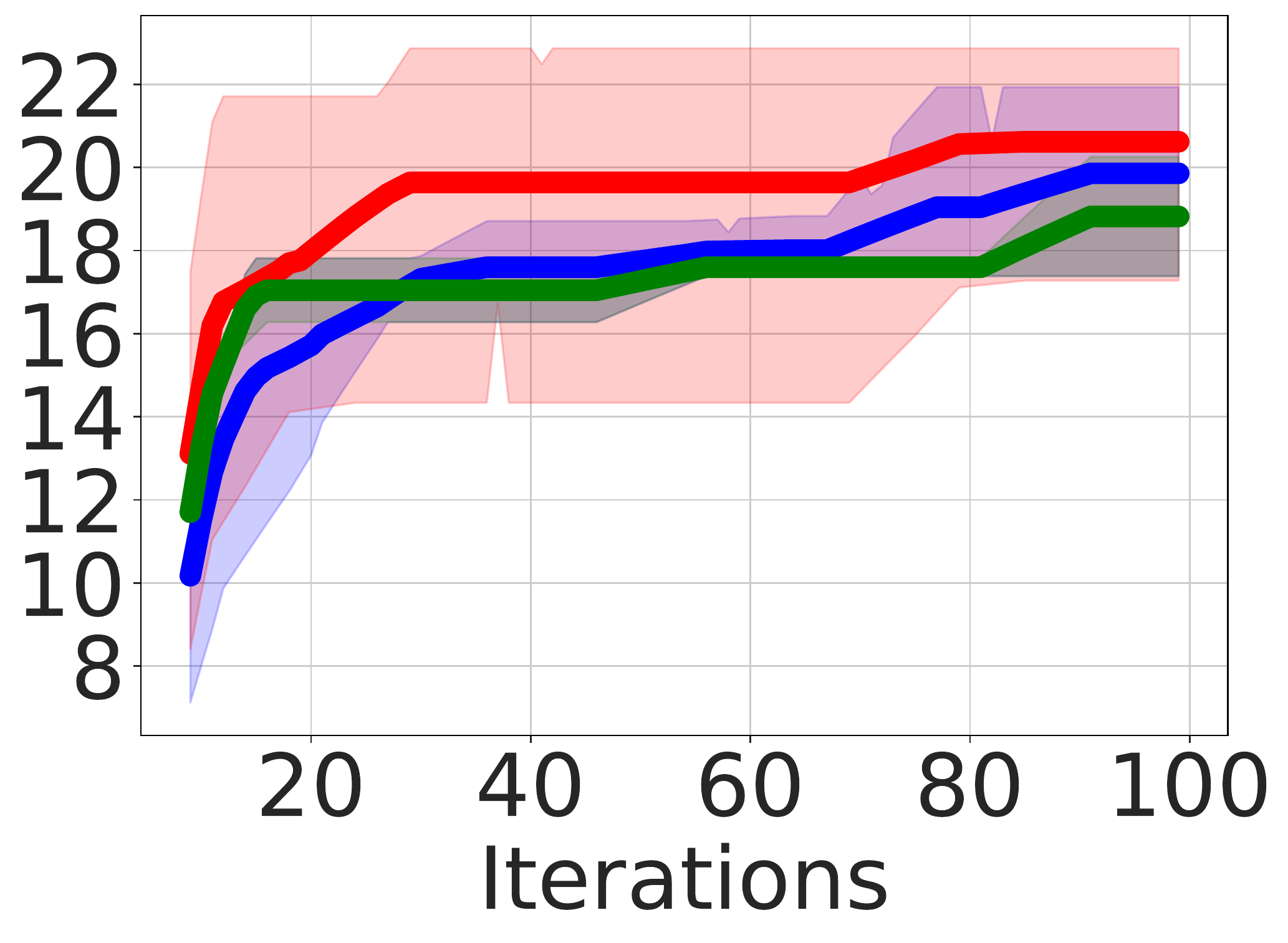} }}
	}
        \subfloat[\color{red}{c7552$^*$}]{{{\includegraphics[width=0.25\columnwidth, valign=c]{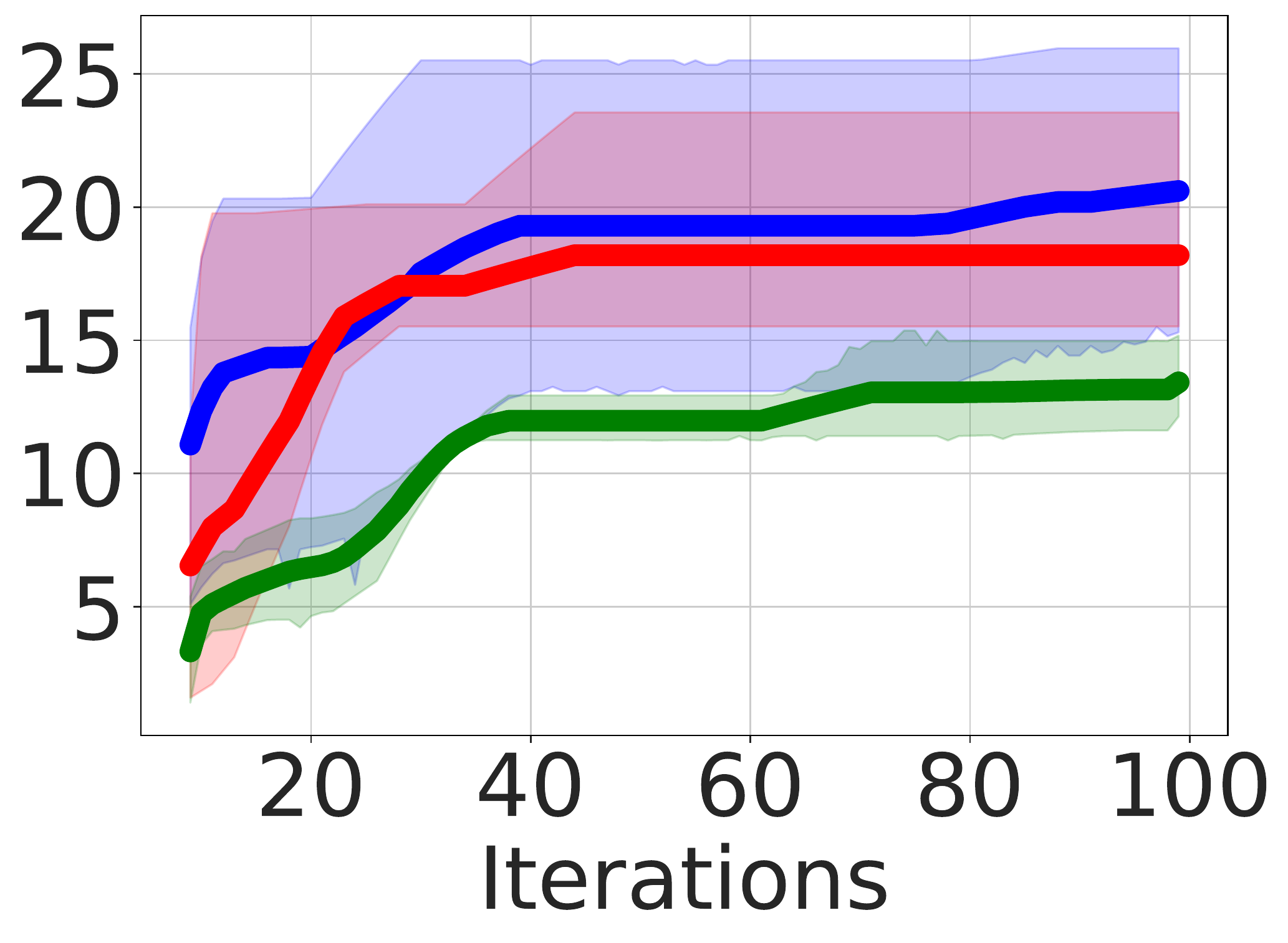} }}
	}
    \caption{Area-delay product reduction (in \%) compared to resyn2 on MCNC circuits. \solution{} classified \texttt{c7552}, \texttt{pair} and \texttt{max1024} as out-of-distribution samples and defaults to pure MCTS search. For rest of the circuits, \solution{} performs pre-trained RL agent-guided search.}
    \label{fig:performance-mcnc}
\end{figure}

We now discuss the performance of \solution{} in reducing area-delay product and improving run-time over state-of-the-art. 

\textbf{MCNC benchmarks:} \autoref{fig:performance-mcnc} demonstrates the effectiveness of \solution{} generated recipes over \ac{SOTA} methods ~\cite{mlcad_abc, neto2022flowtune,bullseye} in terms of percentage \ac{ADP} reduction (all relative to resyn2) and iso-ADP speedup. We report data for \solution{} without ($\alpha=1$) and with our OOD strategy.  Recall that without OOD, \solution{} uses a mix of learning and search. With OOD, however, \solution{} defaults to pure MCTS on OOD inputs. We also report the overall win ratio, i.e., 
the number of circuits on which a particular method achieves the best results. 
The results show that on 10 out of 12 benchmarks, \solution{} generates better recipes than standard MCTS~\cite{neto2022flowtune} and SA with QoR prediction (SA+Pred.)~\cite{bullseye}, with substantial improvements on benchmarks such as \texttt{alu4, apex2, i9, m4, prom1}, and \texttt{c880}. \solution{}'s geo. mean improvements (with OOD) are also the highest overall. Finally, note that \solution{}'s OOD detector correctly defaults to pure search in all three cases where pure MCTS outperforms \solution{}'s history-based RL-guided search.

Figure~\ref{fig:performance-mcnc} plots the ADP reductions over search iterations for MCTS, SA+Pred, and \solution{} ($\alpha=1$, or equivalently without OOD). In \texttt{alu4}, \solution{}'s agent explores paths with higher rewards whereas standard MCTS continues searching without further improvement. A similar trend is observed for \texttt{apex2, m4, prom1} demonstrating that a pre-trained agent helps bias search towards better parts of the search space. SA+Pred.~\cite{bullseye} also leverages past history, but is unable to compete (on average) with MCTS and \solution{}
in part because SA typically underperforms MCTS on tree-based search spaces.


Also note from Figure~\ref{fig:performance-mcnc} that \solution{} in most cases achieves higher ADP reductions earlier than competing methods. This results in significant run-time speedups of $2.5\times$ at iso-\ac{ADP} compared to standard MCTS~\cite{neto2022flowtune}. On in-distribution benchmarks, the speed-up is as high as $6.3\times$; on OOD benchmarks \solution{} performs standard MCTS resulting in  the same speed as prior work.

\begin{wrapfigure}{r}{0.5\textwidth}
\centering
\setlength\tabcolsep{3.5pt}
\resizebox{0.5\textwidth}{!}{%
\begin{tabular}{lrrrrrr}
\toprule
\multirow{3}{*}{Designs}& \multicolumn{5}{c}{ADP reduction (in \%)} & \multirow{3}{*}{\begin{tabular}{@{}c@{}}Iso-\\ADP\\Speed-\\Up\end{tabular}} \\
\cmidrule(lr){2-6}
& \multirow{2}{*}{\begin{tabular}{@{}c@{}}Online-\\RL\\\cite{mlcad_abc}\end{tabular}}& \multirow{2}{*}{\begin{tabular}{@{}c@{}}SA+\\Pred.\\\cite{bullseye}\end{tabular}} & \multirow{2}{*}{\begin{tabular}{@{}c@{}}MCTS\\\cite{neto2022flowtune}\end{tabular}} &
\multicolumn{2}{c}{\solution{}} & \\
\cmidrule(lr){5-6}
& & & & $\alpha=1$ & +OOD & \\
\midrule
adder & \textbf{18.63} & \textbf{18.63} & \textbf{18.63} & \textbf{18.63} &  \textbf{18.63}(\cmark) & 2.2x\\
bar & \textbf{36.89} & \textbf{36.89} & 25.37 & \textbf{36.89} & \textbf{36.89}(\cmark) & 0.6x\\
div & 34.83 & 25.16 & 45.94 & \textbf{55.64} & \textbf{55.64}(\cmark) & 1.4x\\
log2 & 4.73 & 9.58 & 9.09 & \textbf{11.51} & \textbf{11.51}(\cmark) & 1.9x \\
max & 25.09 & 29.87 & 37.50 & \textbf{46.86} & \textbf{46.86}(\cmark) & 1.2x \\
multiplier & 12.41 & 9.75 & 9.90 & \textbf{12.68} & \textbf{12.68}(\cmark) & 1.4x \\
sin & 5.57 & 14.30 & 14.50 & \textbf{15.96} & \textbf{15.96}(\cmark) & 2.4x \\
square & 10.44 & 8.20 & \textbf{12.28} & 9.75  &\textbf{12.28}(\xmark) & 1.0x\\
sqrt & 24.10 & 21.10 & 18.69 & \textbf{24.24} & \textbf{24.24}(\cmark) & 6.3x\\
\midrule
Geomean& 15.41 & 17.01 & 18.42 & 21.51 & \textbf{22.07} & 1.6x\\
Win ratio & 2/9 & 2/9 & 2/9 & 8/9 & 9/9 & 7/9 \\
\bottomrule
\end{tabular}
}
    \captionof{table}{Area-delay reduction over \texttt{resyn2} on EPFL arithmetic benchmarks. \cmark denotes \solution{} perform agent guided search whereas {\xmark} denotes standard MCTS}
\label{tab:qorEPFLarith}
\end{wrapfigure}

\textbf{EPFL arithmetic benchmarks:}

\begin{figure}[h]
\centering
   \subfloat[adder]{{{\includegraphics[width=0.26\columnwidth, valign=c]{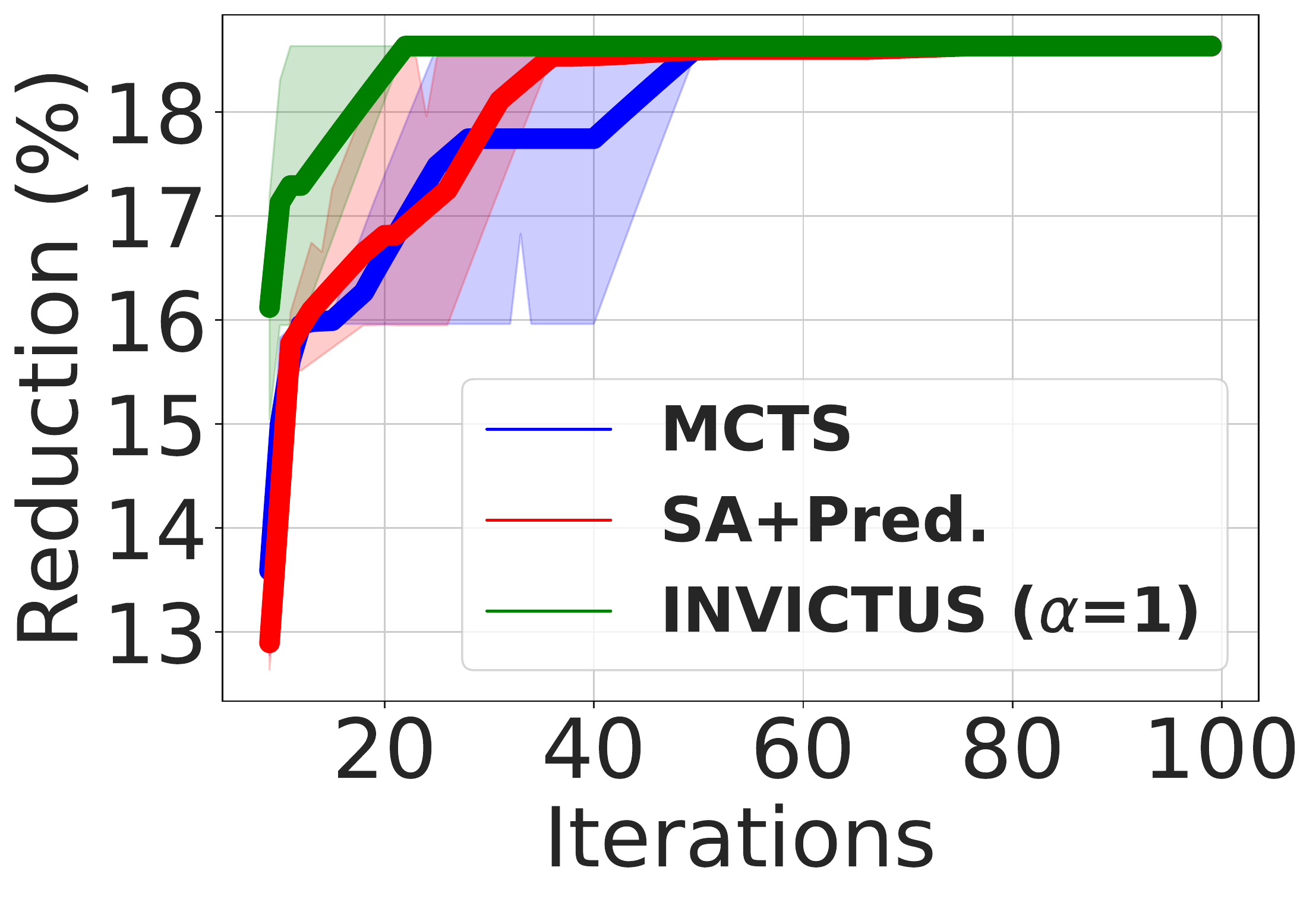} }}%
	}
     \subfloat[sqrt]{{{\includegraphics[width=0.25\columnwidth, valign=c]{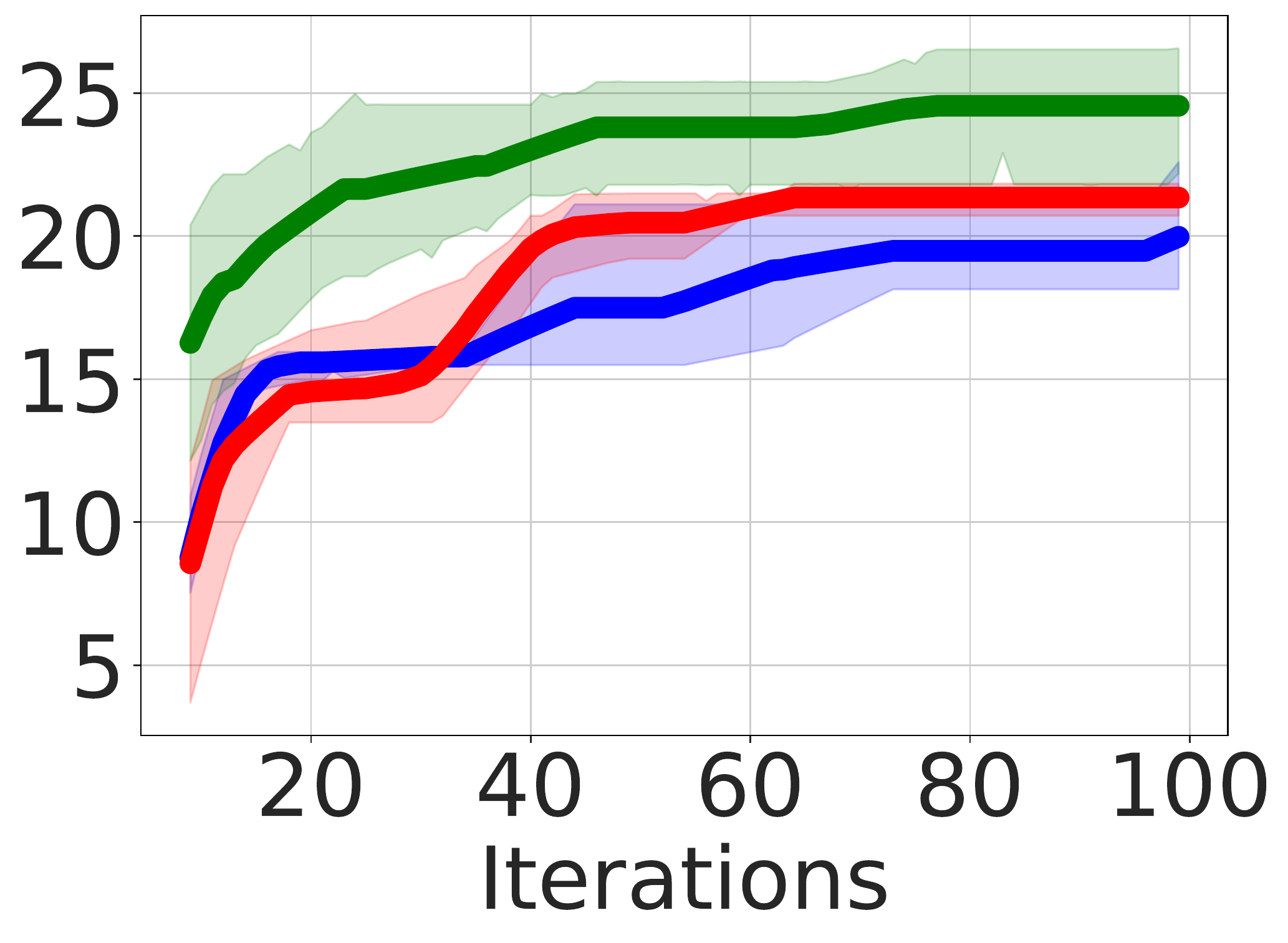} }}
	}
	\subfloat[log2]{{{\includegraphics[width=0.25\columnwidth, valign=c]{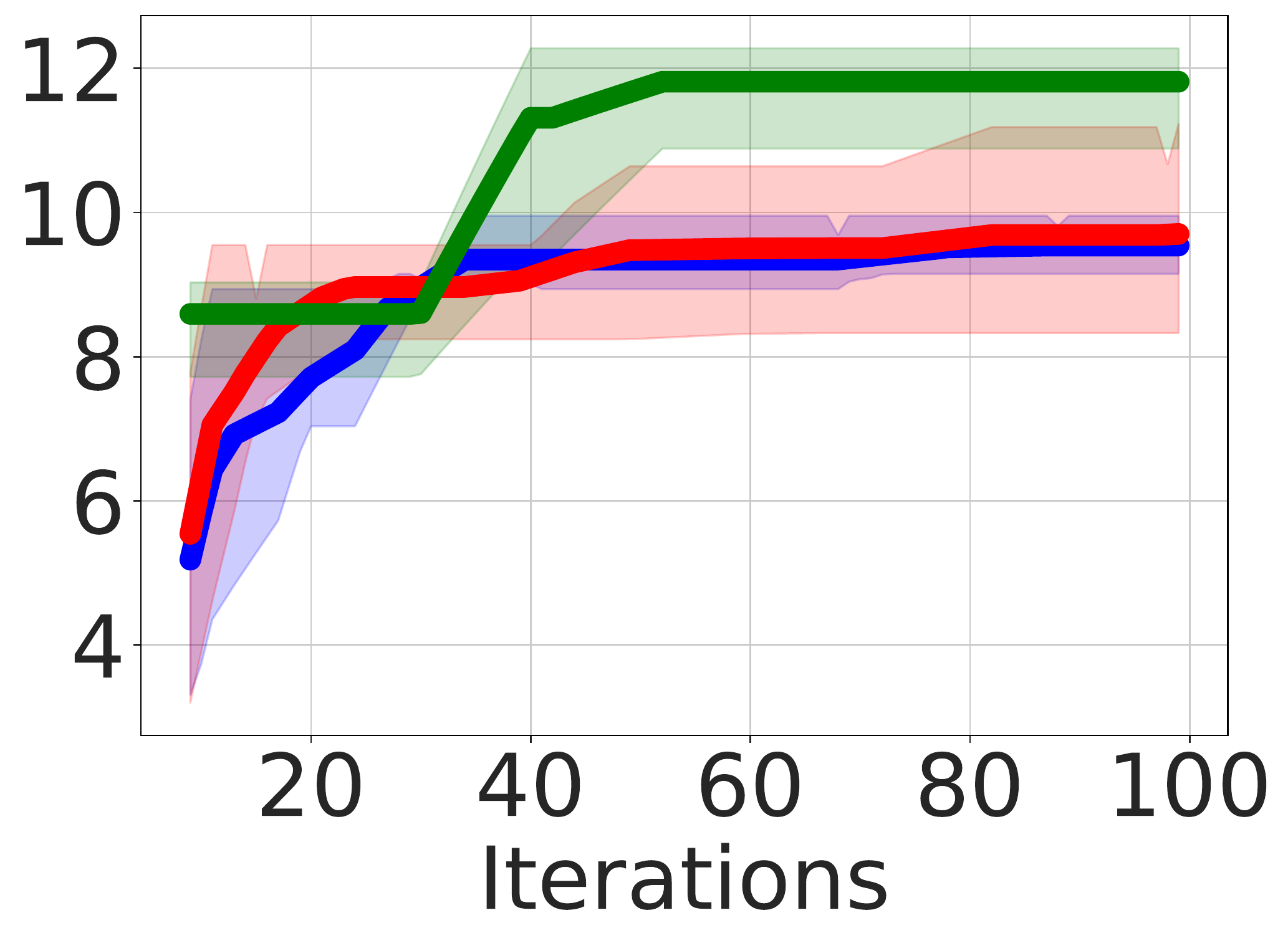} }}
	}
	\subfloat[\color{red}{square$^*$}]{{{\includegraphics[width=0.25\columnwidth, valign=c]{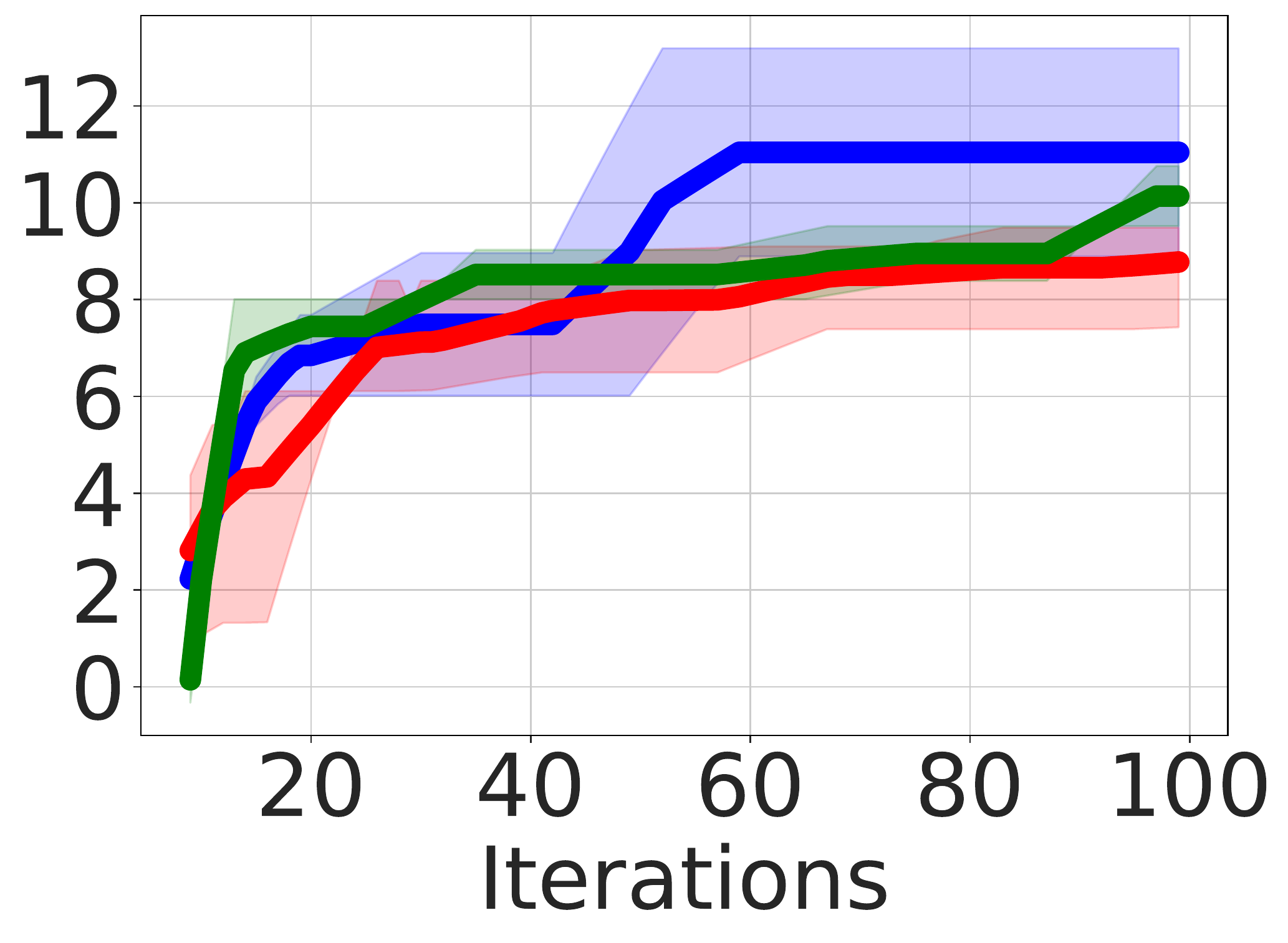} }}
	}
    \caption{Area-delay product reduction (in \%) compared to resyn2 on EPFL arithmetic circuits. \solution{} classified \texttt{square} as out-of-distribution samples and defaults to pure MCTS search. For rest of the circuits, \solution{} performs pre-trained RL agent-guided search.}
    \label{fig:performance-epfl-arith}
\end{figure}

Table~\ref{tab:qorEPFLarith} presents the \ac{ADP} reduction achieved by \solution{} and competing methods on EPFL arithmetic circuits.
On these benchmarks, we correctly classified all except for the \texttt{square} benchmark as in-distribution.
Additionally, \solution{} wins on all nine benchmarks, using OOD detection to match up to pure MCTS on \texttt{square}. Overall, \solution{} achieved a geom. mean \ac{ADP} reduction of 22.07\% over \texttt{resyn2}, representing improvements of $+5.52\%$ and $+6.93\%$ over standard MCTS and SA+Pred., respectively. 
Finally, \solution{} achieves on an average $1.6\times$ runtime speed-up at iso-ADP compared to standard MCTS~\cite{neto2022flowtune}, with up to $6.3\times$ speed-up in the best case (\autoref{fig:performance-epfl-arith}).

\textbf{EPFL random control benchmarks:} 

\begin{wrapfigure}[16]{r}{0.5\textwidth}
\centering
\setlength\tabcolsep{3.5pt}
\resizebox{0.5\textwidth}{!}{%
\begin{tabular}{lrrrrrr}
\toprule
\multirow{3}{*}{Designs}& \multicolumn{5}{c}{ADP reduction (in \%)} & \multirow{3}{*}{\begin{tabular}{@{}c@{}}Iso-\\ADP\\Speed-\\Up\end{tabular}} \\
\cmidrule(lr){2-6}
& \multirow{2}{*}{\begin{tabular}{@{}c@{}}Online-\\RL\\\cite{mlcad_abc}\end{tabular}}& \multirow{2}{*}{\begin{tabular}{@{}c@{}}SA+\\Pred.\\\cite{bullseye}\end{tabular}} & \multirow{2}{*}{\begin{tabular}{@{}c@{}}MCTS\\\cite{neto2022flowtune}\end{tabular}} &
\multicolumn{2}{c}{\solution{}} & \\
\cmidrule(lr){5-6}
& & & & $\alpha=1$ & +OOD & \\
\midrule
arbiter & \textbf{0.03} & \textbf{0.03} & \textbf{0.03} & \textbf{0.03} &  \textbf{0.03}(\cmark) & 2.4x\\
cavlc & 16.28 & \textbf{16.75} & 15.85 & 13.89 & 15.85(\xmark) & 1.0x\\
ctrl & 22.97 & 25.85 & 27.58 & \textbf{30.85} & \textbf{30.85}(\cmark) & 2.6x \\
i2c & 14.91 & 13.10 & 13.45 & \textbf{15.65} & \textbf{15.65}(\cmark) & 3.0x \\
int2float & 7.41 & 7.52 & \textbf{8.10} & 7.52 & \textbf{8.10}(\xmark) & 1.0x \\
mem\_ctrl & 22.54 & 21.45 & 21.55 & \textbf{23.67} & \textbf{23.67}(\cmark) & 2.1x\\
priority & 74.62 & 77.10 & \textbf{77.53} & 75.10 & \textbf{77.53}(\xmark)& 1.0x\\
router & 10.71 & 27.53 & 21.63 & \textbf{25.68}  & 21.63(\xmark) & 1.0x \\
voter & 8.26 & 26.45 & \textbf{27.10} & 26.05 & \textbf{27.10}(\xmark) & 1.0x \\
\midrule
Geomean & 8.29 & 10.47 & 10.38 & 10.69 & \textbf{10.80} & 1.5x \\
Win ratio & 0/9 & 2/9 & 4/9 & 5/9 & 7/9 & 4/9\\
\bottomrule
\end{tabular}
}
    \captionof{table}{Area-delay reduction compared to \texttt{resyn2} on EPFL random-control benchmarks. \cmark denotes \solution{} deploy agent guided search whereas {\xmark} denotes standard MCTS}
\label{tab:qorEPFLrc}
\end{wrapfigure}

Table~\ref{tab:qorEPFLrc} presents the \ac{ADP} reduction achieved by \solution{} and competing methods on EPFL random control circuits. Although our overall conclusions are the same, i.e., \solution{} outperforms competing methods, the trends for this benchmark set are different.
First, \solution{}'s OOD detector is triggered for four of the five benchmark circuits, which is more frequent than for prior benchmark sets.
This aligns with the variability in circuit structure and functionality within the EPFL random control benchmark characterization~\cite{epfl}, causing different synthesis recipe behaviors and train-test distribution shifts.
In the case of \texttt{router}, the OOD detector is triggers incorrectly, i.e., for this benchmark leveraging the pre-trained RL agent would have been helpful.

\solution{}'s pre-trained RL agent is useful for the remaining five benchmarks; outperforming 
standard MCTS~\cite{neto2022flowtune} and simulated annealing with QoR predictor~\cite{bullseye} in each case.  
\solution{} also wins overall on seven of nine benchmarks. \solution{} loses once to SA+Pred. and, interestingly, once to itself without OOD detection.
Finally, \solution{} achieves on an average $1.5\times$ iso-\ac{ADP} runtime speed-up compared to standard MCTS~\cite{neto2022flowtune} with a maximum speed-up of $3\times$.

\section{Discussion and Limitations }
\label{sec:discussion}

We now discuss opportunities to improve \solution{} and its limitations. 

Our results so far indicate that for OOD designs, defaulting to a pure search strategy outperforms RL-guided search. However, this assumes a binary assignment to the $\alpha$ parameter in \autoref{eq:alpha-factor}. Smoothly varying $\alpha$ enables the extent of learning used in search to be varied.
To this end: we modified \autoref{eq:alpha-factor} to a more generalized function as follows:
\begin{align}
    \alpha= 1 - \frac{1}{1+\exp^{-(\frac{\delta^{min}_{test}-\delta_{th}}{T}})}\label{eq:general-alpha-factor}
\end{align}

Here, we introduce temperature $T$ to smoothen our $\alpha$ to real valued function in range $[0,1]$. Setting $T=0$ leads to \autoref{eq:alpha-factor}. We call this soft OOD and tune $T$ on our validation dataset and present our results in Table~\ref{tab:ablation}. Overall, smoothening $\alpha$ yields better geometric mean and win ratio for MCNC and EPFL random control circuits, but falls behind slightly on EPFL arithmetic benchmarks.

\begin{wrapfigure}[9]{r}{0.5\textwidth}
\centering
\setlength\tabcolsep{3.5pt}
\resizebox{0.5\textwidth}{!}{%
\begin{tabular}{lrrrrrr}
\toprule
\multirow{2}{*}{ Settings } & \multicolumn{2}{c}{MCNC} & \multicolumn{2}{c}{EPFL arith} & \multicolumn{2}{c}{EPFL random} \\
\cmidrule(lr){2-3}\cmidrule(lr){4-5}\cmidrule(lr){6-7}
& G.M. & W.R. & G.M & W.R. & G.M. & W.R. \\
\midrule
$\alpha=1$ & 18.84 & 8/12 & 21.46 & 8/9 & 10.78 & 5/9 \\
$T=0$ & 19.76 & 10/12 & 22.07 & 9/9 & 10.80 & 7/9 \\
$T=0.06$ & 20.41 & 11/12 & 21.63 & 8/9 & 11.32 & 7/9 \\
\bottomrule
\end{tabular}
}
\captionof{table}{Varying agent's recommendation during search}
\label{tab:ablation}
\end{wrapfigure}

\begin{figure}[t]
\centering
   \subfloat[pair]{{{\includegraphics[width=0.25\columnwidth, valign=c]{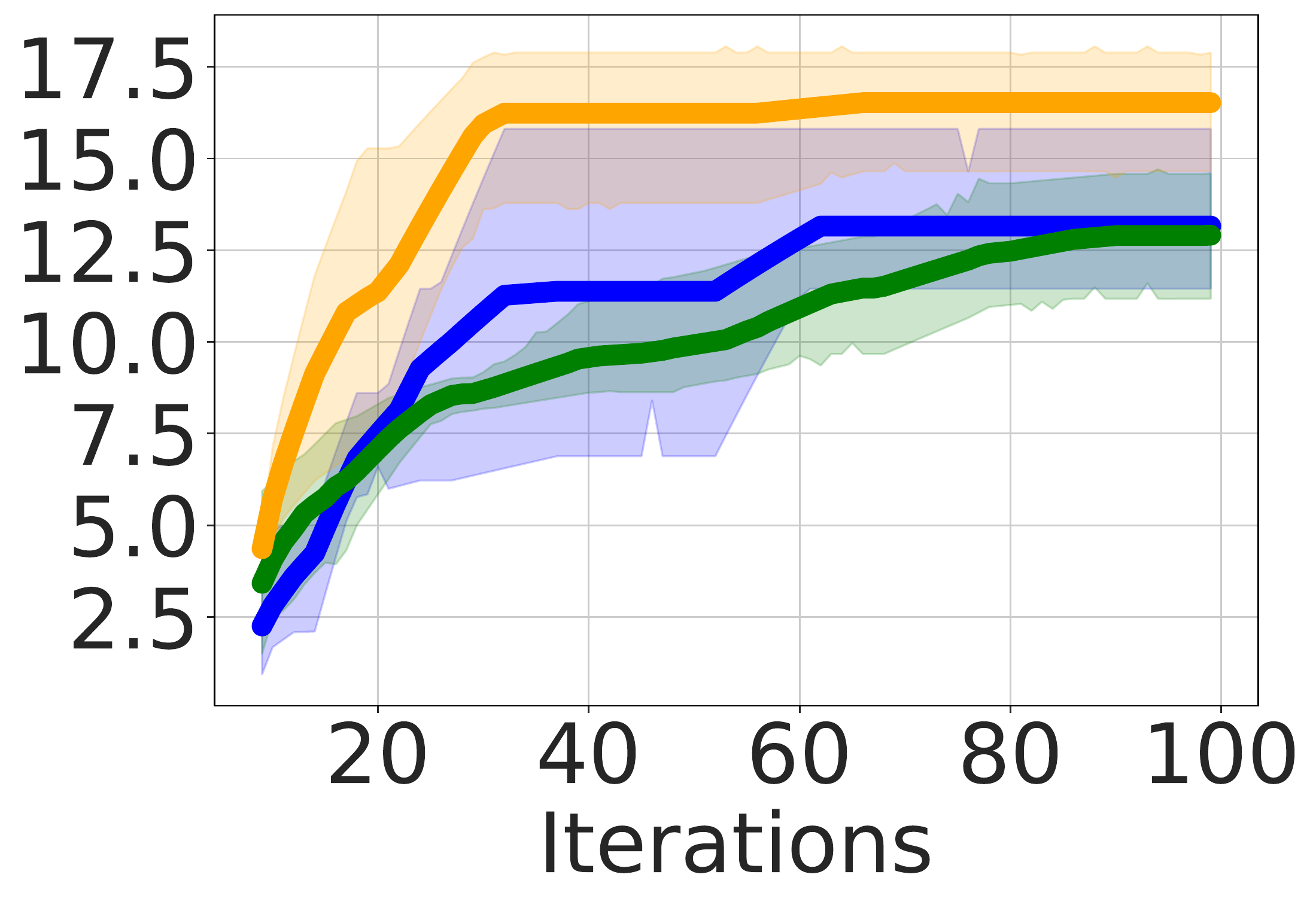} }}%
	}
	\subfloat[max1024]{{{\includegraphics[width=0.25\columnwidth, valign=c]{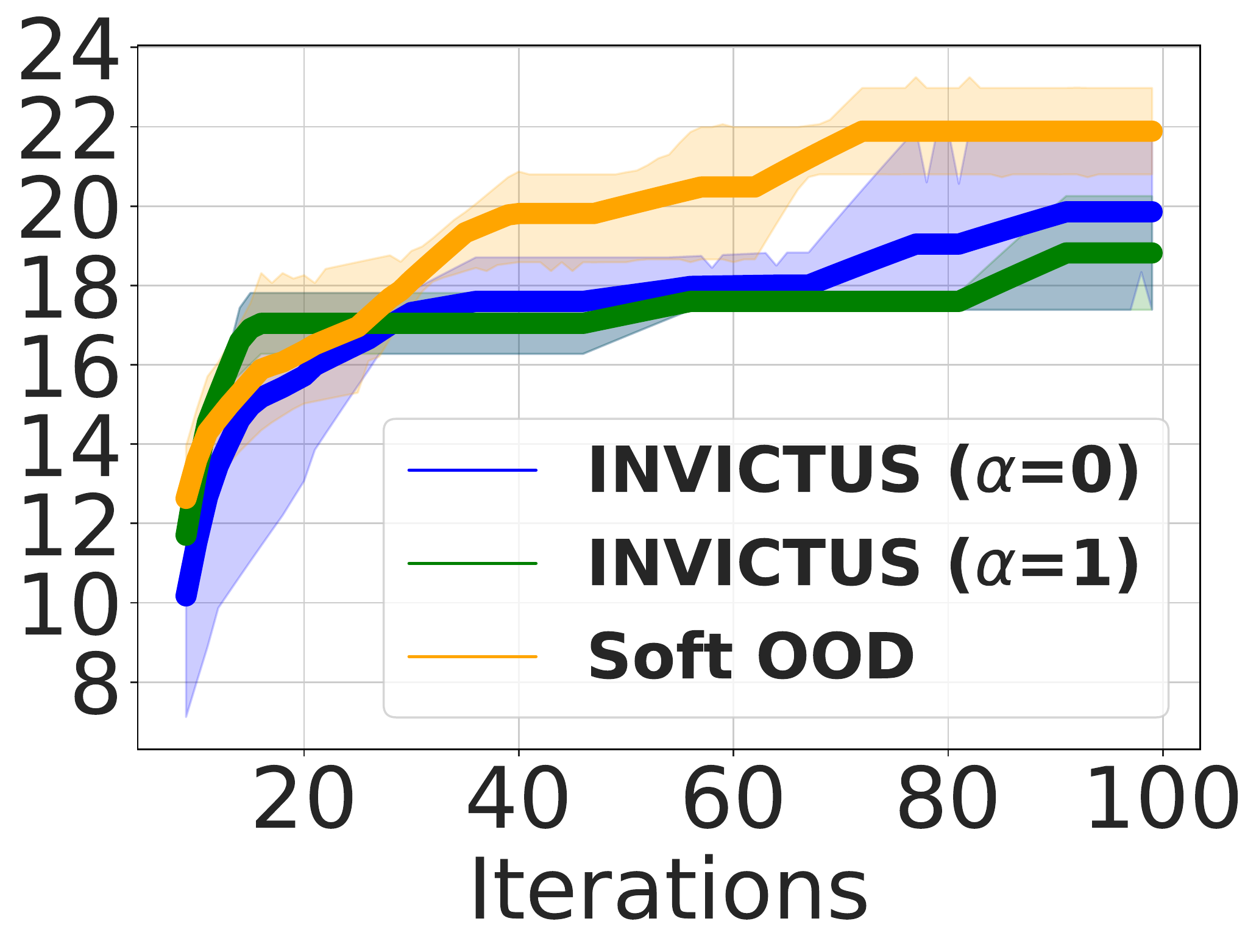} }}
	}
        \subfloat[cavlc]{{{\includegraphics[width=0.25\columnwidth, valign=c]{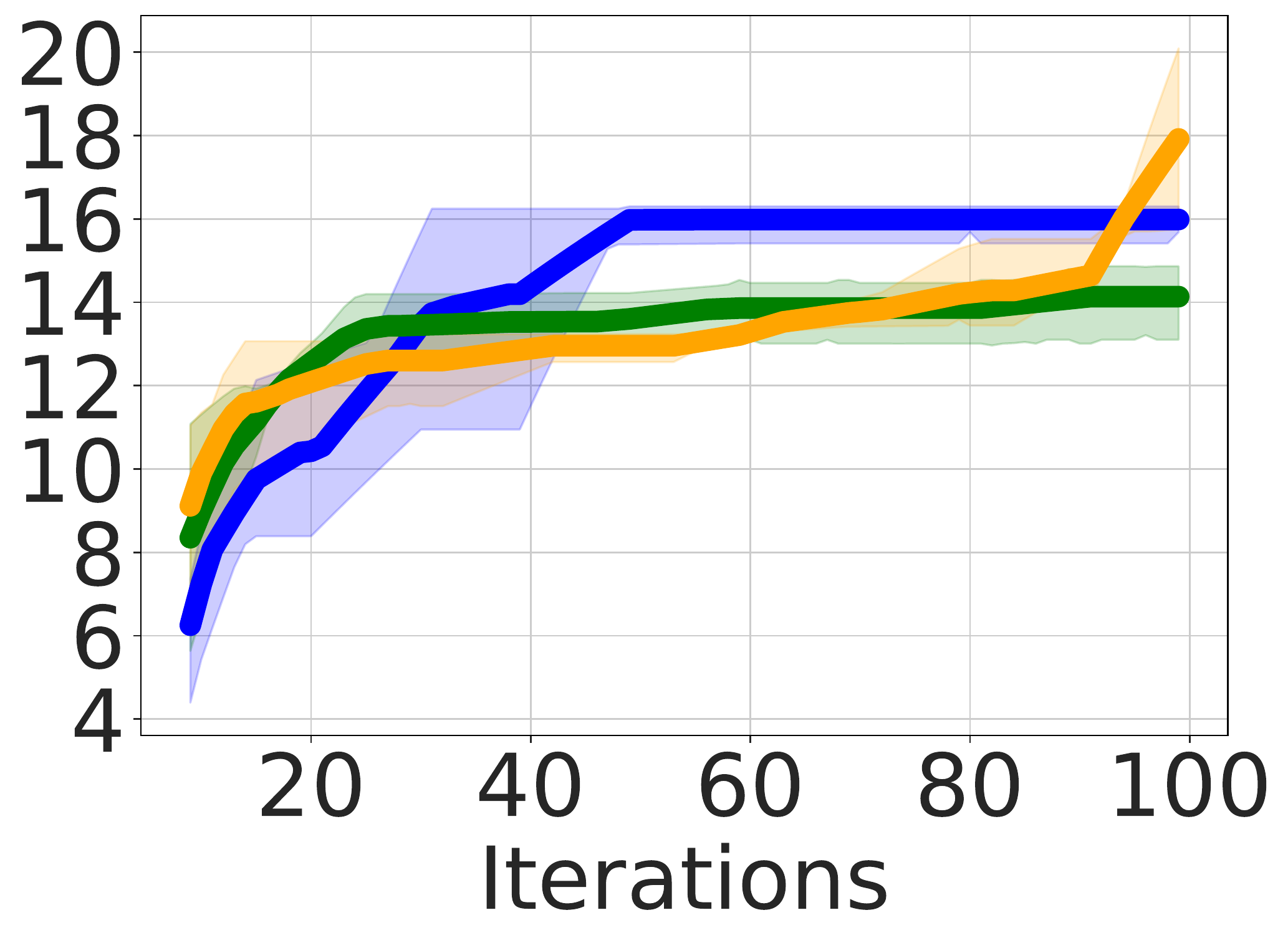} }}
	}
	\subfloat[int2float]{{{\includegraphics[width=0.25\columnwidth, valign=c]{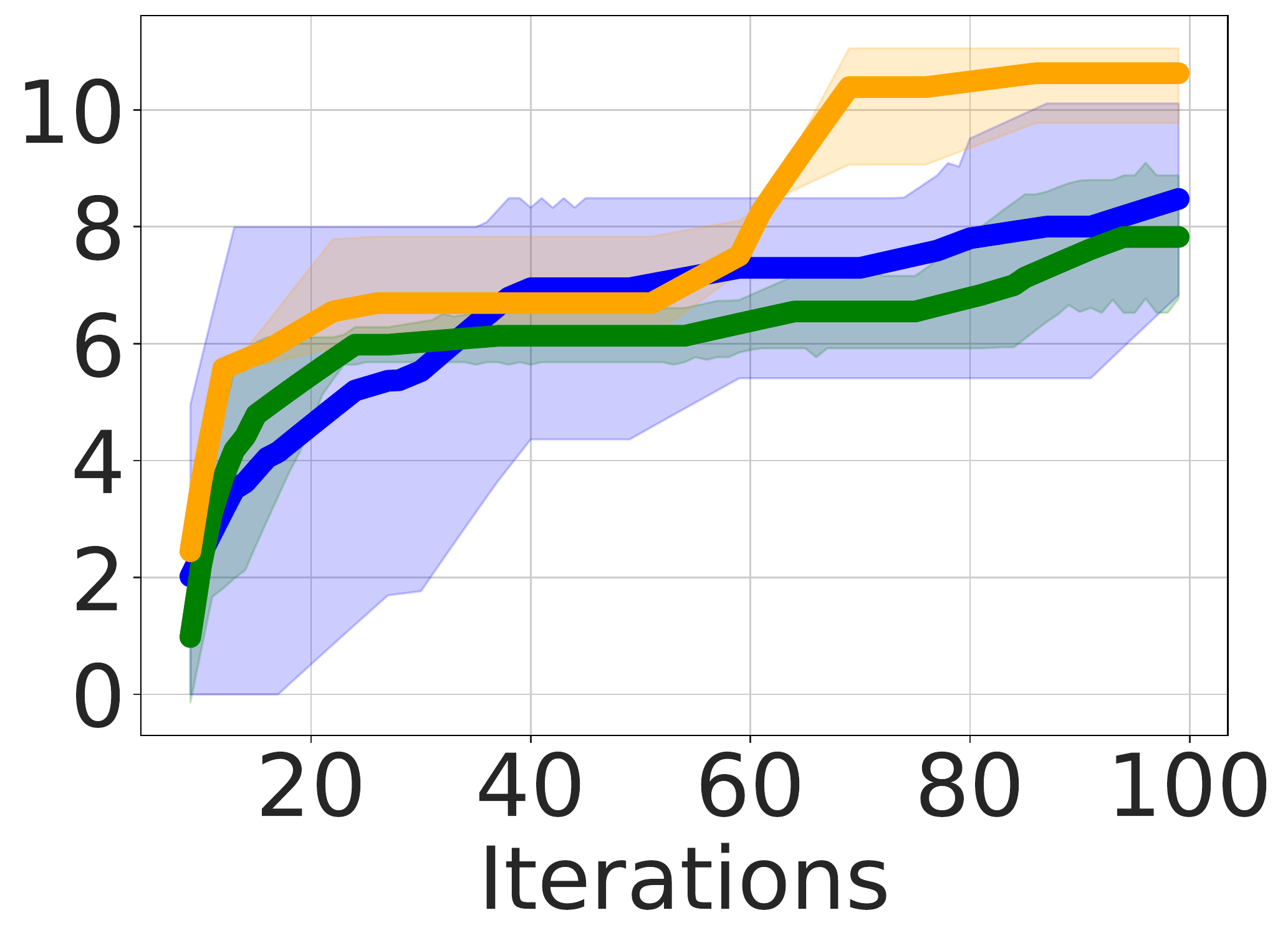} }}
	}
 \caption{Area-delay product reduction (in \%) using soft OOD versus hard OOD on out-of-distribution samples}
    \label{fig:performance-mcnc-synergy}
\end{figure}

\autoref{fig:performance-mcnc-synergy} outlines the performance of smoothing on out-of-distribution benchmarks from MCNC and EPFL random control. We obtain \ac{ADP} reduction upto $4.7\%$ on top of pure MCTS search. This higlights an important insight: a small factor of trained agent's recommendation help bias search towards favourable paths. One strong possibility of such improved performance can be attributed to recipes learned by the agent which broadly works well across different designs. Our future studies include detailed investigation of $\alpha$ smoothing instead of hard \ac{OOD} based decision making and its explainability.
We believe synergistically combining learning and search is going to help yield better results under time-to-market pressure in \ac{EDA} industry. 

\textbf{Limitations}: \solution{} also has some limitations. First, large, standardized, open-source datasets are scarce in the hardware community. Our results suggest that RL agents can be effectively trained on few training circuits (although note that for each circuit we generate thousands of synthesis runs), but evaluations on larger scale datasets will be illuminating. To that point, \solution{}'s training is costly, running into a week for our benchmarks. Interestingly, the bottleneck is the cost of repeatedly running logic synthesis during training; mitigating training time is an area of future improvement. Finally, \solution{} does not address the challenge of continually updating the learned policy agent as new designs are seen.

\section{Related work}
\label{sec:relatedWork}

Prior work in logic synthesis can be classified into three categories: expert-crafted synthesis recipes~\cite{yang2012lazy,dag_mischenko,riener2019scalable}, classical optimization approaches~\cite{cunxi_iccad,neto2022flowtune} and learning-guided approaches~\cite{firstWorkDL_synth,lsoracle,cunxi_dac,drills,mlcad_abc,bullseye}. 
In the first category, researchers have studied the transformations performed by logic minimization heuristics on a variety of circuit benchmarks and devised ``good" synthesis recipes (e.g. \texttt{resyn2}~\cite{dag_mischenko}) which tend to perform reasonably on a wide range of benchmarks. We show that to get good results, synthesis recipes must be tailored to the design.
Classical optimization techniques~\cite{cunxi_iccad,neto2022flowtune} formulate the logic synthesis problem as a black-box optimization problem and leverage heuristics such as MCTS to generate design-specific synthesis recipes. 
Compared to these methods, we show that learning from previous designs can improve quality and run-time.

Learning-based approaches can further be classified into two sub-categories: 1) Synthesis recipe classification~\cite{cunxi_dac,lsoracle} and prediction~\cite{openabc,bullseye} based approaches, and 2) \ac{RL}-based approaches~\cite{firstWorkDL_synth,drills,mlcad_abc}. In ~\cite{cunxi_dac}, the authors train a CNN classifier to classify an unseen synthesis recipe as ``angel'' or ``devil'' recipe using a \ac{QoR} labeled dataset generated by synthesizing the design. \cite{lsoracle} partition the original graph into smaller sub-networks and performs binary classification on sub-networks to pick which recipes work best. However, these methods only work over a small number of pre-selected recipes and have been outperformed by MCTS search methods.
\cite{bullseye,openabc} learns a \ac{QoR} predictor model for a given \ac{AIG} and synthesis recipe using a massive synthesis dataset and use it with simulated annealing for larger search space exploration within a specified time budget. On the other hand, \ac{RL}-based solutions~\cite{firstWorkDL_synth,drills,mlcad_abc} use online \ac{RL} algorithms to craft synthesis recipes, but do not leverage any prior data. We show that \solution{} outperforms these methods.

Finally, ML has been deployed for a range of other EDA problems as well~\cite{googlerl,kurin2020can,lai2022maskplace,schmitt2021neural,yolcu2019learning,vasudevan2021learning,yang2022versatile,cheng2023assessment}. Closer to this work,
\cite{googlerl} trains a deep-RL agent to optimize chip floorplanning, a different EDA problem that seeks to place modules of a synthesized design on the surface of a chip.
Compared to logic synthesis, each action in floorplanning,  i.e., moving the x-y coordinates of modules in the design, is computationally inexpensive, unlike the time-consuming actions in logic synthesis. Thus, a large number of moves can be simulated even on a new design, enabling the trained agent to be fine-tuned. \solution{} uses a different strategy, synergistically combining learning and search with an out-of-distribution detector. 



\section{Conclusion}
\label{sec:conclusion}

We propose \solution{}, a novel approach that combines learning, search, and out-of-distribution (OOD) detection that greatly improves the process of identifying high-quality synthesis recipes for new hardware designs. Particularly, the integrated use of a pre-trained \ac{RL} agent, an RL agent-guided MCTS search over the synthesis recipe space, and an \ac{OOD} selection process between the learned policy and pure search has proved to be effective. These keys ideas backed by empirical results highlight the potential of \solution{} to generate high-quality synthesis recipes making modern complex chips design more efficient and cost-effective.



\small
\bibliographystyle{unsrt}
\bibliography{main}

\newpage
\section{Reproduciblity}

We provide our detailed experimental setup and model hyperparameters in \autoref{sec:empiricalEval}. We have shared the codes for training our model at the following URL: \url{https://anonymous.4open.science/r/invictus-83A3/}. Post acceptance of our work, we will publicly release it with detailed user's instruction.

\section{Broader Impact}

Our work \solution{} attempts to solve one of the main combinatorial optimization problem in the field of \acf{EDA}. We forsee our work to be adopted similarly in other optimization problems such as floorplanning, macro placement, routing and design rule checks. The ideas may nurture promising growth of \acf{ML} in chip design automation: which is regarded is one of the most complex optimization problems and still rely a lot on experienced engineers and practitioners' input to achieve better \acf{QoR}.
\appendix
\section{Appendix 1}

\subsection{Logic Synthesis \label{sec:logic-synth}}

Logic synthesis transforms a hardware design in \acf{RTL} to a Boolean gate-level network, optimizes the number of gates/depth, and then maps it to standard cells in a technology library~\cite{brayton1984logic}. Well-known representations of Boolean networks include sum-of-product form, product-of-sum, Binary decision diagrams, and \acp{AIG} which are a widely accepted format using only AND (nodes) and NOT gates (dotted edges). Several logic minimization heuristics (discussed in \autoref{subsec:ls_heuristics})) have been developed to perform optimization on \ac{AIG} graphs because of its compact circuit representation and \ac{DAG}-based structuring. These heuristics are applied sequentially (``synthesis recipe'') to perform one-pass logic optimization reducing the number of nodes and depth of \ac{AIG}. The optimized network is then mapped using cells from technology library to finally report area, delay and power consumption. 

\subsection{Logic minimization heuristics}
\label{subsec:ls_heuristics}

We now describe optimization heuristics provided by industrial strength academic tool ABC~\cite{abc}:

\noindent\textbf{1. Balance (b)} optimizes \ac{AIG} depth by applying associative and commutative logic function tree-balancing transformations to optimize for delay.

\noindent\textbf{2. Rewrite (rw, rw -z)} is a \acf{DAG}-aware logic rewriting technique that performs template pattern matching on sub-trees and encodes them with equivalent logic functions. 

\noindent\textbf{3. Refactor (rf, rf -z)} performs aggressive changes to the netlist without caring about logic sharing. It iteratively examines all nodes in the \ac{AIG}, lists out the maximum fan-out-free cones, and replaces them with equivalent functions when it improves the cost (e.g., reduces the number of nodes).


\noindent\textbf{4. Re-substitution (rs, rs -z)} creates new nodes in the circuit representing intermediate functionalities using existing nodes; and remove redundant nodes. Re-substitution improves logic sharing. 


The zero-cost (-z) variants of these transformation heuristics perform structural changes to the netlist without reducing nodes or depth of \ac{AIG}. However, previous empirical results show circuit transformations help future passes of other logic minimization heuristics reduce the nodes/depth and achieve the minimization objective.

\subsection{Monte Carlo Tree Search}
\label{subsec:baselineMCTS}

We discuss in detail the \ac{MCTS} algorithm.
During selection, a search tree is built from the current state by following a search policy, with the aim of identifying promising states for exploration.

where $Q_{MCTS}^{k}(s,a)$ denotes estimated $Q$ value (discussed next) obtained after taking action $a$ from state $s$ during the $k^{th}$ iteration of \ac{MCTS} simulation. $U_{MCTS}^{k}(s,a)$ represents upper confidence tree (UCT) exploration factor of \ac{MCTS} search.

\begin{equation}
    U_{MCTS}^k(s, a) = c_\mathrm{UCT}\sqrt{\frac{\log\left(\sum_a N_{MCTS}^k(s, a)\right)}{N_{MCTS}^k(s, a)}},
\end{equation}

$N_{MCTS}^k(s,a)$ denotes the visit count of the resulting state after taking action $a$ from state $s$. $c_{UCT}$ denotes a constant exploration factor~\cite{kocsis2006bandit}.

The selection phase repeats until a leaf node is reached in the search tree. 
A leaf node in MCTS tree denotes either no child nodes have been created or it is a terminal state of the environment.
Once a leaf node is reached the expansion phase begins where an action is picked randomly and its roll out value is returned or $R(s_L)$ is returned for the terminal state $s_L$. Next, back propagation happens where all parent nodes $Q_k(s,a)$ values are updated according to the following equation. 
\begin{equation}
    Q_{MCTS}^k(s, a)=\sum_{i=1}^{N_{MCTS}^k(s,a)} R_{MCTS}^i(s, a) / N_{MCTS}^k(s, a).
    \label{eq:backup}
\end{equation}


\subsection{\solution{} Agent Pre-Training Process}
As discussed in~\autoref{sec:Training-Prior}, we pre-train an agent using available past data to help with choosing which logic minimization heuristic to add to the synthesis recipe. The process is shown as~\autoref{alg:qi}. 

\begin{figure*}[t]
    \centering
\includegraphics[width=\columnwidth]{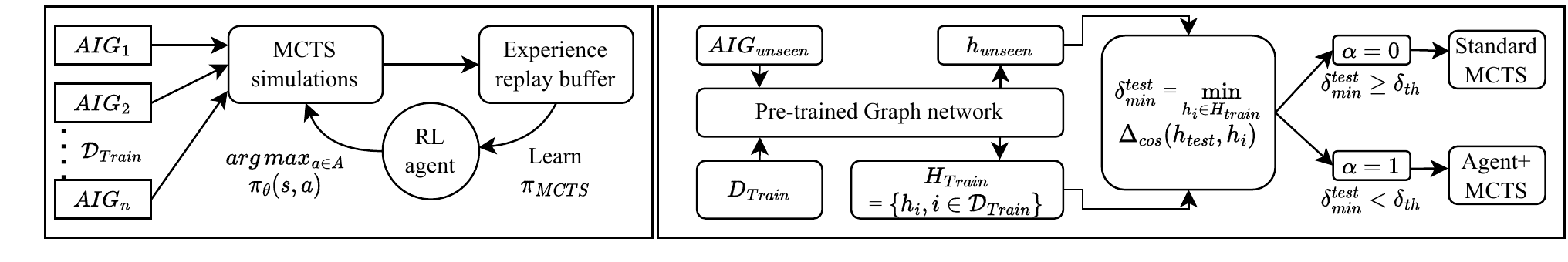}
    \caption{\solution{} flow: Training the agent (left) and Recipe generation at inference-time (right)}
    \label{fig:invictusFlow}
\end{figure*}

\begin{algorithm}[h]
\caption{Invictus: Policy agent pre-training}
\label{alg:qi}
\begin{algorithmic}[1]
\Procedure{Training}{$\theta$}
    \State Replay buffer $(RB) \leftarrow \phi$, $\mathcal{D}_{train} = \{AIG_1,AIG_2,...,AIG_n\}$, num\_epochs=$N$, Recipe length=$L$, AIG embedding network: $\Lambda$, Recipe embedding network: $\mathcal{R}$, Agent policy $\pi_\theta:= U$ (Uniform distribution), MCTS iterations = $K$, Action space = $A$
    \For{$AIG_i \in \mathcal{D}_{train}$}
    \State $r \leftarrow 0$, $depth \leftarrow 0$
    \State $s \leftarrow \Lambda(AIG_i)+\mathcal{R}(r)$
    \While{depth < $L$}
        \State $\pi_{MCTS} = MCTS(s,\pi_\theta,K)$
        \State $a = argmax_{a' \in A}\pi_{MCTS}(s,a')$
        \State $r \leftarrow r+a$, $s' \leftarrow \mathcal{A}(AIG_i)+\mathcal{R}(r)$ 
        \State $RB \leftarrow RB \bigcup (s,a,s',\pi_{MCTS}(s,\cdot))$
        \State $s \leftarrow s', depth \leftarrow depth+1$
    \EndWhile
    \EndFor
    \For{epochs < $N$}
        \State $\theta \leftarrow \theta_i - \alpha\nabla_{\theta}\mathcal{L}(\pi_{MCTS},\pi_{\theta})$
    \EndFor
\EndProcedure
\end{algorithmic}
\end{algorithm}

\section{Network architecture}

\subsection{AIG Network architecture \label{subsec:aigNetwork}}
Starting with a graph G = ($\mathbf{V},\mathbf{E}$) that has vertices $\mathbf{V}$ and edges $\mathbf{E}$, the \ac{GCN} aggregates feature information of a node with its neighbors' node information. The output is then normalized using \texttt{Batchnorm} and passed through a non-linear \texttt{LeakyReLU} activation function. This process is repeated for k layers to obtain information for each node based on information from its neighbours up to a distance of k-hops. A graph-level READOUT operation produces a graph-level embedding. Formally:
\begin{gather}\label{eq:basic}
\tiny
    h^k_u=\sigma ( W_k\sum_{i \in u \cup N(u)} \frac{h^{k-1}_i}{\sqrt{N(u)}\times \sqrt{N(v)}} + b_{k} ) , k \in [1..K] \\
    h_{G}=READOUT(\{h^k_u; u \in V\}) \nonumber
\end{gather}

The embedding for node $u$, generated by the $k^{th}$ layer of the \ac{GCN}, is represented by $h^k_u$. The parameters $W_k$ and $b_k$ are trainable, and $\sigma$ is a non-linear ReLU activation function. $N(\cdot)$  denotes the 1-hop neighbors of a node. The READOUT function combines the activations from the $k^{th}$ layer of all nodes to produce the final output by performing a pooling operation. In our work, we choose $k=2$ and global average and max pooling concatenated as READOUT operation.

\section{Experimental details}
\subsection{Reward normalization}
\label{subsec:qorRewardNorm}
In our work, maximizing \ac{QoR} entails finding a recipe $P$ which is minimizing the area-delay product of transformed \ac{AIG} graph. We consider as a baseline recipe an expert-crafted synthesis recipe \texttt{resyn2}~\cite{dag_mischenko} on top of which we improve our \ac{ADP}. 

\begin{equation*}
R = \begin{cases}
1 - \frac{ADP(\mathcal{S}(G,P))}{ADP(\mathcal{S}(G,resyn2))} & \quad  ADP(\mathcal{S}(G,P)) < 2\times ADP(\mathcal{S}(G,P)), \\
-1 & \quad otherwise.
\end{cases}
\end{equation*}

\section{Results}

\subsection{EPFL arithmetic benchmarks \label{subsec:result-epfl-arith}}

We used agent-based search to evaluate the \ac{ADP} reduction on EPFL arithmetic benchmarks. We treated each circuit as test data and evaluated it with a corresponding \ac{RL} agent. ~\autoref{fig:performance-epfl-arith} shows that, except for the \texttt{square} benchmark, agent-guided search minimized \ac{ADP} more effectively than both standard MCTS~\cite{neto2022flowtune} and simulated annealing~\cite{bullseye}. This performance suggests that the pre-training circuits share similar characteristics with the test data circuits.

\begin{figure}[h]
\centering
   \subfloat[adder]{{{\includegraphics[width=0.25\columnwidth, valign=c]{figures/2023_neurips_plot_adder_compare_ADP_best_adder.pdf} }}%
	}
	\subfloat[bar]{{{\includegraphics[width=0.25\columnwidth, valign=c]{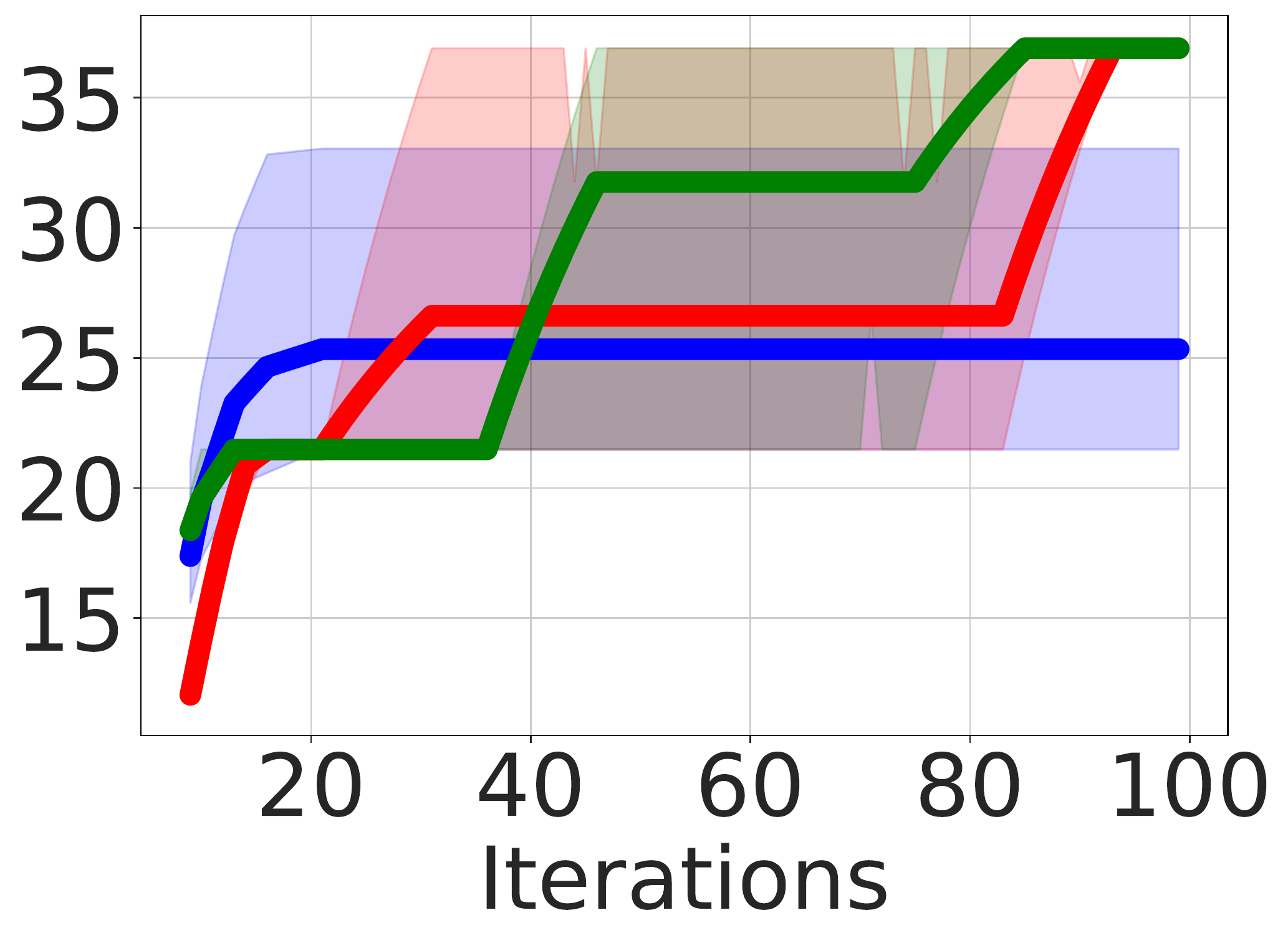} }}
	}
        \subfloat[div]{{{\includegraphics[width=0.25\columnwidth, valign=c]{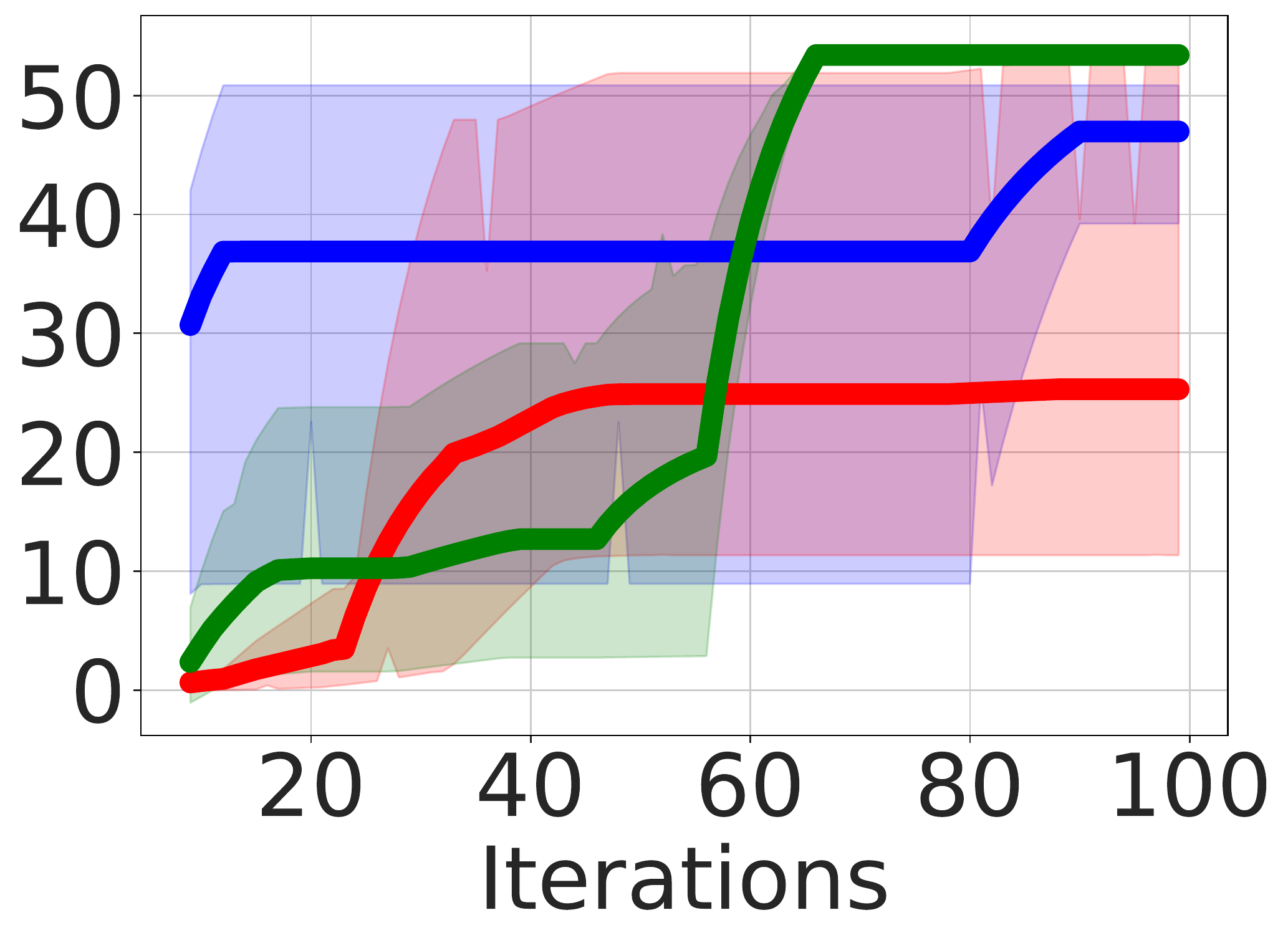} }}
	}
	\subfloat[log2]{{{\includegraphics[width=0.25\columnwidth, valign=c]{figures/2023_neurips_plot_log2_compare_ADP_best_log2.pdf} }}
	}
	
	\subfloat[max]{{{\includegraphics[width=0.25\columnwidth, valign=c]{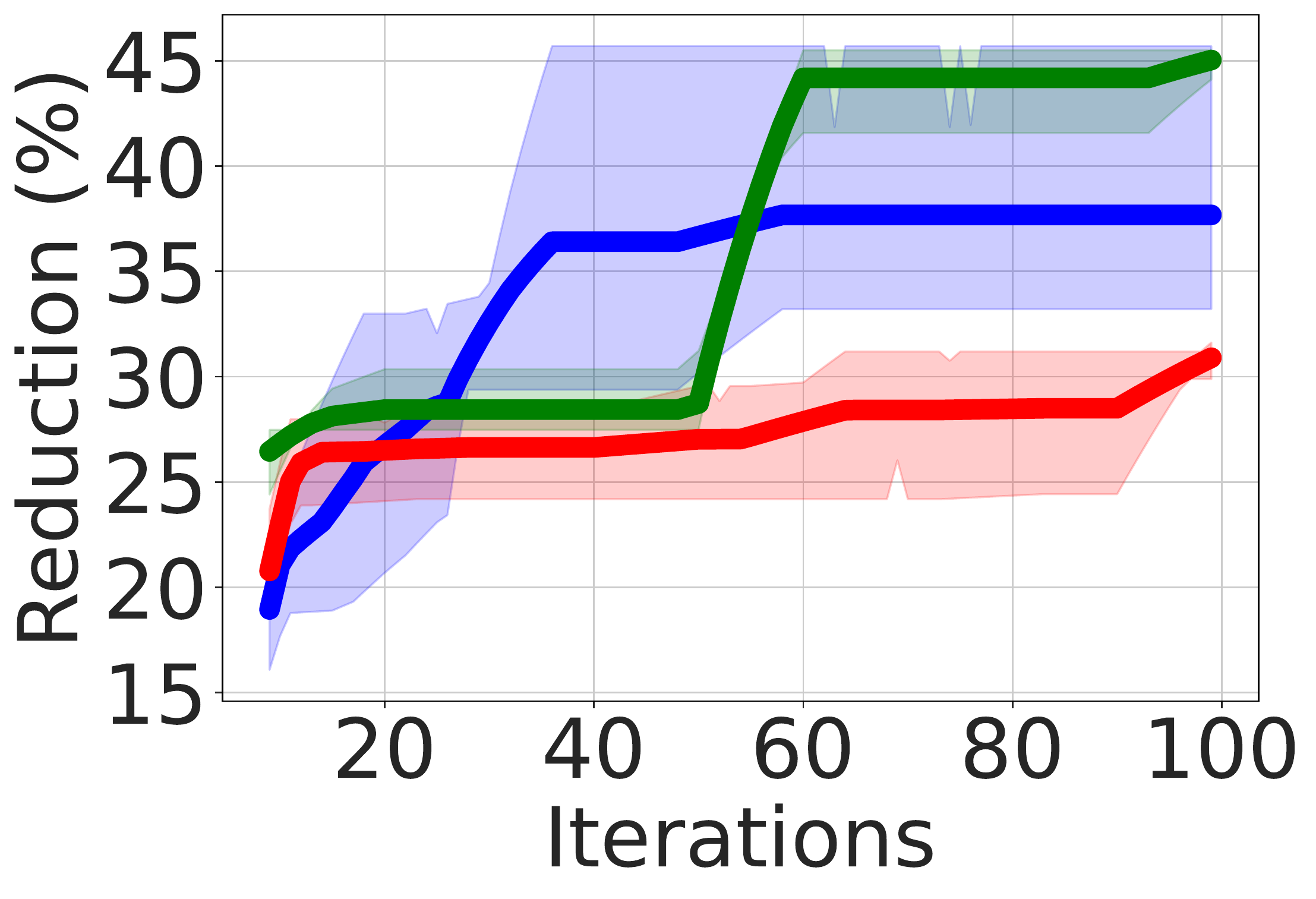} }}
	}
	\subfloat[multiplier]{{{\includegraphics[width=0.25\columnwidth, valign=c]{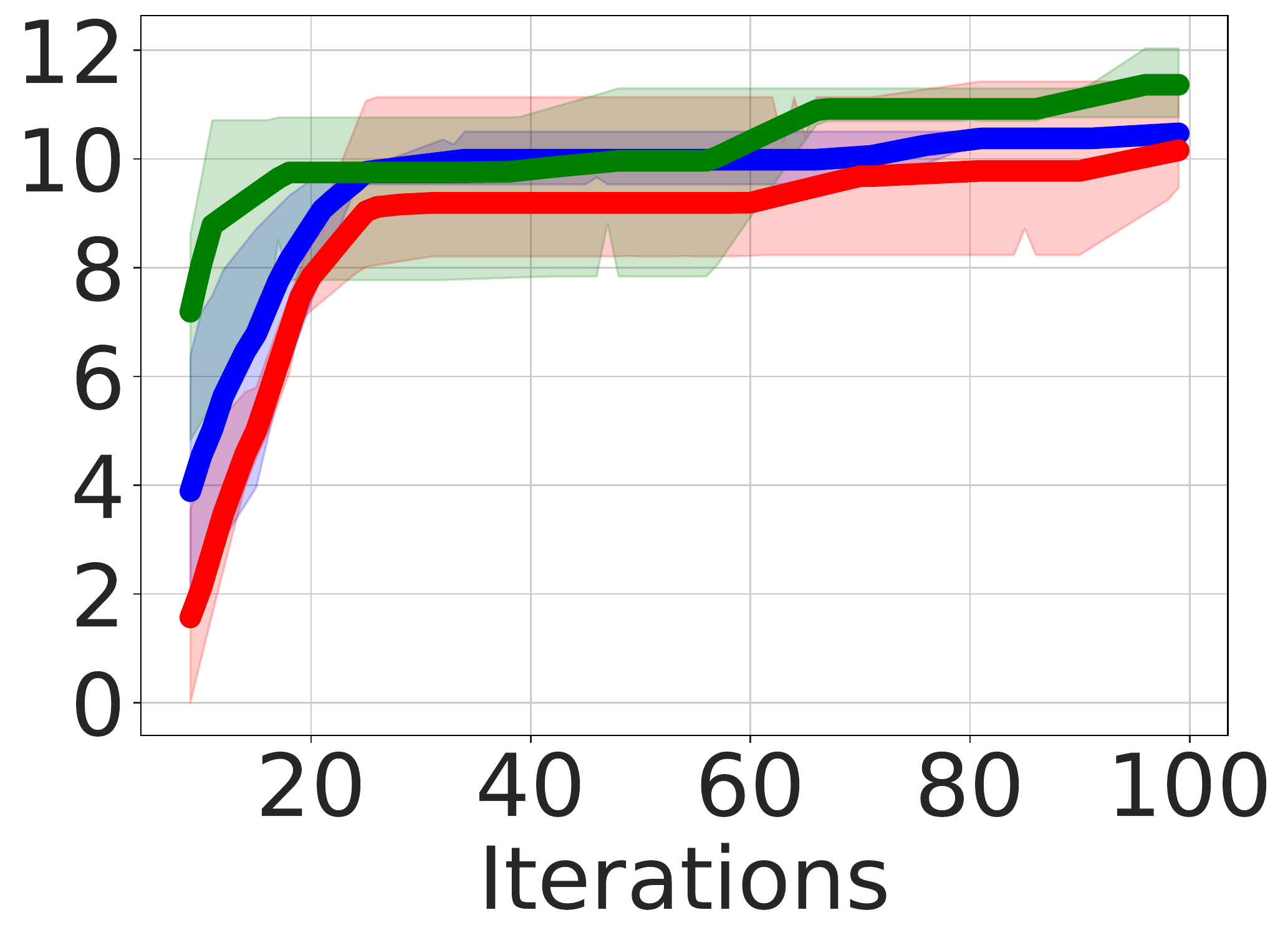} }}
	}
        \subfloat[sqrt]{{{\includegraphics[width=0.25\columnwidth, valign=c]{figures/2023_neurips_plot_sqrt_compare_ADP_best_sqrt.pdf} }}
	}
	\subfloat[\color{red}{square$^*$}]{{{\includegraphics[width=0.25\columnwidth, valign=c]{figures/2023_neurips_plot_square_compare_ADP_best_square.pdf} }}
	}
    \caption{Area-delay product reduction (in \%) compared to resyn2 on EPFL arithmetic benchmarks. We evaluated \texttt{bar} and \texttt{square} with arithmetic agent I, \texttt{adder}, \texttt{sqrt} and \texttt{log2} with arithmetic agent II, \texttt{multiplier} and \texttt{sin} with arithmetic agent III and \texttt{div} and \texttt{max} with arithmetic agent IV. GREEN: INVICTUS ($\alpha=1$), BLUE: MCTS~\cite{neto2022flowtune}, RED: SA+Pred.~\cite{bullseye}}
    \label{fig:performance-epfl-arith2}
\end{figure}

\subsection{EPFL random control benchmarks \label{subsec:result-epfl-rc}}

\begin{figure}[h]
\centering
   \subfloat[\color{red}{cavlc$^*$}]{{{\includegraphics[width=0.25\columnwidth, valign=c]{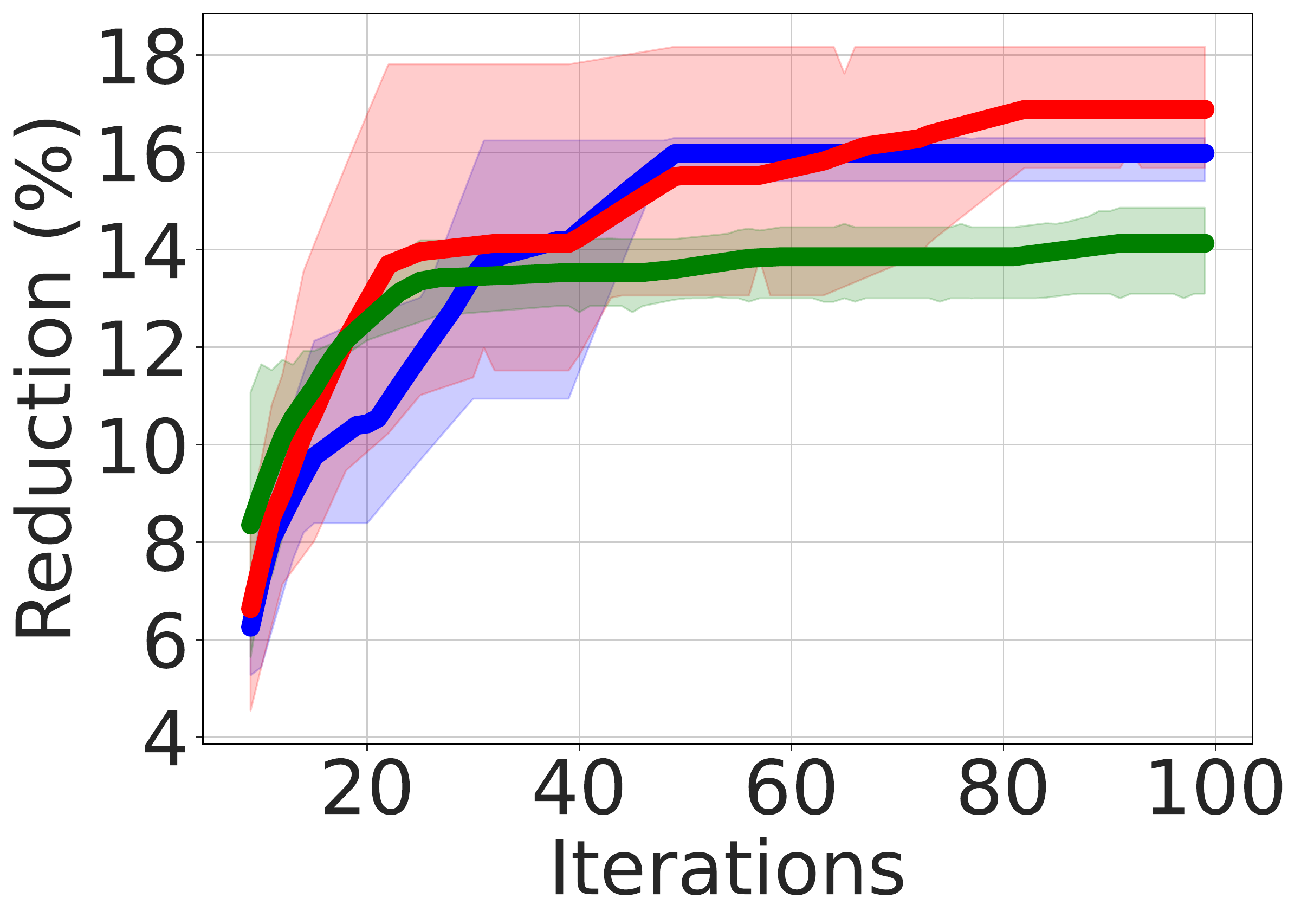} }}%
	}
	\subfloat[ctrl]{{{\includegraphics[width=0.25\columnwidth, valign=c]{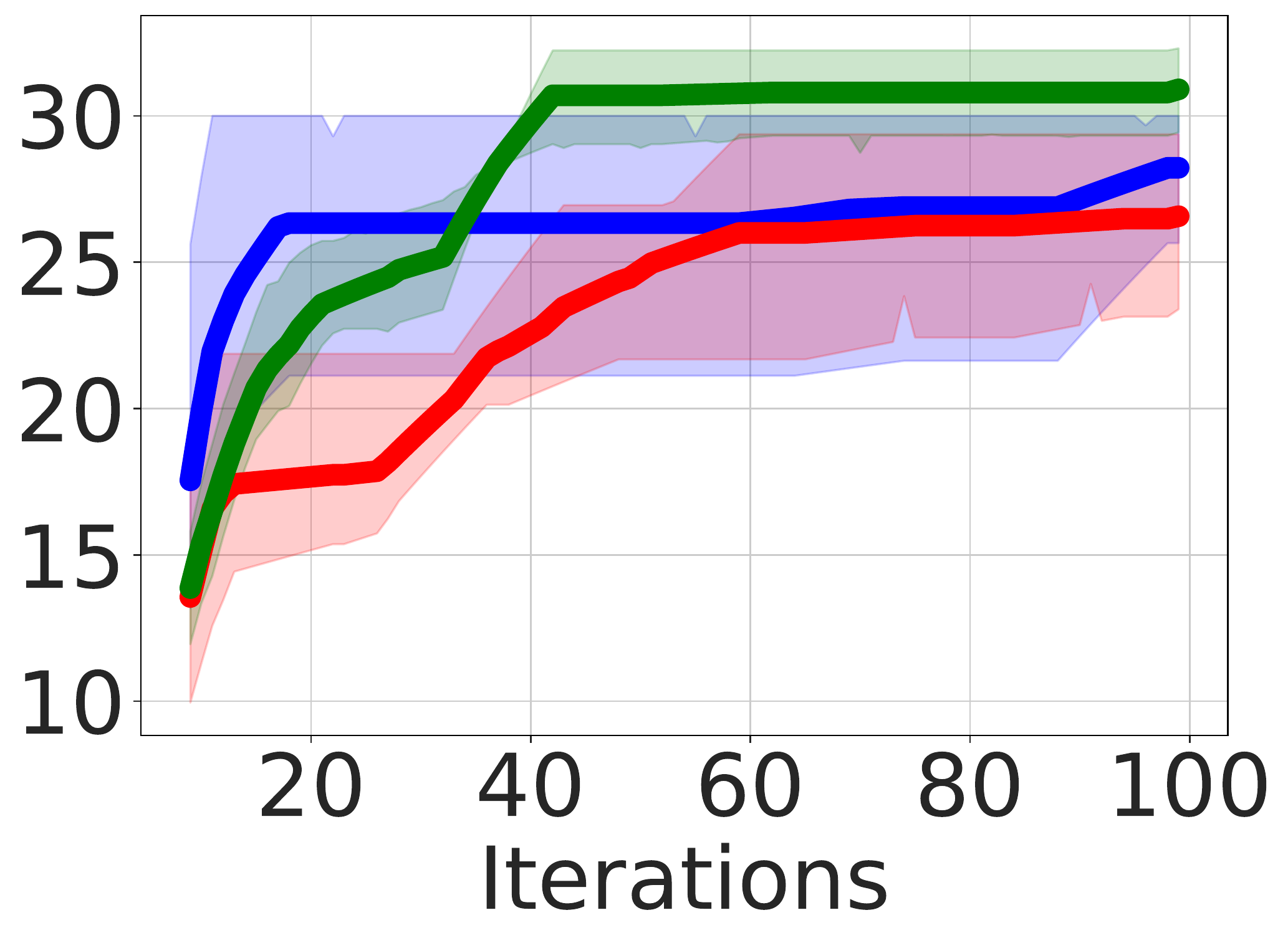} }}
	}
        \subfloat[\color{red}{int2float$^*$}]{{{\includegraphics[width=0.25\columnwidth, valign=c]{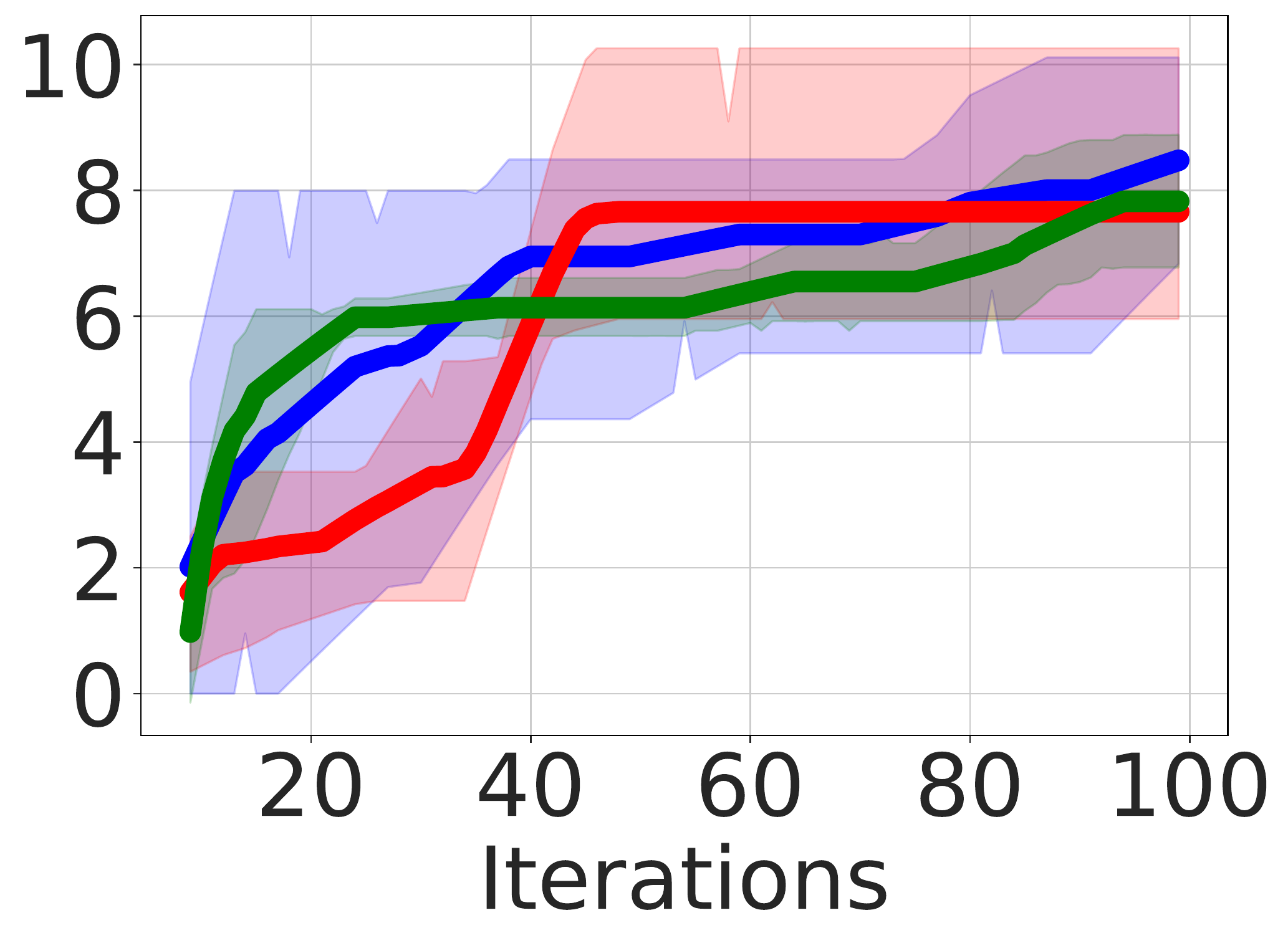} }}
	}
	\subfloat[i2c]{{{\includegraphics[width=0.25\columnwidth, valign=c]{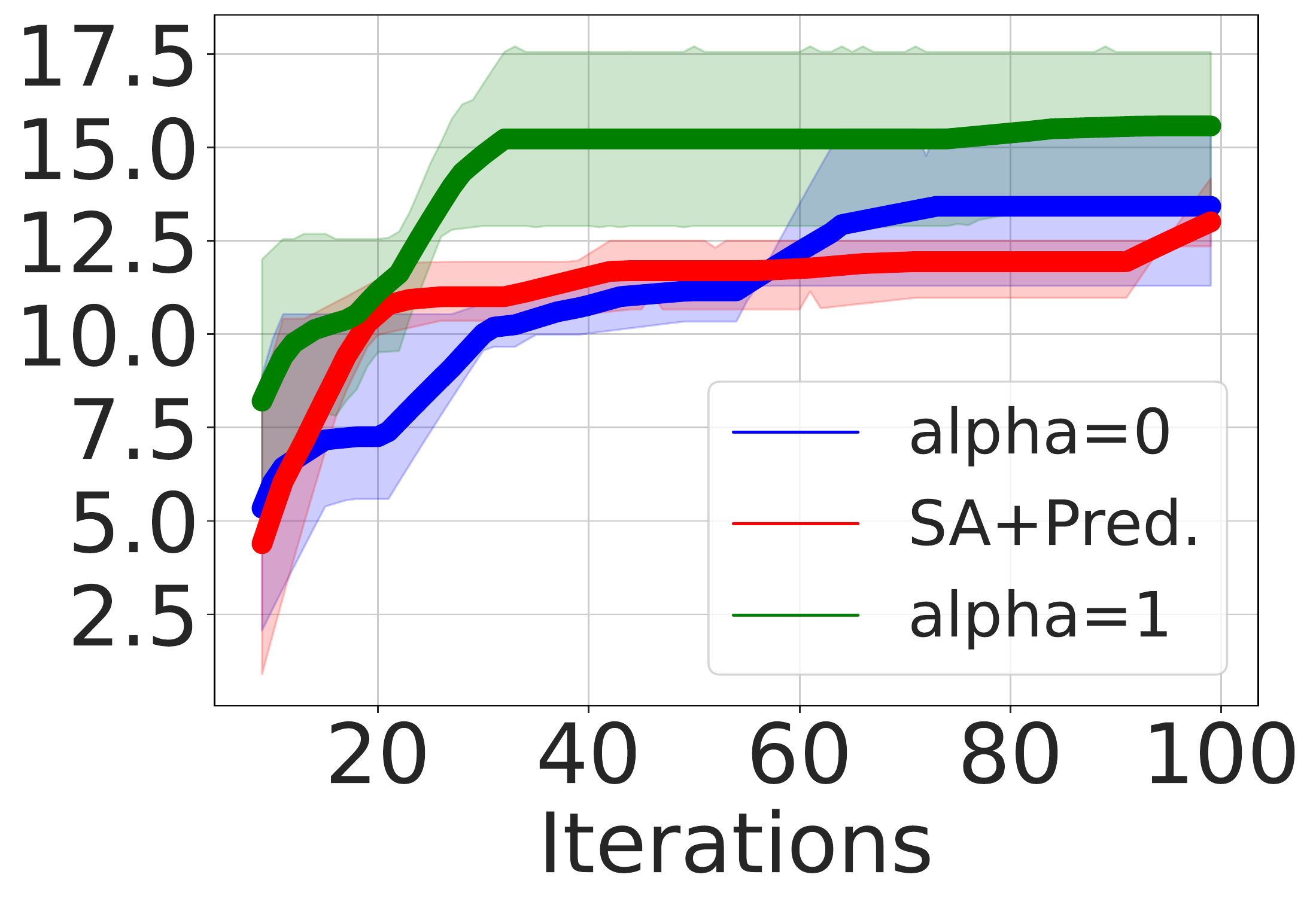} }}
	}
	
\vspace{0.2in}
	\subfloat[mem\_ctrl]{{{\includegraphics[width=0.25\columnwidth, valign=c]{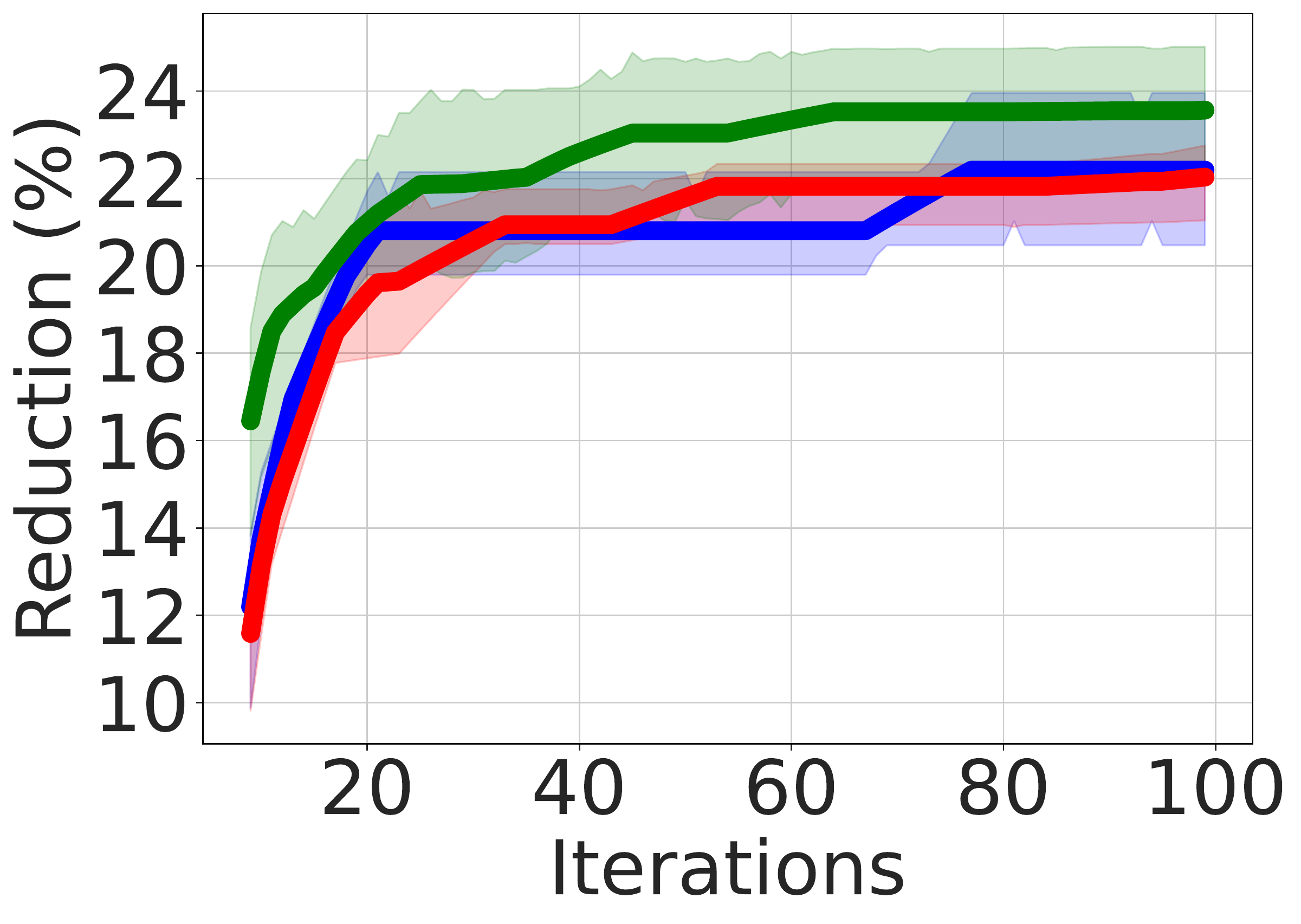} }}%
	}
	\subfloat[priority]{{{\includegraphics[width=0.25\columnwidth, valign=c]{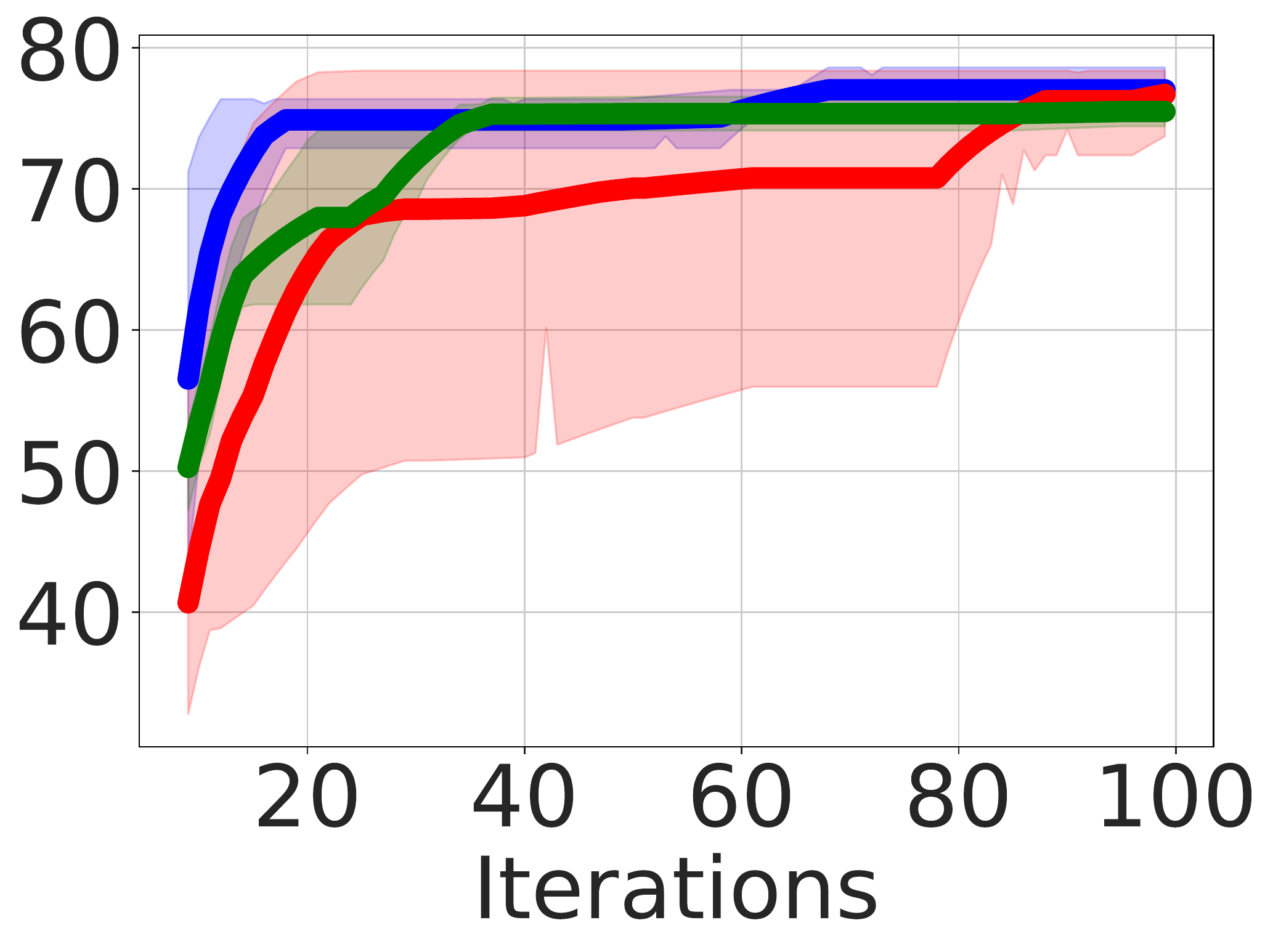} }}
	}
        \subfloat[\textcolor{red}{router$^*$}]{{{\includegraphics[width=0.25\columnwidth, valign=c]{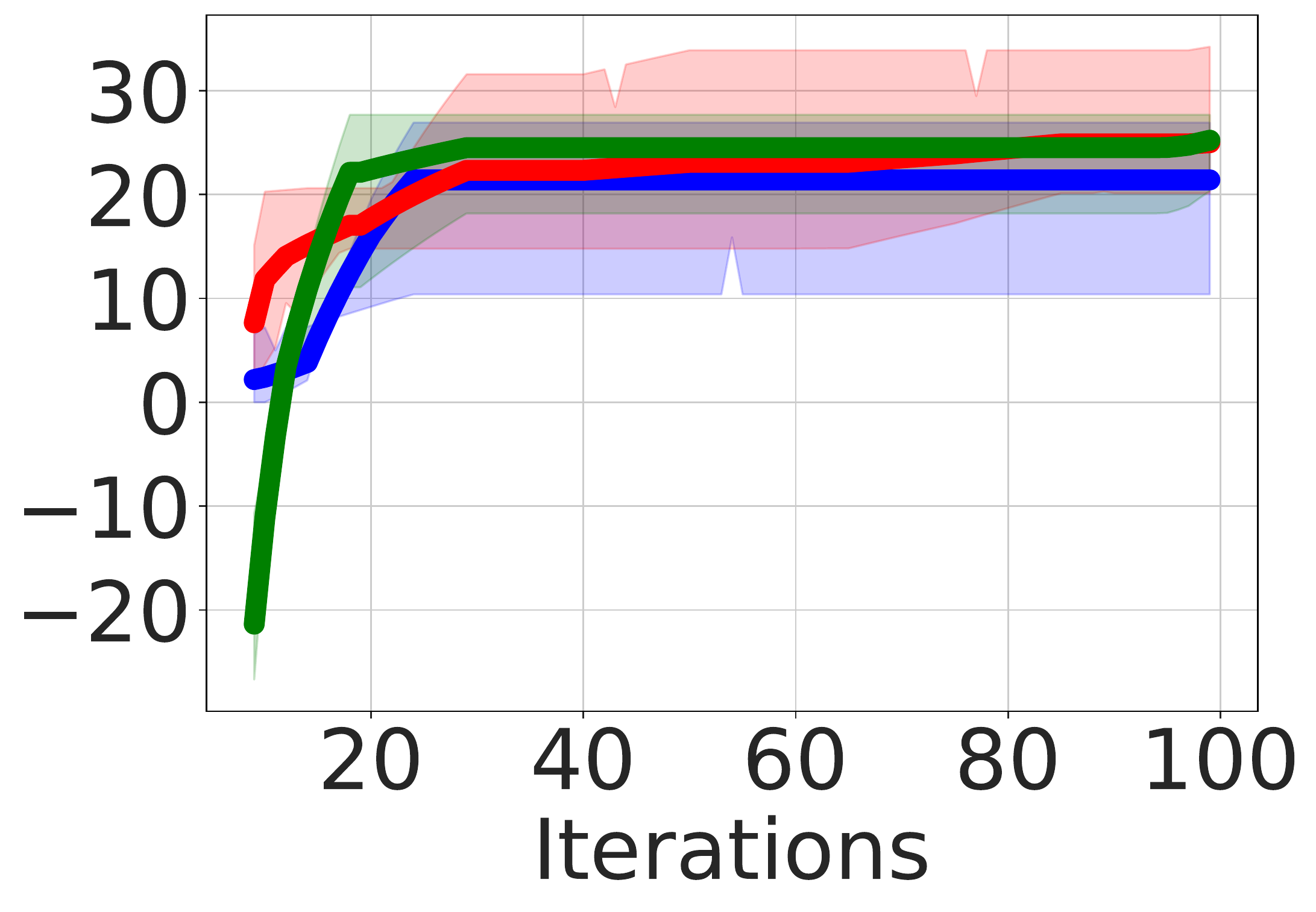} }}
	}
	\subfloat[\color{red}{voter$^*$}]{{{\includegraphics[width=0.25\columnwidth, valign=c]{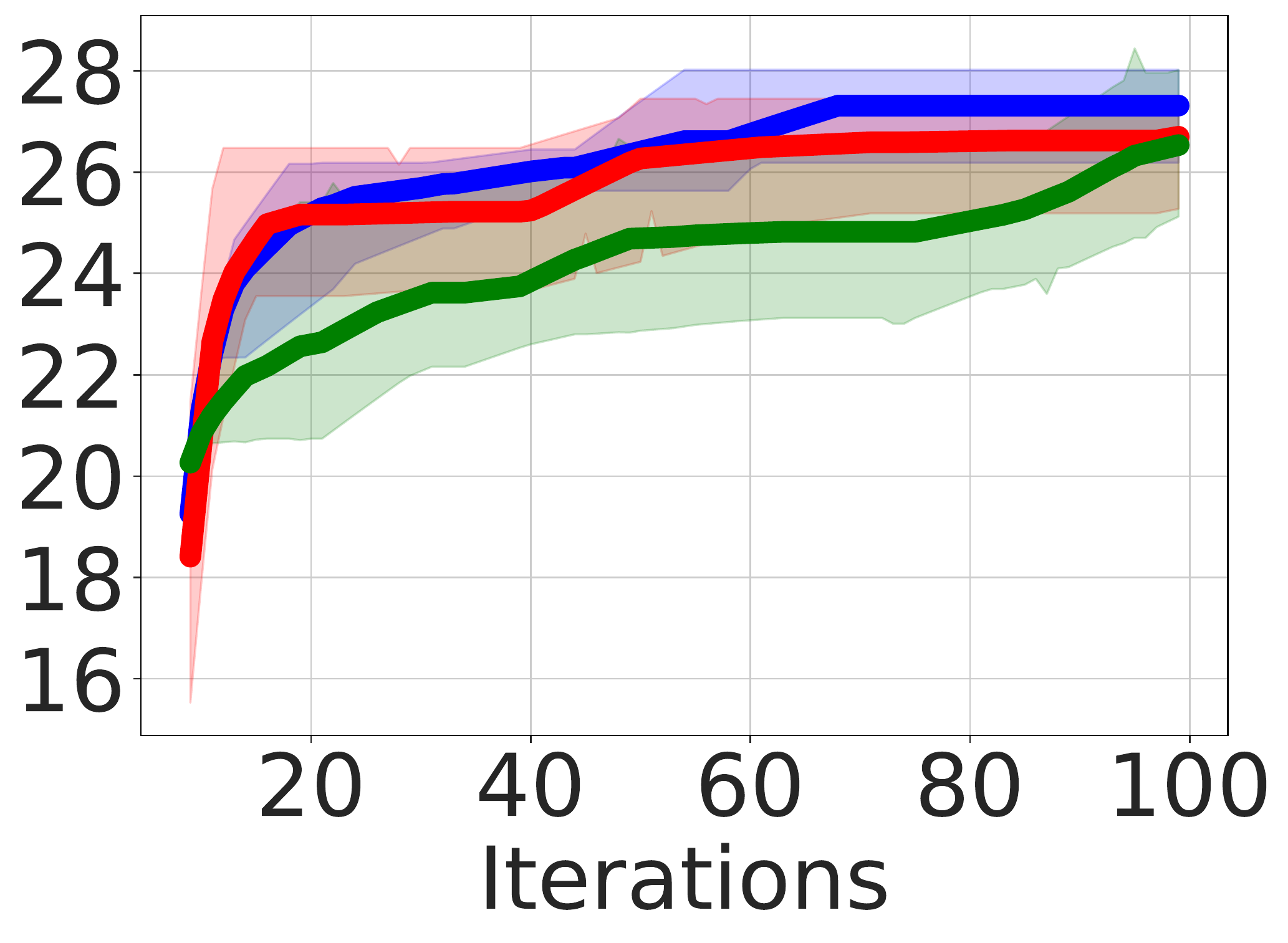} }}
	}
    \caption{Area-delay product reduction (in \%) compared to resyn2 on EPFL random control benchmarks. Except \texttt{router}, our OOD detector successfully identified the average winner approach. GREEN: INVICTUS ($\alpha=1$), BLUE: MCTS~\cite{neto2022flowtune}, RED: SA+Pred.~\cite{bullseye}}
    \label{fig:performance-epfl-rc}
\end{figure}

\newpage
\section{Out-of-distribution detection}

We use cosine distance ($\Delta_{cos}$) based OOD detection to check similarity of a new ``unseen" design bear with training designs embeddings. Next, we report $\delta_{th}$ obtained for various RL agents trained in our experiments using validation data. (See \autoref{sec:empiricalEval}). ~\autoref{tab:aurocMCNC} outlines the data for MCNC agent; ~\autoref{tab:aurocArith1}, \ref{tab:aurocArith2}, \ref{tab:aurocArith3} and \ref{tab:aurocArith4} presents the data for EPFL arithmetic agents and ~\autoref{tab:aurocRC1}, \ref{tab:aurocRC2}, \ref{tab:aurocRC3}, \ref{tab:aurocRC4} and \ref{tab:aurocRC5} report the numbers for EPFL random control agents.

\begin{table}[h]
\centering
\caption{$\Delta_{cosine}$ on validation data (MCNC agent). $\delta_{th}$ computed from AUROC curve: $0.007$}
\scriptsize
\setlength{\tabcolsep}{4pt}
\begin{tabular}{lrrrrrrrrrrrrrrrc}
\toprule
\multirow{2}{*}{ \begin{tabular}{@{}c@{}}Validation\\benchmarks\end{tabular}} & \multicolumn{14}{c}{$\Delta_{cos}$($h_{valid}$, $h_{train}$)} & \multirow{2}{*}{$\delta_{min}$} & \multirow{2}{*}{ \begin{tabular}{@{}c@{}}Winner\\
(0-MCTS+RL\\/1- MCTS)\end{tabular} } \\
\cmidrule(lr){2-15}
& alu2 & apex3 & apex5 & b2 & c1355 & c5315 & c2670 & prom2 & frg1 & i7 & i8 & m3 & max512 & table5 & & \\
\midrule
apex7 & 0.003 & 0.046 & 0.016 & 0.042 & 0.041 & 0.016 & 0.019 & 0.07 & 0.04 & 0.052 & 0.048 & 0.026 & 0.025 & 0.032 & 0.003 & 0 \\
c1908 & 0.022 & 0.098 & 0.036 & 0.088 & 0.009 & 0.039 & 0.043 & 0.125 & 0.023 & 0.11 & 0.103 & 0.047 & 0.059 & 0.076 & 0.009 & 0 \\
c3540 & 0.002 & 0.04 & 0.02 & 0.033 & 0.034 & 0.016 & 0.019 & 0.059 & 0.029 & 0.053 & 0.045 & 0.017 & 0.016 & 0.026 & 0.002 & 0 \\
frg2 & 0.03 & 0.022 & 0.016 & 0.023 & 0.085 & 0.007 & 0.006 & 0.034 & 0.077 & 0.038 & 0.03 & 0.017 & 0.019 & 0.015 & 0.006 & 1 \\
max128 & 0.018 & 0.023 & 0.016 & 0.023 & 0.067 & 0.004 & 0.004 & 0.038 & 0.058 & 0.043 & 0.034 & 0.014 & 0.014 & 0.015 & 0.004 & 0 \\
apex6 & 0.018 & 0.071 & 0.025 & 0.063 & 0.033 & 0.026 & 0.029 & 0.094 & 0.033 & 0.081 & 0.076 & 0.032 & 0.043 & 0.053 & 0.018 & 1 \\
c432 & 0.009 & 0.034 & 0.04 & 0.025 & 0.042 & 0.025 & 0.025 & 0.041 & 0.026 & 0.067 & 0.052 & 0.008 & 0.013 & 0.024 & 0.008 & 1 \\
c499 & 0.03 & 0.124 & 0.064 & 0.112 & 0.006 & 0.059 & 0.055 & 0.153 & 0.012 & 0.147 & 0.136 & 0.069 & 0.078 & 0.1 & 0.006 & 1 \\
seq & 0.028 & 0.009 & 0.033 & 0.012 & 0.115 & 0.027 & 0.035 & 0.008 & 0.098 & 0.01 & 0.016 & 0.015 & 0.015 & 0.01 & 0.008 & 1 \\
table3 & 0.028 & 0.01 & 0.036 & 0.025 & 0.112 & 0.028 & 0.033 & 0.037 & 0.094 & 0.014 & 0.007 & 0.012 & 0.014 & 0.01 & 0.007 & 1 \\
i10 & 0.017 & 0.007 & 0.02 & 0.006 & 0.091 & 0.016 & 0.022 & 0.018 & 0.078 & 0.012 & 0.008 & 0.011 & 0.003 & 0.002 & 0.002 & 0 \\
\bottomrule
\end{tabular}
\label{tab:aurocMCNC}
\end{table}

\begin{table}[h]
\centering
\caption{$\Delta_{cosine}$ on validation data (EPFL arithmetic agent I). $\delta_{th}$ computed from AUROC curve: $0.005$}
\footnotesize
\begin{tabular}{lrrrrrrrrrc}
\toprule
\multirow{2}{*}{ \begin{tabular}{@{}c@{}}Validation\\benchmarks\end{tabular}} & \multicolumn{6}{c}{$\Delta_{cos}$($h_{valid}$, $h_{train}$)} & & & \multirow{2}{*}{$\delta_{min}$} & \multirow{2}{*}{ \begin{tabular}{@{}c@{}}Winner\\
(0-MCTS+RL\\/1- MCTS)\end{tabular} } \\
\cmidrule(lr){2-8}
& adder & div & log2 & multiplier & sin & sqrt & max & & & \\

\midrule
adder\_alu2 & 0.094 & 0.047 & 0.04 & 0.025 & 0.014 & 0.025 & 0.071 & & 0.014 & 1 \\
div\_alu2 & 0.228 & 0 & 0.001 & 0.004 & 0.011 & 0.005 & 0.112 & & 0 & 0 \\
log2\_alu2 & 0.219 & 0.001 & 0 & 0.002 & 0.008 & 0.004 & 0.051 & & 0 & 0 \\
multiplier\_alu2 & 0.182 & 0.004 & 0.002 & 0 & 0.002 & 0.001 & 0.009 & & 0 & 0 \\
sin\_alu2 & 0.152 & 0.012 & 0.008 & 0.003 & 0 & 0.003 & 0.011 & & 0 & 0 \\
sqrt\_alu2 & 0.173 & 0.005 & 0.004 & 0.001 & 0.002 & 0 & 0.004 & & 0 & 0 \\
max\_alu2 & 0.24 & 0.003 & 0.006 & 0.011 & 0.02 & 0.01 & 0.001 & & 0.001 & 0 \\
& & & & & & & & & & \\
adder\_apex7 & 0.075 & 0.092 & 0.082 & 0.06 & 0.044 & 0.062 & 0.078 & & 0.044 & 1 \\
div\_apex7 & 0.228 & 0 & 0.001 & 0.004 & 0.011 & 0.005 & 0.094 & & 0 & 0 \\
log2\_apex7 & 0.219 & 0.001 & 0 & 0.002 & 0.008 & 0.004 & 0.056 & & 0 & 0 \\
multiplier\_apex7 & 0.182 & 0.004 & 0.002 & 0 & 0.002 & 0.001 & 0.011 & & 0 & 0 \\
sin\_apex7 & 0.153 & 0.011 & 0.008 & 0.002 & 0 & 0.003 & 0.019 & & 0 & 0 \\
sqrt\_apex7 & 0.175 & 0.004 & 0.003 & 0.001 & 0.002 & 0 & 0.008 & & 0 & 0 \\
max\_apex7 & 0.242 & 0.003 & 0.006 & 0.012 & 0.021 & 0.011 & 0.003 & & 0.003 & 0 \\
& & & & & & & & & & \\
arbiter & 0.123 & 0.059 & 0.049 & 0.034 & 0.023 & 0.037 & 0.049 & & 0.023 & 1 \\
cavlc & 0.124 & 0.025 & 0.02 & 0.009 & 0.004 & 0.009 & 0.011 & & 0.004 & 0 \\
priority & 0.014 & 0.21 & 0.2 & 0.163 & 0.131 & 0.157 & 0.043 & & 0.014 & 1 \\
router & 0.25 & 0.492 & 0.478 & 0.449 & 0.414 & 0.448 & 0.024 & & 0.25 & 1 \\
\bottomrule
\end{tabular}
\label{tab:aurocArith1}
\end{table}

\begin{table}[h]
\centering
\caption{$\Delta_{cos}$ on validation data (EPFL arithmetic agent II). $\delta_{th}$ computed from AUROC curve: $0.006$}
\footnotesize
\resizebox{\columnwidth}{!}{
\begin{tabular}{lrrrrrrrrc}
\toprule
\multirow{2}{*}{ \begin{tabular}{@{}c@{}}Validation\\benchmarks\end{tabular}} & \multicolumn{6}{c}{$\Delta_{cos}$($h_{valid}$, $h_{train}$)} & & \multirow{2}{*}{$\delta_{min}$} & \multirow{2}{*}{ \begin{tabular}{@{}c@{}}Winner\\
(0-MCTS+RL\\/1- MCTS)\end{tabular} } \\
\cmidrule(lr){2-7}
& bar & div & max & multiplier & sin & square & & & \\
\midrule
bar\_alu2 & 0 & 0.005 & 0.014 & 0.009 & 0.018 & 0.071 & & 0 & 0 \\
div\_alu2 & 0.005 & 0 & 0.003 & 0.004 & 0.011 & 0.082 & & 0 & 0 \\
max\_alu2 & 0.014 & 0.003 & 0 & 0.011 & 0.02 & 0.112 & & 0 & 0 \\
multiplier\_alu2 & 0.009 & 0.004 & 0.011 & 0 & 0.002 & 0.054 & & 0 & 0 \\
sin\_alu2 & 0.018 & 0.012 & 0.021 & 0.003 & 0 & 0.043 & & 0 & 0 \\
square\_alu2 & 0.06 & 0.069 & 0.096 & 0.043 & 0.033 & 0.001 & & 0.001 & 0 \\
& & & & & & & & & \\
bar\_apex7 & 0 & 0.005 & 0.014 & 0.009 & 0.018 & 0.071 & & 0 & 0 \\
div\_apex7 & 0.005 & 0 & 0.003 & 0.004 & 0.011 & 0.082 & & 0 & 0 \\
max\_apex7 & 0.014 & 0.003 & 0 & 0.012 & 0.021 & 0.113 & & 0 & 0 \\
multiplier\_apex7 & 0.009 & 0.004 & 0.011 & 0 & 0.002 & 0.055 & & 0 & 0 \\
sin\_apex7 & 0.018 & 0.011 & 0.02 & 0.002 & 0 & 0.044 & & 0 & 0 \\
square\_apex7 & 0.07 & 0.081 & 0.112 & 0.054 & 0.043 & 0 & & 0 & 0 \\
& & & & & & & & & \\
arbiter & 0.055 & 0.059 & 0.083 & 0.034 & 0.023 & 0.008 & & 0.008 & 1 \\
cavlc & 0.051 & 0.047 & 0.064 & 0.025 & 0.014 & 0.018 & & 0.014 & 1 \\
priority & 0.236 & 0.21 & 0.228 & 0.163 & 0.131 & 0.108 & & 0.108 & 1 \\
router & 0.508 & 0.492 & 0.504 & 0.449 & 0.414 & 0.386 & & 0.386 & 1 \\
\bottomrule
\end{tabular}
}
\label{tab:aurocArith2}
\end{table}

\begin{table}[h]
\centering
\caption{$\Delta_{cosine}$ on validation data (EPFL arithmetic agent III). $\delta_{th}$ computed from AUROC curve: $0.01$}
\footnotesize
\begin{tabular}{lrrrrrrrrrc}
\toprule
\multirow{2}{*}{ \begin{tabular}{@{}c@{}}Validation\\benchmarks\end{tabular}} & \multicolumn{6}{c}{$\Delta_{cos}$($h_{valid}$, $h_{train}$)} & & & \multirow{2}{*}{$\delta_{min}$} & \multirow{2}{*}{ \begin{tabular}{@{}c@{}}Winner\\
(0-MCTS+RL\\/1- MCTS)\end{tabular} } \\
\cmidrule(lr){2-8}
& adder & bar & div & log2 & max & square & sqrt & & & \\

\midrule
adder\_alu2 & 0.094 & 0.051 & 0.047 & 0.04 & 0.064 & 0.01 & 0.025 & & 0.01 & 0 \\
bar\_alu2 & 0.257 & 0 & 0.005 & 0.004 & 0.014 & 0.071 & 0.012 & & 0 & 0 \\
div\_alu2 & 0.228 & 0.005 & 0 & 0.001 & 0.003 & 0.082 & 0.005 & & 0 & 0 \\
log2\_alu2 & 0.219 & 0.004 & 0.001 & 0 & 0.006 & 0.07 & 0.004 & & 0 & 0 \\
max\_alu2 & 0.24 & 0.014 & 0.003 & 0.006 & 0 & 0.112 & 0.01 & & 0 & 0 \\
square\_alu2 & 0.133 & 0.06 & 0.069 & 0.058 & 0.096 & 0.001 & 0.048 & & 0.001 & 0 \\
sqrt\_alu2 & 0.173 & 0.012 & 0.005 & 0.004 & 0.01 & 0.059 & 0 & & 0 & 0 \\
& & & & & & & & & & \\
adder\_apex7 & 0.075 & 0.093 & 0.092 & 0.082 & 0.117 & 0.01 & 0.062 & & 0.01 & 0 \\
bar\_apex7 & 0.259 & 0 & 0.005 & 0.004 & 0.014 & 0.071 & 0.013 & & 0 & 0 \\
div\_apex7 & 0.228 & 0.005 & 0 & 0.001 & 0.003 & 0.082 & 0.005 & & 0 & 0 \\
log2\_apex7 & 0.219 & 0.004 & 0.001 & 0 & 0.006 & 0.07 & 0.004 & & 0 & 0 \\
max\_apex7 & 0.242 & 0.014 & 0.003 & 0.006 & 0 & 0.113 & 0.011 & & 0 & 0 \\
square\_apex7 & 0.14 & 0.07 & 0.081 & 0.07 & 0.112 & 0 & 0.059 & & 0 & 0 \\
sqrt\_apex7 & 0.175 & 0.012 & 0.004 & 0.003 & 0.01 & 0.059 & 0 & & 0 & 0 \\
& & & & & & & & & & \\
arbiter & 0.123 & 0.055 & 0.059 & 0.049 & 0.083 & 0.008 & 0.037 & & 0.008 & 0 \\
cavlc & 0.124 & 0.03 & 0.025 & 0.02 & 0.038 & 0.027 & 0.009 & & 0.009 & 0 \\
priority & 0.014 & 0.236 & 0.21 & 0.2 & 0.228 & 0.108 & 0.157 & & 0.014 & 1 \\
router & 0.25 & 0.508 & 0.492 & 0.478 & 0.504 & 0.386 & 0.448 & & 0.25 & 1 \\
\bottomrule
\end{tabular}
\label{tab:aurocArith3}
\end{table}

\begin{table}[h]
\centering
\caption{$\Delta_{cosine}$ on validation data (EPFL arithmetic agent IV). $\delta_{th}$ computed from AUROC curve: $0.01$}
\footnotesize
\begin{tabular}{lrrrrrrrrrc}
\toprule
\multirow{2}{*}{ \begin{tabular}{@{}c@{}}Validation\\benchmarks\end{tabular}} & \multicolumn{6}{c}{$\Delta_{cos}$($h_{valid}$, $h_{train}$)} & & & \multirow{2}{*}{$\delta_{min}$} & \multirow{2}{*}{ \begin{tabular}{@{}c@{}}Winner\\
(0-MCTS+RL\\/1- MCTS)\end{tabular} } \\
\cmidrule(lr){2-8}
& adder & bar & log2 & multiplier & sin & square & sqrt & & & \\

\midrule
adder\_alu2 & 0.094 & 0.051 & 0.04 & 0.025 & 0.01 & 0.018 & 0.025 & & 0.01 & 0 \\
bar\_alu2 & 0.257 & 0 & 0.004 & 0.009 & 0.018 & 0.071 & 0.012 & & 0 & 0 \\
log2\_alu2 & 0.219 & 0.004 & 0 & 0.002 & 0.008 & 0.07 & 0.004 & & 0 & 0 \\
multiplier\_alu2 & 0.182 & 0.009 & 0.002 & 0 & 0.002 & 0.054 & 0.001 & & 0 & 0 \\
sin\_alu2 & 0.152 & 0.018 & 0.008 & 0.003 & 0 & 0.043 & 0.003 & & 0 & 0 \\
square\_alu2 & 0.133 & 0.06 & 0.058 & 0.043 & 0.033 & 0.001 & 0.048 & & 0.001 & 0 \\
sqrt\_alu2 & 0.173 & 0.012 & 0.004 & 0.001 & 0.002 & 0.059 & 0 & & 0 & 0 \\
& & & & & & & & & & \\
adder\_apex7 & 0.075 & 0.093 & 0.082 & 0.06 & 0.044 & 0.01 & 0.062 & & 0.01 & 0 \\
bar\_apex7 & 0.259 & 0 & 0.004 & 0.009 & 0.018 & 0.071 & 0.013 & & 0 & 0 \\
log2\_apex7 & 0.219 & 0.004 & 0 & 0.002 & 0.008 & 0.07 & 0.004 & & 0 & 0 \\
multiplier\_apex7 & 0.182 & 0.009 & 0.002 & 0 & 0.002 & 0.055 & 0.001 & & 0 & 0 \\
sin\_apex7 & 0.153 & 0.018 & 0.008 & 0.002 & 0 & 0.044 & 0.003 & & 0 & 0 \\
square\_apex7 & 0.14 & 0.07 & 0.07 & 0.054 & 0.043 & 0 & 0.059 & & 0 & 0 \\
sqrt\_apex7 & 0.175 & 0.012 & 0.003 & 0.001 & 0.002 & 0.059 & 0 & & 0 & 0 \\
& & & & & & & & & & \\
arbiter & 0.123 & 0.055 & 0.049 & 0.034 & 0.023 & 0.008 & 0.037 & & 0.008 & 0 \\
cavlc & 0.124 & 0.03 & 0.02 & 0.009 & 0.004 & 0.027 & 0.009 & & 0.004 & 0 \\
priority & 0.014 & 0.236 & 0.2 & 0.163 & 0.131 & 0.108 & 0.157 & & 0.014 & 1 \\
router & 0.25 & 0.508 & 0.478 & 0.449 & 0.414 & 0.385 & 0.448 & & 0.25 & 1 \\
\bottomrule
\end{tabular}
\label{tab:aurocArith4}
\end{table}

\begin{table}[h]
\centering
\caption{$\Delta_{cosine}$ on validation data (EPFL random control agent I). $\delta_{th}$ computed from AUROC curve: $0.008$}
\footnotesize
\begin{tabular}{lrrrrrrrrrc}
\toprule
\multirow{2}{*}{ \begin{tabular}{@{}c@{}}Validation\\benchmarks\end{tabular}} & \multicolumn{6}{c}{$\Delta_{cos}$($h_{valid}$, $h_{train}$)} & & & \multirow{2}{*}{$\delta_{min}$} & \multirow{2}{*}{ \begin{tabular}{@{}c@{}}Winner\\
(0-MCTS+RL\\/1- MCTS)\end{tabular} } \\
\cmidrule(lr){2-8}
& cavlc & ctrl & i2c & int2float & mem\_ctrl & priority & router & & & \\

\midrule
cavlc\_apex7 & 0.001 & 0.005 & 0.032 & 0.016 & 0.022 & 0.117 & 0.402 & & 0.001 & 0 \\
ctrl\_apex7 & 0.006 & 0.001 & 0.02 & 0.027 & 0.025 & 0.123 & 0.386 & & 0.001 & 0 \\
i2c\_apex7 & 0.037 & 0.023 & 0 & 0.062 & 0.051 & 0.153 & 0.393 & & 0 & 0 \\
int2float\_apex7 & 0.008 & 0.017 & 0.049 & 0.002 & 0.057 & 0.073 & 0.339 & & 0.002 & 1 \\
mem\_ctrl\_apex7 & 0.029 & 0.031 & 0.051 & 0.072 & 0 & 0.233 & 0.498 & & 0 & 0 \\
priority\_apex7 & 0.031 & 0.047 & 0.088 & 0.005 & 0.108 & 0.042 & 0.314 & & 0.005 & 0 \\
router\_apex7 & 0.286 & 0.278 & 0.292 & 0.236 & 0.366 & 0.218 & 0.022 & & 0.022 & 0 \\
& & & & & & & & & & \\
cavlc\_alu2 & 0 & 0.006 & 0.034 & 0.013 & 0.027 & 0.106 & 0.398 & & 0 & 0 \\
ctrl\_alu2 & 0.004 & 0.002 & 0.024 & 0.018 & 0.037 & 0.096 & 0.377 & & 0.002 & 0 \\
i2c\_alu2 & 0.035 & 0.022 & 0 & 0.059 & 0.052 & 0.146 & 0.391 & & 0 & 0 \\
int2float\_alu2 & 0.007 & 0.017 & 0.049 & 0.001 & 0.059 & 0.065 & 0.35 & & 0.001 & 0 \\
mem\_ctrl\_alu2 & 0.029 & 0.031 & 0.051 & 0.072 & 0 & 0.233 & 0.498 & & 0 & 0 \\
priority\_alu2 & 0.008 & 0.018 & 0.051 & 0.001 & 0.062 & 0.062 & 0.35 & & 0.001 & 0 \\
router\_alu2 & 0.195 & 0.187 & 0.203 & 0.158 & 0.266 & 0.161 & 0.049 & & 0.049 & 0 \\
& & & & & & & & & & \\
adder & 0.124 & 0.12 & 0.156 & 0.079 & 0.255 & 0.014 & 0.25 & & 0.014 & 1 \\
bar & 0.03 & 0.031 & 0.05 & 0.075 & 0.001 & 0.236 & 0.508 & & 0.001 & 0 \\
square & 0.027 & 0.047 & 0.094 & 0.016 & 0.068 & 0.108 & 0.385 & & 0.016 & 1 \\
sqrt & 0.009 & 0.009 & 0.031 & 0.041 & 0.012 & 0.157 & 0.448 & & 0.009 & 1 \\
\bottomrule
\end{tabular}
\label{tab:aurocRC1}
\end{table}

\begin{table}[h]
\centering
\caption{$\Delta_{cosine}$ on validation data (EPFL random control agent II). $\delta_{th}$ computed from AUROC curve: $0.005$}
\footnotesize
\begin{tabular}{lrrrrrrrrrc}
\toprule
\multirow{2}{*}{ \begin{tabular}{@{}c@{}}Validation\\benchmarks\end{tabular}} & \multicolumn{6}{c}{$\Delta_{cos}$($h_{valid}$, $h_{train}$)} & & & \multirow{2}{*}{$\delta_{min}$} & \multirow{2}{*}{ \begin{tabular}{@{}c@{}}Winner\\
(0-MCTS+RL\\/1- MCTS)\end{tabular} } \\
\cmidrule(lr){2-8}
& arbiter & ctrl & i2c & int2float & mem\_ctrl & priority & voter & & & \\

\midrule
arbiter\_apex7 & 0 & 0.029 & 0.064 & 0.007 & 0.053 & 0.092 & 0.111 & & 0 & 0 \\
ctrl\_apex7 & 0.029 & 0.001 & 0.02 & 0.027 & 0.025 & 0.123 & 0.142 & & 0.001 & 0 \\
i2c\_apex7 & 0.069 & 0.023 & 0 & 0.062 & 0.051 & 0.153 & 0.176 & & 0 & 0 \\
int2float\_apex7 & 0.006 & 0.017 & 0.049 & 0.002 & 0.057 & 0.073 & 0.091 & & 0.002 & 0 \\
mem\_ctrl\_apex7 & 0.053 & 0.031 & 0.051 & 0.072 & 0 & 0.233 & 0.259 & & 0 & 0 \\
priority\_apex7 & 0.015 & 0.047 & 0.088 & 0.005 & 0.108 & 0.042 & 0.055 & & 0.005 & 0 \\
voter\_apex7 & 0.02 & 0.055 & 0.097 & 0.004 & 0.121 & 0.034 & 0.047 & & 0.004 & 0 \\
& & & & & & & & & & \\
arbiter\_alu2 & 0.001 & 0.022 & 0.054 & 0.006 & 0.047 & 0.09 & 0.11 & & 0.001 & 1 \\
ctrl\_alu2 & 0.024 & 0.002 & 0.024 & 0.018 & 0.037 & 0.096 & 0.113 & & 0.002 & 0 \\
i2c\_alu2 & 0.067 & 0.022 & 0 & 0.059 & 0.052 & 0.146 & 0.169 & & 0 & 0 \\
int2float\_alu2 & 0.006 & 0.017 & 0.049 & 0.001 & 0.059 & 0.065 & 0.082 & & 0.001 & 0 \\
mem\_ctrl\_alu2 & 0.053 & 0.031 & 0.051 & 0.072 & 0 & 0.233 & 0.259 & & 0 & 0 \\
priority\_alu2 & 0.007 & 0.018 & 0.051 & 0.001 & 0.062 & 0.062 & 0.078 & & 0.001 & 0 \\
voter\_alu2 & 0.008 & 0.019 & 0.053 & 0.001 & 0.065 & 0.059 & 0.075 & & 0.001 & 1 \\
& & & & & & & & & & \\
adder & 0.123 & 0.12 & 0.156 & 0.079 & 0.255 & 0.014 & 0.012 & & 0.012 & 1 \\
bar & 0.055 & 0.031 & 0.05 & 0.075 & 0.001 & 0.236 & 0.261 & & 0.001 & 0 \\
square & 0.008 & 0.047 & 0.094 & 0.016 & 0.068 & 0.108 & 0.127 & & 0.008 & 1 \\
sqrt & 0.037 & 0.009 & 0.031 & 0.041 & 0.012 & 0.157 & 0.177 & & 0.009 & 1 \\
\bottomrule
\end{tabular}
\label{tab:aurocRC2}
\end{table}

\begin{table}[h]
\centering
\caption{$\Delta_{cosine}$ on validation data (EPFL random control agent III). $\delta_{th}$ computed from AUROC curve: $0.007$}
\footnotesize
\begin{tabular}{lcccccccccc}
\toprule
\multirow{2}{*}{ \begin{tabular}{@{}c@{}}Validation\\benchmarks\end{tabular}} & \multicolumn{6}{c}{$\Delta_{cos}$($h_{valid}$, $h_{train}$)} & & & \multirow{2}{*}{$\delta_{min}$} & \multirow{2}{*}{ \begin{tabular}{@{}c@{}}Winner\\
(0-MCTS+RL\\/1- MCTS)\end{tabular} } \\
\cmidrule(lr){2-8}
& arbiter & cavlc & i2c & int2float & mem\_ctrl & router & voter & & & \\

\midrule
arbiter\_apex7 & 0 & 0.013 & 0.064 & 0.007 & 0.053 & 0.359 & 0.111 & & 0 & 0 \\
cavlc\_apex7 & 0.014 & 0.001 & 0.032 & 0.016 & 0.022 & 0.402 & 0.136 & & 0.001 & 0 \\
i2c\_apex7 & 0.069 & 0.037 & 0 & 0.062 & 0.051 & 0.393 & 0.176 & & 0 & 0 \\
int2float\_apex7 & 0.006 & 0.008 & 0.049 & 0.002 & 0.057 & 0.339 & 0.091 & & 0.002 & 1 \\
mem\_ctrl\_apex7 & 0.053 & 0.029 & 0.051 & 0.072 & 0 & 0.498 & 0.259 & & 0 & 0 \\
router\_apex7 & 0.249 & 0.286 & 0.292 & 0.236 & 0.366 & 0.022 & 0.229 & & 0.022 & 0 \\
voter\_apex7 & 0.02 & 0.037 & 0.097 & 0.007 & 0.121 & 0.317 & 0.047 & & 0.007 & 0 \\
& & & & & & & & & & \\
arbiter\_alu2 & 0.001 & 0.008 & 0.054 & 0.006 & 0.047 & 0.361 & 0.11 & & 0.001 & 0 \\
cavlc\_alu2 & 0.012 & 0 & 0.034 & 0.013 & 0.027 & 0.398 & 0.125 & & 0 & 0 \\
i2c\_alu2 & 0.067 & 0.035 & 0 & 0.059 & 0.052 & 0.391 & 0.169 & & 0 & 0 \\
int2float\_alu2 & 0.006 & 0.007 & 0.049 & 0.001 & 0.059 & 0.35 & 0.082 & & 0.001 & 0 \\
mem\_ctrl\_alu2 & 0.053 & 0.029 & 0.051 & 0.072 & 0 & 0.498 & 0.259 & & 0 & 0 \\
router\_alu2 & 0.171 & 0.195 & 0.203 & 0.158 & 0.266 & 0.049 & 0.173 & & 0.049 & 0 \\
voter\_alu2 & 0.008 & 0.008 & 0.053 & 0.001 & 0.065 & 0.351 & 0.075 & & 0.001 & 1 \\
& & & & & & & & & & \\
adder & 0.123 & 0.124 & 0.156 & 0.079 & 0.255 & 0.25 & 0.012 & & 0.012 & 1 \\
bar & 0.055 & 0.03 & 0.05 & 0.075 & 0.001 & 0.508 & 0.261 & & 0.001 & 0 \\
square & 0.008 & 0.027 & 0.094 & 0.016 & 0.068 & 0.385 & 0.127 & & 0.008 & 1 \\
sqrt & 0.037 & 0.009 & 0.031 & 0.041 & 0.012 & 0.448 & 0.177 & & 0.009 & 0 \\
\bottomrule
\end{tabular}
\label{tab:aurocRC3}
\end{table}

\begin{table}[h]
\centering
\caption{$\Delta_{cosine}$ on validation data (EPFL random control agent IV). $\delta_{th}$ computed from AUROC curve: $0.015$}
\footnotesize
\begin{tabular}{lccccccccccc}
\toprule
\multirow{2}{*}{ \begin{tabular}{@{}c@{}}Validation\\benchmarks\end{tabular}} & \multicolumn{7}{c}{$\Delta_{cos}$($h_{valid}$, $h_{train}$)} & & & \multirow{2}{*}{$\delta_{min}$} & \multirow{2}{*}{ \begin{tabular}{@{}c@{}}Winner\\
(0-MCTS+RL\\/1- MCTS)\end{tabular} } \\
\cmidrule(lr){2-9}
& arbiter & cavlc & ctrl & i2c & int2float & priority & router & voter & & & \\

\midrule
arbiter\_apex7 & 0 & 0.013 & 0.029 & 0.064 & 0.3231 & 0.092 & 0.359 & 0.111 & & 0 & 0 \\
cavlc\_apex7 & 0.014 & 0.001 & 0.005 & 0.032 & 0.3618 & 0.117 & 0.402 & 0.136 & & 0.001 & 1 \\
ctrl\_apex7 & 0.029 & 0.006 & 0.001 & 0.02 & 0.3474 & 0.123 & 0.386 & 0.142 & & 0.001 & 0 \\
i2c\_apex7 & 0.069 & 0.037 & 0.023 & 0 & 0.3537 & 0.153 & 0.393 & 0.176 & & 0 & 0 \\
int2float\_apex7 & 0.016 & 0.018 & 0.017 & 0.049 & 0.3051 & 0.073 & 0.339 & 0.091 & & 0.016 & 1 \\
priority\_apex7 & 0.015 & 0.031 & 0.047 & 0.088 & 0.2826 & 0.042 & 0.314 & 0.055 & & 0.015 & 0 \\
router\_apex7 & 0.249 & 0.286 & 0.278 & 0.292 & 0.0108 & 0.218 & 0.012 & 0.229 & & 0.011 & 0 \\
voter\_apex7 & 0.012 & 0.037 & 0.055 & 0.097 & 0.2853 & 0.034 & 0.317 & 0.047 & & 0.012 & 0 \\
& & & & & 0 & & & & & \\
arbiter\_alu2 & 0.001 & 0.008 & 0.022 & 0.054 & 0.3249 & 0.09 & 0.361 & 0.11 & & 0.001 & 0 \\
cavlc\_alu2 & 0.012 & 0 & 0.006 & 0.034 & 0.3582 & 0.106 & 0.398 & 0.125 & & 0 & 0 \\
ctrl\_alu2 & 0.024 & 0.004 & 0.002 & 0.024 & 0.3393 & 0.096 & 0.377 & 0.113 & & 0.002 & 0 \\
i2c\_alu2 & 0.067 & 0.035 & 0.022 & 0 & 0.3519 & 0.146 & 0.391 & 0.169 & & 0 & 0 \\
int2float\_alu2 & 0.006 & 0.007 & 0.017 & 0.049 & 0.315 & 0.065 & 0.35 & 0.082 & & 0.006 & 0 \\
priority\_alu2 & 0.007 & 0.008 & 0.018 & 0.051 & 0.315 & 0.062 & 0.35 & 0.078 & & 0.007 & 0 \\
router\_alu2 & 0.171 & 0.195 & 0.187 & 0.203 & 0.0441 & 0.161 & 0.049 & 0.173 & & 0.0441 & 0 \\
voter\_alu2 & 0.008 & 0.008 & 0.019 & 0.053 & 0.3159 & 0.059 & 0.351 & 0.075 & & 0.008 & 1 \\
& & & & & 0 & & & & & \\
adder & 0.123 & 0.124 & 0.12 & 0.156 & 0.225 & 0.019 & 0.25 & 0.016 & & 0.019 & 1 \\
bar & 0.055 & 0.03 & 0.031 & 0.05 & 0.4572 & 0.236 & 0.508 & 0.261 & & 0.03 & 1 \\
square & 0.008 & 0.027 & 0.047 & 0.094 & 0.3474 & 0.108 & 0.386 & 0.127 & & 0.008 & 0 \\
sqrt & 0.037 & 0.009 & 0.009 & 0.031 & 0.4032 & 0.157 & 0.448 & 0.177 & & 0.009 & 0 \\
\bottomrule
\end{tabular}
\label{tab:aurocRC4}
\end{table}

\begin{table}[h]
\centering
\caption{$\Delta_{cosine}$ on validation data (EPFL random control agent V). $\delta_{th}$ computed from AUROC curve: $0.005$}
\footnotesize
\begin{tabular}{lrrrrrrrrrr}
\toprule
\multirow{2}{*}{ \begin{tabular}{@{}c@{}}Validation\\benchmarks\end{tabular}} & \multicolumn{6}{c}{$\Delta_{cos}$($h_{valid}$, $h_{train}$)} & & & \multirow{2}{*}{$\delta_{min}$} & \multirow{2}{*}{ \begin{tabular}{@{}c@{}}Winner\\
(0-MCTS+RL\\/1- MCTS)\end{tabular} } \\
\cmidrule(lr){2-8}
& arbiter & cavlc & ctrl & mem\_ctrl & priority & router & voter & & & \\

\midrule
arbiter\_apex7 & 0.011 & 0.013 & 0.029 & 0.053 & 0.092 & 0.359 & 0.111 & & 0.011 & 1 \\
cavlc\_apex7 & 0.014 & 0.001 & 0.005 & 0.022 & 0.117 & 0.402 & 0.136 & & 0.001 & 0 \\
ctrl\_apex7 & 0.029 & 0.006 & 0.01 & 0.025 & 0.123 & 0.386 & 0.142 & & 0.006 & 1 \\
mem\_ctrl\_apex7 & 0.053 & 0.029 & 0.031 & 0 & 0.233 & 0.498 & 0.259 & & 0 & 0 \\
priority\_apex7 & 0.015 & 0.031 & 0.047 & 0.108 & 0.005 & 0.314 & 0.055 & & 0.005 & 0 \\
router\_apex7 & 0.249 & 0.286 & 0.278 & 0.366 & 0.218 & 0.01 & 0.229 & & 0.01 & 0 \\
voter\_apex7 & 0.02 & 0.037 & 0.055 & 0.121 & 0.034 & 0.317 & 0.004 & & 0.004 & 0 \\
& & & & & & & & & & \\
arbiter\_alu2 & 0.001 & 0.008 & 0.022 & 0.047 & 0.09 & 0.361 & 0.11 & & 0.001 & 0 \\
cavlc\_alu2 & 0.012 & 0 & 0.006 & 0.027 & 0.106 & 0.398 & 0.125 & & 0 & 0 \\
ctrl\_alu2 & 0.024 & 0.014 & 0.006 & 0.037 & 0.096 & 0.377 & 0.113 & & 0.006 & 1 \\
mem\_ctrl\_alu2 & 0.053 & 0.029 & 0.031 & 0 & 0.233 & 0.498 & 0.259 & & 0 & 0 \\
priority\_alu2 & 0.007 & 0.008 & 0.018 & 0.062 & 0.062 & 0.35 & 0.078 & & 0.007 & 1 \\
router\_alu2 & 0.171 & 0.195 & 0.187 & 0.266 & 0.161 & 0.049 & 0.173 & & 0.049 & 0 \\
voter\_alu2 & 0.008 & 0.008 & 0.019 & 0.065 & 0.059 & 0.351 & 0.075 & & 0.008 & 1 \\
& & & & & & & & & & \\
adder & 0.123 & 0.124 & 0.12 & 0.255 & 0.014 & 0.25 & 0.012 & & 0.012 & 1 \\
bar & 0.055 & 0.03 & 0.031 & 0.001 & 0.236 & 0.508 & 0.261 & & 0.001 & 0 \\
square & 0.008 & 0.027 & 0.047 & 0.068 & 0.108 & 0.385 & 0.127 & & 0.008 & 1 \\
sqrt & 0.037 & 0.009 & 0.004 & 0.012 & 0.157 & 0.448 & 0.177 & & 0.004 & 0 \\
\bottomrule
\end{tabular}
\label{tab:aurocRC5}
\end{table}





\end{document}